\documentclass{article}

\usepackage{arxiv}

\usepackage[utf8]{inputenc} 
\usepackage[T1]{fontenc}    
\usepackage{hyperref}       
\usepackage{url}            
\usepackage{booktabs}       
\usepackage{amsfonts}       
\usepackage{nicefrac}       
\usepackage{microtype}      
\usepackage{lipsum}
\usepackage{graphicx}
\usepackage{xcolor}
\usepackage{cite}
\usepackage{isomath}
\usepackage{amsmath, bm}
\usepackage{cleveref}
\usepackage[ruled,vlined]{algorithm2e}
\usepackage[most]{tcolorbox}
\usepackage{graphicx,wrapfig}
\usepackage[frozencache,cachedir=.]{minted}

\usepackage{tcolorbox}
\tcbuselibrary{minted,breakable,xparse,skins}

\definecolor{bg}{gray}{0.95}
\DeclareTCBListing{mintedbox}{O{}m!O{}}{%
  breakable=true,
  listing engine=minted,
  listing only,
  minted language=#2,
  minted style=default,
  minted options={%
    linenos,
    gobble=0,
    breaklines=true,
    breakafter=,,
    fontsize=\small,
    numbersep=8pt,
    #1},
  boxsep=0pt,
  left skip=0pt,
  right skip=0pt,
  left=25pt,
  right=0pt,
  top=3pt,
  bottom=3pt,
  arc=5pt,
  leftrule=0pt,
  rightrule=0pt,
  bottomrule=2pt,
  toprule=2pt,
  colback=bg,
  colframe=orange!70,
  enhanced,
  overlay={%
    \begin{tcbclipinterior}
    \fill[orange!20!white] (frame.south west) rectangle ([xshift=20pt]frame.north west);
    \end{tcbclipinterior}},
  #3}

\DeclareMathOperator*{\argmax}{arg\,max}

\newcommand{\vect}[1]{\boldsymbol{#1}}
\graphicspath{ {./images/} }

\definecolor{light-gray}{gray}{0.95}

\usepackage{hyperref}
\hypersetup{
    linkcolor=blue,
    filecolor=magenta,      
    urlcolor=cyan,
    pdftitle={Overleaf Example},
    pdfpagemode=FullScreen,
    }

\title{Training Spiking Neural Networks Using Lessons from Deep Learning}

\author{
\hspace{-5pt} Jason K. Eshraghian*\\
\hspace{-5pt}  UC Santa Cruz\\
\hspace{-5pt}  University of Michigan\\
\hspace{-5pt}  jeshragh@ucsc.edu\\
    \And
\hspace{40pt} Max Ward\\
\hspace{40pt} UWA\\ 
\hspace{40pt} Harvard University\\
  \And
\hspace{30pt} Emre Neftci\\
\hspace{30pt} Forschungszentrum J\"ulich\\
\hspace{30pt} RWTH Aachen\\
  \AND
\hspace{-20pt} Xinxin Wang\\
\hspace{-20pt}  University of Michigan\\
 \And
  \hspace{30pt} Gregor Lenz\\
\hspace{30pt} SynSense\\
 \And
\hspace{45pt} Girish Dwivedi \\
\hspace{45pt} UWA\\
 \AND
 \hspace{-15pt} Mohammed Bennamoun\\
  \hspace{-15pt}UWA\\
 \And
\hspace{30pt} Doo Seok Jeong\\
\hspace{30pt} Hanyang University\\
 \And
\hspace{35pt} Wei D. Lu*\\
\hspace{35pt}  University of Michigan\\
 \hspace{35pt} wluee@umich.edu\\
}
\begin{document}
\maketitle 
\begin{abstract}
The brain is the perfect place to look for inspiration to develop more efficient neural networks. The inner workings of our synapses and neurons provide a glimpse at what the future of deep learning might look like. This paper serves as a tutorial and perspective showing how to apply the lessons learnt from several decades of research in deep learning, gradient descent, backpropagation and neuroscience to biologically plausible spiking neural neural networks. 

We also explore the delicate interplay between encoding data as spikes and the learning process; the challenges and solutions of applying gradient-based learning to spiking neural networks (SNNs); the subtle link between temporal backpropagation and spike timing dependent plasticity, and how deep learning might move towards biologically plausible online learning.
Some ideas are well accepted and commonly used amongst the neuromorphic engineering community, while others are presented or justified for the first time here. 

The fields of deep learning and spiking neural networks evolve very rapidly. We endeavour to treat this document as a `dynamic' manuscript that will continue to be updated as the common practices in training SNNs also change.


A series of companion interactive tutorials complementary to this paper using our Python package, \textit{snnTorch}, are also made available.\footnote{Link to interactive tutorials: \url{https://snntorch.readthedocs.io/en/latest/tutorials/index.html}.}

\end{abstract}

\newpage
\tableofcontents


\newpage
\section{Introduction} \label{sec:1}
Deep learning has solved numerous problems in computer vision \cite{krizhevsky2012imagenet, girshick2014rich, girshick2015fast, ren2015faster, redmon2016you, eshraghian2020human}, speech recognition \cite{graves2014towards, chan2016listen, zhang2017very}, and natural language processing \cite{mikolov2013distributed, collobert2008unified, luong2015effective, vaswani2017attention, mikolov2010recurrent}. Neural networks have been instrumental in outperforming world champions in a diverse range of games, from Go to Starcraft \cite{silver2016mastering, vinyals2019grandmaster}. They are now surpassing the diagnostic capability of clinical specialists in numerous medical tasks \cite{mckinney2020international, hannun2019cardiologist, yang2021multimodal, rahimiazghadi2020hardware}. But for all the state-of-the-art models designed every day, a Kaggle \cite{kaggle} contest for state-of-the-art energy efficiency would go to the brain, every time. A new generation of brain-inspired spiking neural networks (SNNs) is poised to bridge this efficiency gap.

The amount of computational power required to run top performing deep learning models has increased at a rate of 10$\times$ per year from 2012 to 2019 \cite{thompson2020computational,perrault2019AI}. The rate of data generation is likewise increasing at an exponential rate. The backbone of OpenAI's ChatGPT language model, GPT-3, contains 175 billion learnable parameters, estimated to consume roughly 190,000~kWh to train \cite{brown2020language, dhar2020carbon, anthony2020carbontracker}.
Meanwhile, our brains operate within $\sim$12-20~W of power. This is in addition to churning through a multitude of sensory input, all the while ensuring our involuntary biological processes do not shut down \cite{levy2020computation}. If our brains dissipated as much heat as state-of-the-art deep learning models, then natural selection would have wiped humanity out long before we could have invented machine learning. To be fair, none of the authors can emulate the style of Shakespeare, or write up musical guitar tabs with the same artistic flair of GPT-4.

\begin{figure*}[!h]
    \centering
    \includegraphics[scale=0.6]{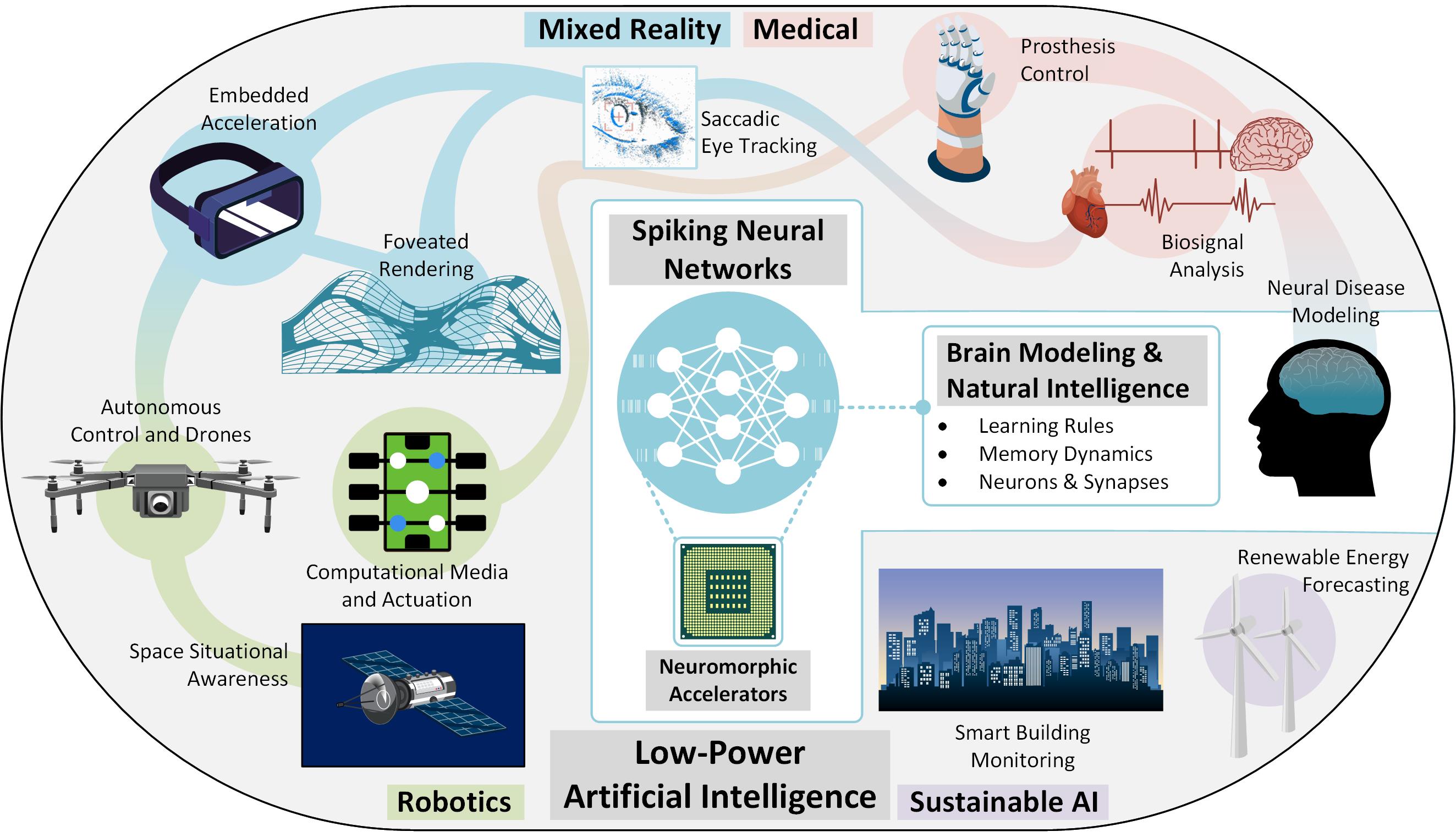}
    \caption{Spiking neural networks (SNNs) have pervaded many streams of deep learning that are in need of low-power, resource-constrained, and often portable operation. The utility of SNNs even extends to the modeling of neural dynamics across individual neurons and higher-level neural systems.}
    \label{fig:abstract}
\end{figure*}

\subsection{Neuromorphic Computing: A Quick Snapshot}
Neuromorphic (`brain-like') engineering strives to imitate the computational principles of the brain to drive down the energy cost of artificial intelligence systems. To replicate a biological system, we build on three parts:

\begin{enumerate}
    \item \textbf{Neuromorphic sensors} that take inspiration from biological sensors, such as the retina or cochlear, and typically record \textit{changes} in a signal instead of sampling it at regular intervals. Signals are only generated when a change occurs, and the signal is referred to as a `spike'.
    \item \textbf{Neuromorphic algorithms} that learn to make sense of spikes are known as spiking neural networks (SNNs). Instead of floating point values, SNNs work with single-bit, binary activations (spikes) that encode information over time, rather than in an intensity. As such, SNNs take advantage of low-precision parameters and high spatial and temporal sparsity.\footnote{A subtle caveat: it is possible for an SNN to accept non-spiking, continuous-valued input, and train the model to find the most efficient spike-based representation.}
    \item These models are designed with power-efficient execution on specialized \textbf{neuromorphic hardware} in mind. Sparse activations reduce data movement both on and off a chip to accelerate neuromorphic workloads, which can lead to large power and latency gains when compared to the same task on conventional hardware.
\end{enumerate}

Armed with these three components, neuromorphic systems are equipped to bridge the efficiency gap between today's and future intelligent systems.

What lessons can be learnt from the brain to build more efficient neural networks? Should we replicate the genetic makeup of a neuron right down to the molecular level~\cite{paun2005dna, qian2011neural}? Do we look at the way memory and processing coalesce within neurons and synapses~\cite{chi2016prime, rahimi2020complementary}? Or should we aim to extract the learning algorithms that underpin the brain~\cite{bi1998synaptic}?
This paper hones in on the intricacies of training brain-inspired neuromorphic algorithms, ultimately moving towards the goal of harnessing natural intelligence to further improve our use of artificial intelligence.
SNNs can already be optimized using the tools available to the deep learning community. But the brain-inspired nature of these emerging sensors, neuron models, and training methods are different enough to warrant a deep dive into biologically-inspired neural networks.

\subsection{Neuromorphic Systems in the Wild}
The overarching aim is to combine artificial neural networks (ANNs), which have already proven their worth in a broad range of domains, with the potential efficiency of SNNs~\cite{hamilton2021best}. 
So far, SNNs have staked their claim to a range of applications where power efficiency is of utmost importance.

\Cref{fig:abstract} offers a small window into the uses of SNNs, and their domain only continues to expand. Spiking algorithms have been used to implement low-power artificial intelligence algorithms across the medical, robotics, and mixed-reality domains, amongst many other fields. Given their power efficiency, initial commercial products often target Edge computing applications, close to where the data is recorded. 

In biosignal monitoring, nerve implants for brain-machine or biosignal interfaces have to pre-process information locally at minimum power and lack the bandwidths to transmit data for cloud computation. Work done in that direction using SNNs includes on-chip spike sorting~\cite{liu2016clockless, haessig2020mixed}, biosignal anomaly detection~\cite{bauer2019real, yan2021energy, yang2023neuromorphic, he2022implantable} and brain-machine interfaces~\cite{corradi2015neuromorphic, boi2016bidirectional}.
Beyond biomedical intervention, SNN models are also used in robotics in an effort to make them more human-like and to drive down the cost of operation~\cite{sandamirskaya2022neuromorphic, bartolozzi2022embodied, hussaini2022spiking}. Unmanned aerial vehicles must also operate in low-power environments to extract as much value from lightweight batteries, and have benefited from using neuromorphic processors \cite{dupeyroux2021neuromorphic}.
Audio signals can be processed with sub-mW power consumption and low latency on neuromorphic hardware as SNNs provide an efficient computational mechanism for temporal signal processing~\cite{bos2023sub}.

A plethora of efficient computer vision applications using spiking neural networks are reviewed in Ref.~\cite{gallego2020event}. SNNs are equally suitable to track objects such as satellites in the sky for space situational awareness~\cite{cohen2019event, afshar2020event}, scientific computing~\cite{kosters2023benchmarking}, and have been researched to promote sustainable uses of artificial intelligence, such as in monitoring material strain in smart-buildings \cite{henkes2022spiking} and wind power forecasting in remote areas that face power delivery challenges \cite{wei2021wind}. At the 2018-19 Telluride Neuromorphic and Cognition Workshops, a neuromorphic robot was even built to play foosball!\cite{cohen2022gooaall}

Beyond neuromorphic applications, SNNs are also used to test theories about how natural intelligence may arise, from the higher-level learning rules of the brain \cite{seung2003learning} and how memories are formed \cite{luboeinski2021memory}, down to lower-level neuronal and synaptic dynamics~\cite{perez2021neural}.

\subsection{Overview of Paper}
The brain's neural circuitry is a physical manifestation of its neural algorithm; understanding one will likely lead to an understanding of the other. This paper will hone in on one particular aspect of neural models: those that are compatible with modern deep learning. \Cref{fig:tax} provides an illustrated overview of the structure of this paper, and we will start from the ground up:

\begin{figure}
    \centering
    \includegraphics[scale=0.6]{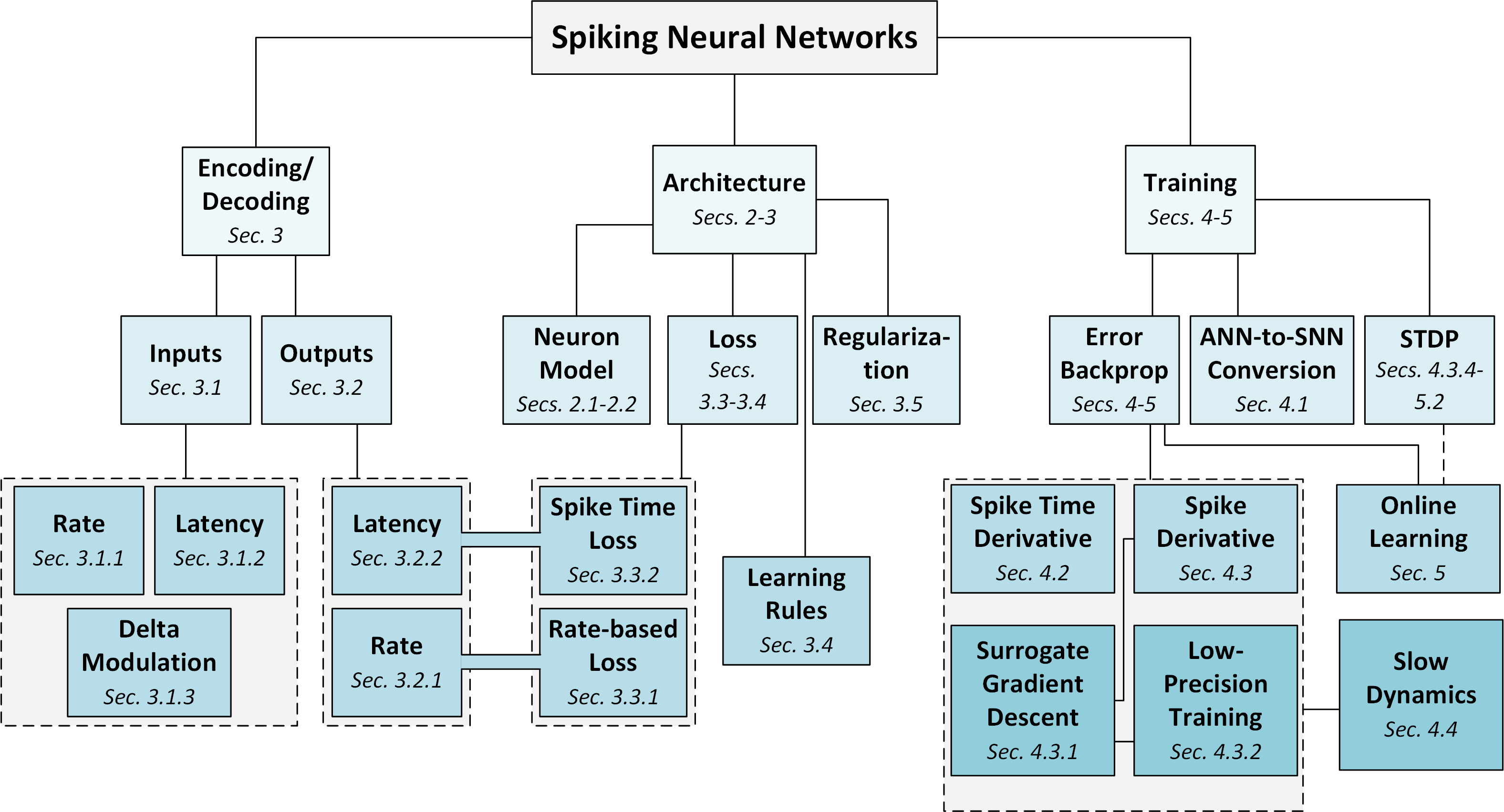}
    \caption{An overview of the paper structure.}
    \label{fig:tax}
\end{figure}

\begin{itemize}
    \item In \Cref{sec:2}, we will rationalise the commonly accepted advantages of using spikes, and derive a spiking neuron model from basic principles.
    \item These spikes will be assigned meaning in \Cref{sec:3} by exploring various spike encoding strategies, how they impact the learning process, and how objective and regularisation functions can be used to sway the spiking patterns of an SNN. 
    \item In \Cref{sec:4}, the challenges of training SNNs using gradient-based optimisation will be explored, and several solutions will be derived. These include defining derivatives at spike times and using approximations of the gradient.
    \item In doing so, a subtle link between the backpropagation algorithm and the spike-timing-dependent plasticity (STDP) learning rule will emerge, and be used in the subsequent section to derive online variants of backprop that move towards biologically plausible learning mechanisms (\Cref{fig:tax}).
\end{itemize}

 The aim is to combine artificial neural networks (ANNs), which have already proven their worth in a broad range of domains, with the potential efficiency of SNNs~\cite{hamilton2021best}. 


\newpage
\section{From Artificial to Spiking Neural Networks} \label{sec:2} 
\begin{figure}
    \centering
    \includegraphics[scale=0.32]{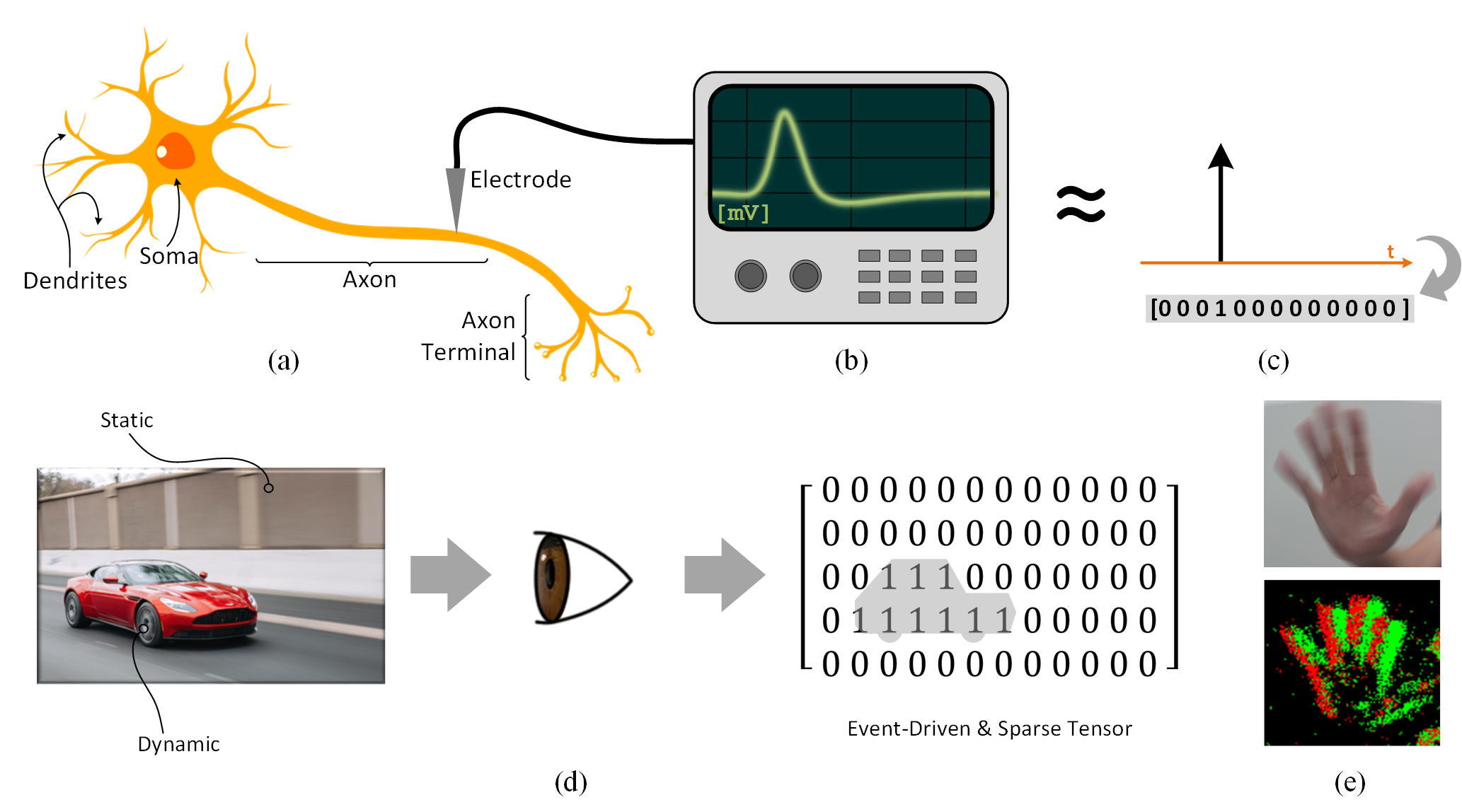}
    \caption{Neurons communicate via spikes. (a) Diagram of a neuron. (b) Measuring an action potential propagated along the axon of a neuron. Fluctuating subthreshold voltages are present in the soma, but become severely attenuated over distances beyond 1~mm \cite{dayan2001theoretical}. Only the action potential is detectable along the axon. (c) The neuron's spike is approximated with a binary representation. (d) Event-Driven Processing. Only dynamic segments of a scene are passed to the output (`1'), while static regions are suppressed (`0'). (e) Active Pixel Sensor and Dynamic Vision Sensor.}
    \label{fig:1}
\end{figure}
The neural code refers to how the brain represents information, and while many theories exist, the code is yet to be cracked. There are several persistent themes across these theories, which can be distilled down to \textit{`the three S's'}: spikes, sparsity, and static suppression. These traits are a good starting point to show \textit{why} the neural code might improve the efficiency of ANNs. Our first observation is:

\hfill\begin{minipage}{\dimexpr\textwidth-24pt}
1. \textbf{Spikes: } Biological neurons interact via \textit{spikes}
\xdef\tpd{\prevdepth}
\end{minipage}

Neurons primarily process and communicate with action potentials, or ``spikes'', which are electrical impulses of approximately 100~mV in amplitude. In most neurons, the occurrence of an action potential is far more important than the subtle variations of the action potential \cite{dayan2001theoretical}. Many computational models of neurons simplify the representation of a spike to a discrete, single-bit, all-or-nothing event (\Cref{fig:1}(a-c)). Communicating high-precision activations between layers, routing them around and between chips is an expensive undertaking. 
Multiplying a high-precision activation with a high-precision weight requires conversion into integers, decomposition of multiplication into multiple additions which introduces a carry propagation delay. On the other hand, a spike-based approach only requires a weight to be multiplied by a spike (`1'). This trades the cumbersome multiplication process with a simple memory read-out of the weight value.

Despite the activation being constrained to a single bit, spiking networks are vastly different from binarised neural networks. What actually matters is the \emph{timing} of the spike. Time is not a binarised quantity, and can be implemented using clock signals that are already distributed across a digital circuit. After all, why not use what is already available?

\hfill\begin{minipage}{\dimexpr\textwidth-24pt}
2. \textbf{Sparsity: } Biological neurons spend most of their time at rest, silencing a majority of activations to \textit{zero} at any given time
\xdef\tpd{\prevdepth}
\end{minipage}


Sparse tensors are cheap to store. The space that a simple data structure requires to store a matrix grows with the number of entries to store. In contrast, a data structure to store a sparse matrix only consumes memory with the number of non-zero elements. Take the following list as an example:

\begin{center}
     \fontfamily{courier}\selectfont
 \colorbox{light-gray}{[0, 0, 0, 0, 0, 0, 0, 0, 0, 0, 1, 0, 0, 0, 0, 0, 0, 0, 0, 0, 1]}
\end{center}

Since most of the entries are zero, we could save time by writing out only the non-zero elements as would occur in run-length encoding (indexing from zero):

\begin{center}
    \textit{``1 at position 10; 1 at position 20''}
\end{center}

For example, \Cref{fig:1}(c) shows how a single action potential can be represented by a sparsely populated vector. The sparser the list, the more space can be saved.

\hfill\begin{minipage}{\dimexpr\textwidth-24pt}
3. \textbf{Static Suppression (a.k.a., event-driven processing): } The sensory system is more responsive to changes than to static input
\xdef\tpd{\prevdepth}
\end{minipage}

The sensory periphery features several mechanisms that promote neuron excitability when subject to dynamic, changing stimuli, while suppressing its response to static, unchanging information. In retinal ganglion cells and the primary visual cortex, the spatiotemporal receptive fields of neurons promote excitable responses to regions of spatial contrast (or edges) over regions of spatial invariance \cite{hubel1962receptive}. Analogous mechanisms in early auditory processing include spectro-temporal receptive fields, which cause neurons to respond more favourably to changing frequencies in sound over static frequencies \cite{aertsen1981spectro}. These processes occur on short timescales (milliseconds), while perceptual adaptation has also been observed on longer timescales (seconds) \cite{benucci2013adaptation, wark2007sensory, eshraghian2018formulation}, causing neurons to become less responsive to prolonged exposure to fixed stimuli. 

\begin{figure}[!b]
    \centering
    \includegraphics[width=0.85\linewidth]{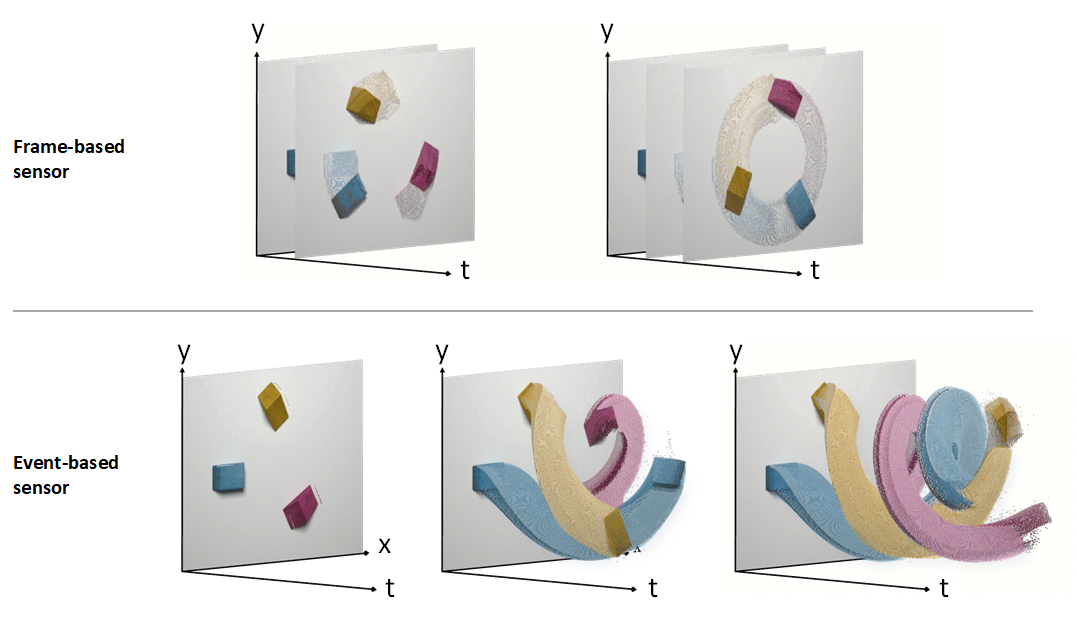}
    \caption{Functional difference between a conventional frame-based camera (above) and an event-based camera/silicon retina (below). The former records the scene as a sequence of images at a fixed frame rate. It operates independently of activity in the scene and can result in motion blur due to the global shutter. The silicon retina's output is directly driven by visual activity in the scene, as every pixel reacts to a change in illuminance.}
    \label{fig:event}
\end{figure}

A real-world engineering example of event-driven processing is 
the dynamic vision sensor (DVS), or the `silicon retina', which is a camera that reports changes in brightness and stays silent otherwise (\Cref{fig:1}(d-e)) \cite{ruedi2003128, lichtsteiner2006128, eshraghian2018neuromorphic, robey2021naturalizing, gallego2019event}. This also means that each pixel activates independently of all other pixels, as opposed to waiting for a global shutter to produce a still frame. The reduction of active pixels leads to huge energy savings when compared to conventional CMOS image sensors. This mix of low-power and asynchronous pixels allows for fast clock speeds, giving commercially available DVS cameras a microsecond temporal resolution without breaking a sweat \cite{brandli2014240}. The difference between a conventional frame-based camera and an event-based camera is illustrated in \Cref{fig:event}.

\subsection{Spiking Neurons}
\begin{figure}[!t]
    \centering
    \includegraphics[scale=0.7]{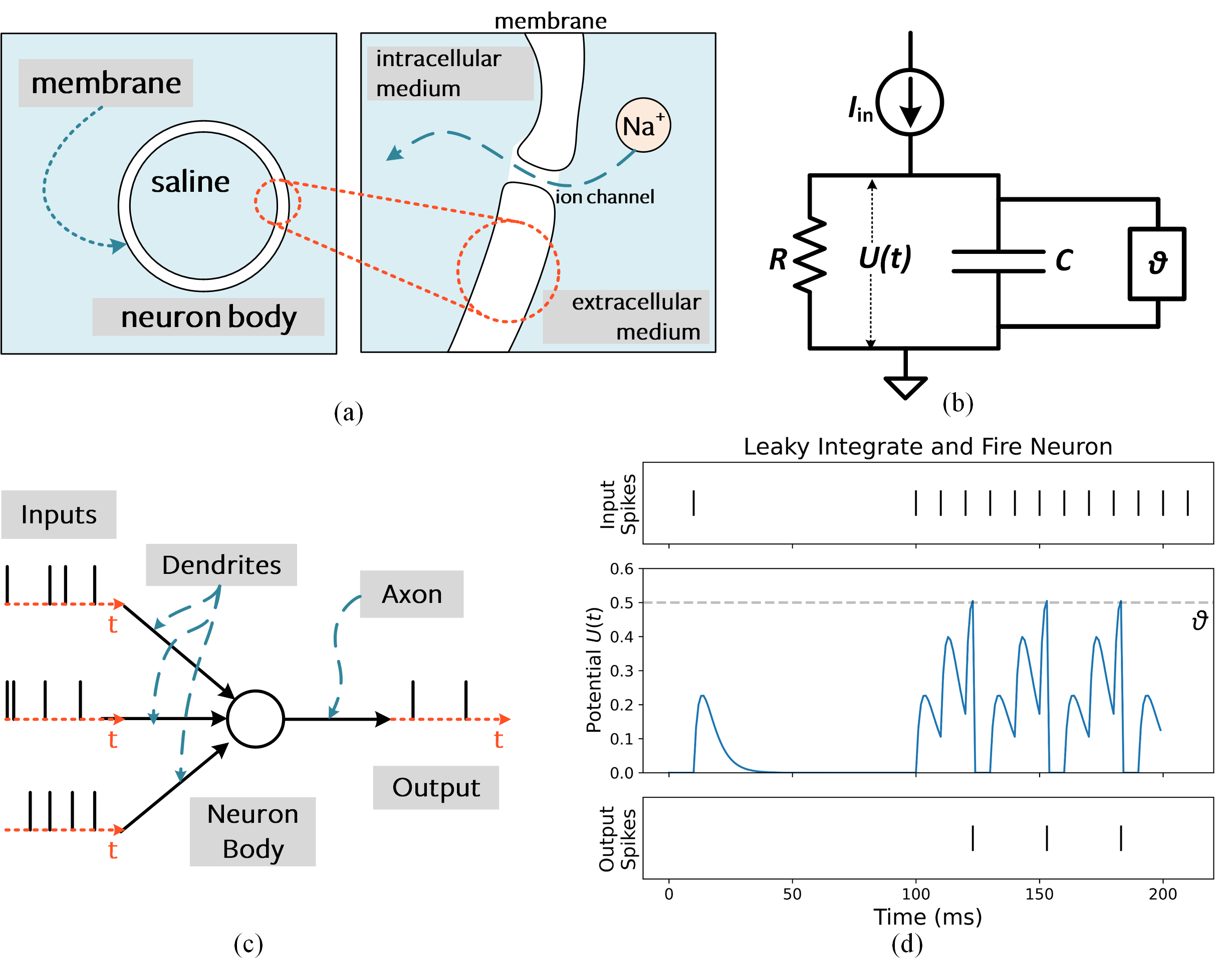}
    \caption{Leaky Integrate-and-Fire (LIF) Neuron Model. (a) An insulating bilipid membrane separates the intracellular and extracellular medium. Gated ion channels allow charge carriers, such as Na$^+$, to diffuse through the membrane. (b) The capacitive membrane and resistive ion channels form an RC circuit. When the membrane potential exceeds a threshold $\theta$, a spike is generated. (c) Input spikes generated by $I_{\rm in}$ are passed to the neuron body via the dendritic tree. Sufficient excitation will cause spike emission at the output. (d) A simulation depicting the membrane potential $U(t)$ reaching the threshold, arbitrarily set to $\theta=0.5 V$, which generates output spikes.}
    \label{fig:2}
\end{figure}

ANNs and SNNs can model the same types of network topologies, but SNNs trade the artificial neuron model with a spiking neuron model instead (\Cref{fig:2}). Much like the artificial neuron model \cite{rosenblatt1958perceptron}, spiking neurons operate on a weighted sum of inputs. Rather than passing the result through a sigmoid or ReLU nonlinearity, the weighted sum contributes to the membrane potential $U(t)$ of the neuron. If the neuron is sufficiently excited by this weighted sum, and the membrane potential reaches a threshold $\theta$, then the neuron will emit a spike to its subsequent connections. But most neuronal inputs are spikes of very short bursts of electrical activity. It is quite unlikely for all input spikes to arrive at the neuron body in unison (\Cref{fig:2}(c)). This indicates the presence of temporal dynamics that `sustain' the membrane potential over time.

These dynamics were quantified back in 1907 \cite{lapicque1907louis}. Louis Lapicque stimulated the nerve fiber of a frog leg using a hacked-together current source, and observed how long it took the frog leg to twitch based on the amplitude and duration of the driving current $I_{\rm in}$ \cite{brunel2007lapicque}. He concluded that a spiking neuron coarsely resembles a low-pass filter circuit consisting of a resistor $R$ and a capacitor $C$, later dubbed the leaky integrate-and-fire (LIF) neuron (\Cref{fig:2}(b)). This holds up a century later: physiologically, the capacitance arises from the insulating lipid bilayer forming the membrane of a neuron. The resistance arises from gated ion channels that open and close, modulating charge carrier diffusion across the membrane (\Cref{fig:2}(a--b)) \cite{hodgkin1952quantitative}. The dynamics of the passive membrane modeled using an RC circuit can be represented as:

\begin{equation}
\label{eq:1}
    \tau\frac{dU(t)}{dt} = -U(t) + I_{\rm in}(t)R
\end{equation}

where $\tau=RC$ is the time constant of the circuit. Typical values of $\tau$ fall on the order of 1-100 milliseconds. The solution of (\ref{eq:1}) for a constant current input is:

\begin{equation}\label{eq:generalsolution}
    U(t) = I_{\rm in}R + [U_0 - I_{\rm in}R]e^{-\frac{t}{\tau}}
\end{equation}

\noindent which shows how exponential relaxation of $U(t)$ to a steady state value follows current injection, where $U_0$ is the initial membrane potential at $t=0$. To make this time-varying solution compatible with a sequence-based neural network, the forward Euler method is used in the simplest case to find an approximate solution to \Cref{eq:1}:

\begin{equation}
\label{eq:euler}
    U[t]=\beta U[t-1] + (1-\beta)I_{\rm in}[t]
\end{equation}

where time is explicitly discretised, $\beta = e^{-1/\tau}$ is the decay rate (or `inverse time constant') of $U[t]$, and the full derivation is provided in \Cref{app:a1}. 

In deep learning, the weighting factor of an input is typically a learnable parameter. Relaxing the physically viable assumptions made thus far, the coefficient of input current in \Cref{eq:euler}, $(1-\beta)$, can be subsumed into a learnable weight $W$. The simplification $I_{\rm in}[t] = WX[t]$ is made to decouple the effect of $\beta$ on the input $X[t]$. Here, $X[t]$ is treated as a single input. A full-scale network would vectorise $X[t]$ and $W$ would be a matrix, but is treated here as a single input to a single neuron for simplicity. Finally, accounting for spiking and membrane potential reset gives:

\begin{equation}\label{eq:2}
    U[t] = \underbrace{\beta U[t-1]}_\text{\rm decay} + \underbrace{WX[t]}_\text{\rm input} -
    \underbrace{S_{\rm out}[t-1]\theta}_\text{\rm reset}
\end{equation}

$S_{\rm out}[t] \in \{0,1\}$ is the output spike generated by the neuron, where if activated ($S_{\rm out}=1$), the reset term subtracts the threshold $\theta$ from the membrane potential. Otherwise, the reset term has no effect ($S_{\rm out}=0$). 
A complete derivation of \Cref{eq:2} with all simplifying assumptions is provided in \Cref{app:a1}. A spike is generated if the membrane potential exceeds the threshold:

\begin{equation} \label{eq:3}
    S_{\rm out}[t] =
    \begin{cases}
      1,  & \text{\rm if $U[t] >\theta$} \\ 
      0, & \text{otherwise} \\
    \end{cases}  
\end{equation}

An example of how this might be coded in Python is as follows:

\begin{mintedbox}{python}

def lif(X, U):
    beta = 0.9 # set decay rate
    W = 0.5 # learnable parameter
    theta = 1 # set threshold
    S = 0 # initialize output spike
    
    U = beta * U + W*X - S * theta # iterate over one time step of Eq. 4
    S = (U > theta) # Eq. 5
    return S, U
\end{mintedbox}

In your exploration of LIF neurons, you may come across many slight variants. Some variations include:

\begin{itemize}
    \item The spike threshold (\textit{Line 8}) might be applied \textit{before} updating the membrane potential (\textit{Line 7}). This induces a one-step delay between the input signal $X$, and when it can trigger a spike.
    \item The above derivations use a `reset-by-subtraction` (or \textit{soft reset}) mechanism. But an alternative shown in \Cref{app:a1} is a `reset-to-zero' mechanism (or \textit{hard reset}).
    \item The factor $(1-\beta)$ from \Cref{eq:euler} may be included as a co-efficient to the input term, $WX$. This will allow you to simulate a neuron model with realistic time constants, but does not offer any advantages when ultimately applied to deep learning.
\end{itemize}

Alternatively to the above code-block, this can be continuously executed in snnTorch over discrete time steps:

\begin{mintedbox}{python}
import snntorch as snn

lif = snn.Leaky(beta=0.9, threshold=1) # initialize neuron

infinite_loop = True
while infinite_loop:
    S, U = lif(X*W, U) # Eq.4 and Eq. 5 are recurrently returned
\end{mintedbox}

An extensive list of alternative neuron types (both LIF and otherwise) is detailed in \Cref{sec:alt-neurons}, along with a brief overview of their use cases.

\begin{figure}[!h]
    \centering
    \includegraphics[scale=0.8]{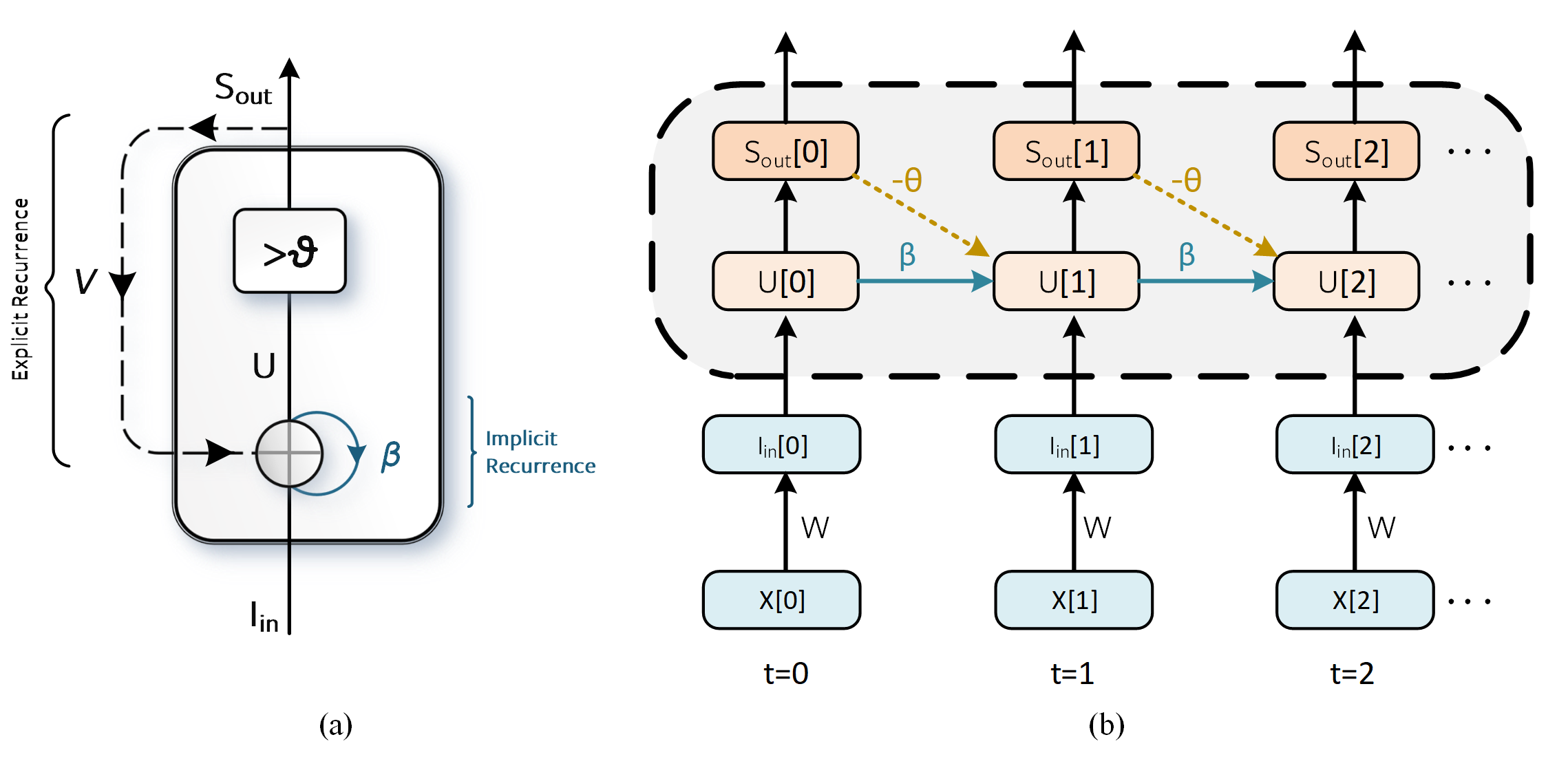}
    \caption{Computational steps in solving the LIF neuron model. (a) A recurrent representation of a spiking neuron. Hidden state decay is referred to as `implicit recurrence', and external feedback from the spike is `explicit recurrence', where $V$ is the recurrent weight (omitted from \Cref{eq:2})\cite{zenke2021brain}. (b) An unrolled computational graph of the neuron where time flows from left to right. $-\theta$ represents the reset term from \Cref{eq:2}, while $\beta$ is the decay rate of $U$ over time. Explicit recurrence is omitted for clarity. Note that kernel-based neuron models replace implicit recurrence with a time-varying filter \cite{gerstner2001framework, shrestha2018slayer, zhang2020spike}.}
    \label{fig:2.5}
\end{figure}

A graphical depiction of the LIF neuron is provided in \Cref{fig:2.5}. The recurrent neuron in (a) is `unrolled' across time steps in (b), where the reset mechanism is included via $S_{\rm out}\theta$, and explicit recurrence $S_{\rm out}V$ is omitted for brevity. The unrolled computational graph is a good way to think about leaky integrate-and-fire neurons. As you will soon see in \Cref{sec:4}, this unrolled representation allows us to borrow many of the tricks developed by deep learning researchers in training networks of LIF neurons.

\begin{tcolorbox}[colback=bg,colframe=black,title=Practical Note: Soft Reset Enables Better Performance]
  In our experience, neuron models that apply a variant of the soft reset to \Cref{eq:2} can lead to networks with higher performance (e.g., accuracy), shown in the expression below. That is, the reset mechanism in \Cref{eq:2} is scaled by the decay rate $\beta$. Why this improves performance is an open question.

  \begin{equation*}
      \underbrace{\beta S_{\rm out}[t-1] \theta}_{\rm reset}
  \end{equation*}
\end{tcolorbox}

\newpage
\subsection{Alternative Spiking Neuron Models}\label{sec:alt-neurons}
The LIF neuron is but one of many spiking neuron models. Some other models you might encounter are listed below:
\begin{itemize}
    \item \textbf{Integrate-and-Fire (IF):} The leakage mechanism is removed; $\beta=1$ in \Cref{eq:2}. 
    \item \textbf{Current-based:} Often referred to as CuBa neuron models, these incorporate synaptic conductance variation into leaky integrate and fire neurons. If the default LIF neuron is a first-order low-pass filter, then CuBa neurons are a second-order low-pass filter. The input spike train undergoes two rounds of `smoothing’, which means the membrane potential has a finite rise time rather than experiencing discontinuous jumps in response to incoming spikes~\cite{burkitt2006review, vogels2005signal, gerstner2014neuronal}. A depiction of such a neuron with a finite rise time of membrane potential is shown in Figure~\ref{fig:2}(d). 
    \item \textbf{Recurrent Neurons:} The output spikes of a neuron are routed back to the input, labelled in \Cref{fig:2.5}(a) with explicit recurrence. Rather than an alternative model, recurrence is a topology that can be applied to any other neuron, and can be implemented in different ways; i) one-to-one recurrence, where each neuron routes its own spike to itself, or ii) all-to-all recurrence, where the output spikes of a full layer are weighted and summed (e.g., via a dense or convolutional layer), before being fed back to the full layer~\cite{sun2022intelligence}.
    \item \textbf{Kernel-based Models:} Also knows as the spike-response model, where a pre-defined kernel (such as the `alpha function': see \Cref{app:c1}) is convolved with input spikes \cite{gerstner2001framework, shrestha2018slayer, zhang2020spike}. Having the option to define the kernel to be any shape offers significant flexibility.
    \item \textbf{Deep learning inspired spiking neurons:} Rather than drawing upon neuroscience, it is just as possible to start with primitives from deep learning and apply spiking thresholds. This helps with extending the short-term capacity of basic recurrent neurons. A couple of examples include spiking LSTMs \cite{yang2023neuromorphic}, and Legendre Memory Units \cite{voelker2019legendre}. More recently, transformers have been used to further improve long-range memory dependencies in data. In a similar manner, SpikeGPT approximated self-attention into a recurrent model, providing the first demonstration of natural language generation with SNNs~\cite{zhu2023spikegpt}.
    \item \textbf{Higher-complexity neuroscience-inspired models:} A large variety of more detailed neuron models are out there. These account for biophysical realism and/or morphological details that are not represented in simple leaky integrators. The most renowned models include the Hodgkin-Huxley model~\cite{hodgkin1952quantitative}, and the Izhikevich (or Resonator) model~\cite{izhikevich2003simple}, which can reproduce electrophysiological results with better accuracy.
\end{itemize}

The main takeaway is: use the neuron model that suits your task. Power efficient deep learning will call for LIF models. Improving performance may call for using recurrent SNNs. Driving performance even further (often at the expense of efficiency) may demand methods derived from deep learning, such as spiking LSTMs, and recurrent spiking transformers~\cite{zhu2023spikegpt, leroux2023online}. Or perhaps deep learning is not your goal. If you are aiming to construct a brain model, or are tasked with an exploration of linking low-level dynamics (ionic, conductance-driven, or otherwise) with higher-order brain function, then perhaps more detailed, biophysically accurate models will be your friend.
The following code-snippets show how some of these neurons can be instantiated in snnTorch.

\begin{mintedbox}{python}
import snntorch as snn

# all thresholds default to threshold=1
lif = snn.Leaky(beta=0.9) # vanilla leaky integrate-and-fire neuron
int_fire = snn.Leaky(beta=1.0) # integrate-and-fire neuron
cuba = snn.Synaptic(beta=0.9, alpha=0.8) # current-based neuron
rlif_1 = snn.RLeaky(beta=0.9, all_to_all=True) # all-to-all recurrent lif neuron
rlif_2 = snn.RLeaky(beta=0.9, all_to_all=False) # one-to-one recurrent lif neuron
slstm = snn.SLSTM(input_size=10, hidden_size=1000) # spiking LSTM: 10 inputs, 1000 outputs
\end{mintedbox}

With the neurons instantiated, it is just a matter of passing in two arguments: i) input data, and ii) their hidden state(s). The hidden states will be updated recursively.

\begin{mintedbox}{python}

import torch

X = torch.rand(10) # vector of 10 random inputs
U = torch.zeros(10) # initialize hidden states of 10 neurons to 0 V

infinite_loop = True
while infinite_loop:
    S, U = lif(X, U) # forward-pass of leaky integrate-and-fire neuron
\end{mintedbox}

    
Having formulated a spiking neuron in a discrete-time, recursive form, we can now `borrow' the developments in training RNNs and sequence-based models. This recursion is illustrated using an `implicit' recurrent connection for the decay of the membrane potential, and is distinguished from `explicit' recurrence where the output spikes $S_{\rm out}$ are fed back to the input as in recurrent SNNs (\Cref{fig:2.5}). 

While there are plenty more physiologically accurate neuron models \cite{hodgkin1952quantitative}, the leaky integrate and fire model is the most prevalent in gradient-based learning due to its computational efficiency and ease of training. Before moving onto training SNNs in \Cref{sec:4}, let us gain some insight to what spikes actually mean, and how they might represent information in the next section.

\begin{tcolorbox}[colback=bg,colframe=black,title=Practical Note: A Comparison of Various Neuron Models]
  With all of these neuron variants, what is the `correct' approach? This ultimately depends on your goals, and an actively researched question. In our experience:
  \begin{itemize}
      \item \textbf{LIF} neurons are the most commonly used in the deep learning context. They sit in the sweet spot between biological plausibility and practicality. It is relatively straightforward to adopt techniques used in training recurrent neural networks (RNNs) to SNNs, as will be explained in detail in \Cref{sec:4}.
      \item \textbf{IF} models are commonly used in low-power neuromorphic hardware (e.g., SynSense's Speck and Xylo systems; BrainChip's Akida). It removes the `multiply' step by $\beta$ when updating the membrane potential. This simplification is at the cost of treating all historical data in the same way. I.e., an event that occurred 10 seconds in the past will be weighted equally as an event that occurred 1 second in the past.
      \item \textbf{CuBa} neuron models add additional complexity without improving learning capacity. They tend not to be less common as a result. On the plus side, by smoothing out the membrane potential evolution (see \Cref{fig:2}(d)), the spike time becomes differentiable with respect to the membrane potential. More on this in \Cref{sec:4}.
      \item \textbf{Recurrent spiking neurons} have shown to improve learning convergence when training SNNs on time-varying tasks when using all-to-all connections~\cite{knight2023easy}. One-to-one connections may not appear to help on the face of it, but by applying a fixed, negative recurrent weight, it can be used as an alternative way to model an adaptive leaky integrate-and-fire neuron. For each output spike, rather than updating the threshold, the output spike can be used to inhibit the membrane potential instead.
      \item \textbf{Kernel-based Models} are, in theory, equally performant as all other first-order neuron models. They also demand more memory, as solving the state of the neuron at each time step depends on knowledge of the full history of spikes. In contrast, the formulation in \Cref{eq:2} only requires information from the previous time step. SLAYER~\cite{shrestha2018slayer} popularized their use in multi-layer deep learning, and its implementation was parallelized to improve training time in EXODUS~\cite{bauer2022exodus}.
      \item \textbf{Deep learning inspired neurons} deviate from biology and adopt models from deep learning instead. For example, LSTMs are known to have better memory capacity than basic neurons with recurrent connections. Adding a firing threshold to the state of the LSTM can reduce inter-layer spike traffic. However, some of these neurons may introduce higher-dimensional states which can be more expensive to compute. Where leaky integrate-and-fire neurons and recurrent SNNs are insufficient, deep learning-derived models may be a good alternative.
      \item \textbf{Higher-complexity neuroscience-inspired neurons} are less often used in deep learning as they involve solving highly stiff ordinary differential equations. I.e., a tiny perturbation may lead to negligible influence on the membrane potential, but the very same perturbation may skyrocket into an action potential. Harnessing gradients from such an unstable signal is not a pleasant experience. As a glimmer of hope, the neuromorphic folk at Intel Labs have been aiming to bridge the gap between low-level neuronal detail in Resonator neurons and higher-level functionality. They have recently discovered emergent properties of such models that allow for efficient signal processing (e.g., spectral decomposition) and factorization of high-dimensional vectors~\cite{frady2022efficient}.
    \end{itemize}
\end{tcolorbox}

\newpage 
\section{The Neural Code} \label{sec:3}

\begin{figure}[!b]
    \centering
    \includegraphics[scale=0.85]{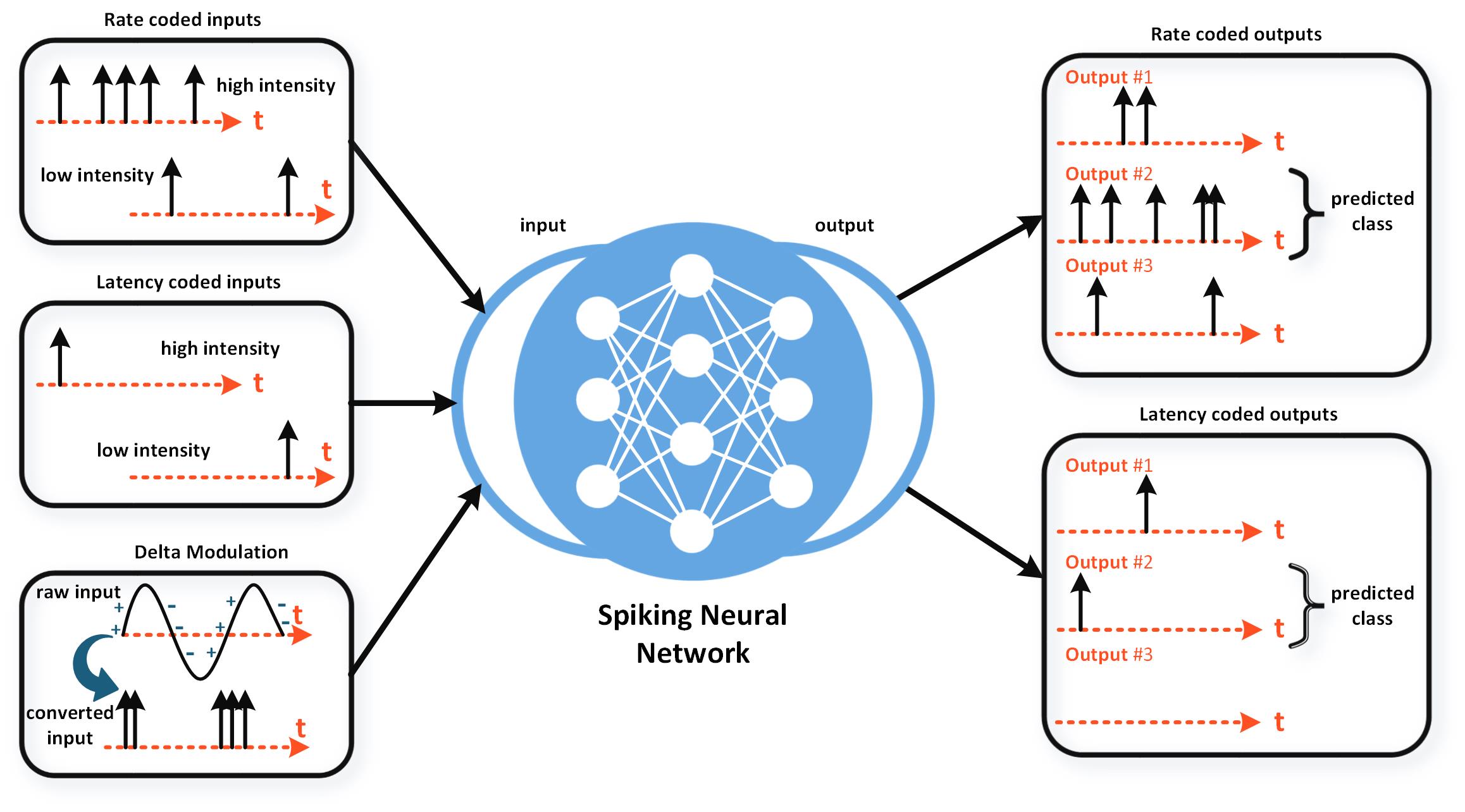}
    \caption{Input data to an SNN may be converted into a firing rate, firing time, or the data can be delta modulated. Alternatively, the input to the network can also be passed in without conversion which experimentally represents a direct or variable current source applied to the input layer of neurons. The network itself may be trained to enable the correct class to have the highest firing rate or to fire first, amongst many other encoding strategies.}
    \label{fig:3}
\end{figure}

Light is what we see when the retina converts photons into spikes. Odors are what we smell when the nose processes volatilised molecules into spikes. Tactile perceptions are what we feel when our nerve endings turn pressure into spikes. The brain trades in the global currency of the \textit{spike}. If all spikes are treated identically, then how do they carry meaning? With respect to spike encoding, there are two parts of a neural network that must be treated separately (\Cref{fig:3}):

\begin{enumerate}
    \item \textbf{Input encoding:} Conversion of input data into spikes which is then passed to a neural network
    \item \textbf{Output decoding:} Train the output of a network to spike in a way that is meaningful and informative
\end{enumerate}

\subsection{Input encoding}
Input data to an SNN does not necessarily have to be encoded into spikes. It is acceptable to pass continuous values as input, much like how the perception of light starts with a continuous number of photons impinging upon our photoreceptor cells.

Static data, such as an image, can be treated as a direct current (DC) input with the same features passed to the input layer of the SNN at every time step. But this does not exploit the way SNNs extract meaning from temporal data. In general, three encoding mechanisms have been popularised with respect to input data:

\begin{enumerate}
    \item \textbf{Rate coding} converts input intensity into a \textbf{firing rate} or \textbf{spike count}
    \item \textbf{Latency (or temporal) coding} converts input intensity to a spike \textbf{time}
    \item \textbf{Delta modulation} converts a temporal \textbf{change} of input intensity into spikes, and otherwise remains silent
\end{enumerate}

This is a non-exhaustive list, and these codes are not necessarily independent of each other.


\subsubsection{Rate Coded Inputs}
How does the sensory periphery encode information about the world into spikes? When bright light is incident upon our photoreceptor cells, the retina triggers a spike train to the visual cortex. Hubel and Wiesel's Nobel prize-winning research on visual processing indicates that a brighter input or a favourable orientation of light corresponds to a higher firing rate \cite{hubel1962receptive}. As a rudimentary example, a bright pixel is encoded into a high firing rate, whereas a dark pixel would result in low-frequency firing. Measuring the firing rate of a neuron can become quite nuanced. The simplest approach is to apply an input stimulus to a neuron, count up the total number of action potentials it generates, and divide that by the duration of the trial. Although straightforward, the problem here is that the dynamics of a neuron vary across time. There is no guarantee the firing rate at the start of the trial is anything near the rate at the end. 



An alternative method counts the spikes over a very short time interval $\Delta t$. For a small enough $\Delta t$, the spike count can be constrained to either 0 or 1, limiting the total number of possible outcomes to only two. By repeating this experiment multiple times, the average number of spikes (over trials) occurring within $\Delta t$ can be found. This average must be equal to or less than 1, interpreted as the observed probability that a neuron will fire within the brief time interval. To convert it into a \textit{time-dependent} firing rate, the trial average is divided by the duration of the interval. This probabilistic interpretation of the rate code can be distributed across multiple neurons, where counting up the spikes from a collection of neurons advocates for a population code \cite{mello2015scalable}.
This representation is quite convenient for sequential neural networks. Each discrete time step in an RNN can be thought of as lasting for a brief duration $\Delta t$ in which a spike either occurs or does not occur. A formal example of how this takes place is provided in \Cref{app:a2}.
Data can be rate-coded using the \textit{spikegen} module within snnTorch:

\begin{mintedbox}{python}
from snntorch import spikegen
import torch

steps = 100 # number of time steps

X = torch.rand(10) # vector of 10 random inputs
S = spikegen.rate(X, num_steps=steps) 

print(X.size())
>> torch.Size([10])
print(S.size())
>> torch.Size([100, 10])
\end{mintedbox}

\subsubsection{Latency Coded Inputs}
A latency, or temporal, code is concerned with the timing of a spike. The total number of spikes is no longer consequential. Rather, \textit{when} the spike occurs is what matters. For example, a time-to-first-spike mechanism encodes a bright pixel as an early spike, whereas a dark input will spike last, or simply never spike at all. When compared to the rate code, latency-encoding mechanisms assign much more meaning to each individual spike.

Neurons can respond to sensory stimuli over an enormous dynamic range. In the retina, neurons can detect individual photons to an influx of millions of photons \cite{hecht1942energy, van1946number, rieke1998single, eshraghian2020nonlinear, baek2020real}. To handle such widely varying stimuli, sensory transduction systems likely compress stimulus intensity with a logarithmic dependency. For this reason, a logarithmic relation between spike times and input feature intensity is ubiquitous in the literature (\Cref{app:a3}) \cite{dehaene2003neural, arrow2021prosthesis}.

Although sensory pathways appear to transmit rate coded spike trains to our brains, it is likely that temporal codes dominate the actual processing that goes on within the brain. More on this in \Cref{rvl}. 

\begin{mintedbox}{python}
from snntorch import spikegen
import torch

steps = 100 # number of time steps

X = torch.rand(10) # vector of 10 random inputs
S = spikegen.latency(X, num_steps=steps) 

print(X.size())
>> torch.Size([10])
print(S.size())
>> torch.Size([100, 10])
\end{mintedbox}

\subsubsection{Delta Modulated Inputs}

Delta modulation is based on the notion that neurons thrive on change, which underpins the operation of the silicon retina camera that only generates an input when there has been a sufficient change of input intensity over time. If there is no change in your field of view, then your photoreceptor cells are much less prone to firing. Computationally, this would take a time-series input and feed a thresholded matrix difference to the network. 
While the precise implementation may vary, a common approach requires the difference to be both \textit{positive} and \textit{greater} than some pre-defined threshold for a spike to be generated. This encoding technique is also referred to as `threshold crossing'. Alternatively, changes in intensity can be tracked over multiple time steps, and other approaches account for negative changes. For an illustration, see \Cref{fig:event}, where the `background' is not captured over time. Only the moving blocks are recorded, as it is those pixels that are changing.

Assuming the variable $X$ stores a batch of videos, say with 100 frames in each sample, delta modulation can be applied using the spikegen module:

\begin{mintedbox}{python}
from snntorch import spikegen
import torch

print(X.size())
>> torch.Size([100, 128, 1, 28, 28]) # time X batch-size X channel X x-dim X y-dim

S = spikegen.delta(X, num_steps=steps) # convert X to delta modulated spikes in S

print(S.size())
>> torch.Size([100, 128, 1, 28, 28]) # no change to the size; only to the elements
\end{mintedbox}


The previous techniques tend to `convert' data into spikes. But it is more efficient to natively capture data in `pre-encoded', spiking form. Each pixel in a DVS camera and channel in a silicon cochlear uses delta modulation to record changes in the visual or audio scene. Some examples of neuromorphic benchmark datasets are described in Table~\ref{tab:datasets}. A comprehensive series of neuromorphic-relevant datasets are accounted for in NeuroBench~\cite{yik2023neurobench}\footnote{https://neurobench.ai/}.

\begin{table}[htb]
    \centering
    \caption{Examples of neuromorphic datasets recorded with event-based cameras and cochlear models.}
    \resizebox{1.\columnwidth}{!}{
    \begin{tabular}{ll}
    \textbf{Vision datasets} & \\ \midrule
    ASL-DVS~\cite{bi2019graph} & 100,800 samples of American sign language recorded with DAVIS. \\
    DAVIS Dataset~\cite{mueggler2017event} & Includes spikes, frames and inertial measurement unit recordings of interior and outdoor scenes.\\
    DVS Gestures~\cite{amir2017low} & 11 different hand gestures recorded under 3 different lighting conditions. \\
    DVS Benchmark~\cite{hu2016dvs} & DVS benchmark datasets for object tracking, action recognition, and object recognition. \\
    MVSEC~\cite{zihao2018multi} & Spikes, frames and optical flow from stereo cameras for indoor and outdoor scenarios. \\
    N-MNIST~\cite{orchard2015converting} & Spiking version of the classic MNIST dataset by converting digits from a screen using saccadic motion. \\
    POKER DVS~\cite{serrano2015poker} & 4 classes of playing cards flipped in rapid succession in front of a DVS.
    \\
    DSEC~\cite{gehrig2021dsec} & A stereo event camera dataset for driving scenarios.\\
    \\
    \textbf{Audio datasets} & \\ \midrule
    N-TIDIGITS~\cite{anumula2018feature}  & Speech recordings from the TIDIGITS dataset converted to spikes with a silicon cochlear. \\
    SHD~\cite{cramer2020heidelberg}  & Spiking version of the Heidelberg Digits dataset converted using a simulated cochlear model.\\
    SSC~\cite{cramer2020heidelberg} & Spiking version of the \textit{Speech Commands} dataset converted using a simulated cochlear model. \\
    \end{tabular}
    }
    \label{tab:datasets}
\end{table}

\begin{tcolorbox}[colback=bg,colframe=black,title=Practical Note: Input Encoding]
What if you start off with a non-spiking dataset? Applying these input encoding mechanisms will almost always lead to accuracy/performance degradation. Information loss is guaranteed. It is often wiser to apply the pre-processing techniques that are prevalent in the literature; e.g., if you have EEG data, a Fourier transform is often appropriate. Alternatively, SpikeGPT introduced binarized embeddings that `learn' the optimal spike-based encoding of a sequence of input tokens~\cite{zhu2023spikegpt}. 
If you \textit{must} encode data into spikes, we have found that rate codes do less harm to accuracy/loss minimization than latency codes. In the ideal scenario, your sensor would naturally capture data as spikes rather than having to go through conversion and compression steps.
\end{tcolorbox}

\subsection{Output Decoding}
Encoding input data into spikes can be thought of as how the sensory periphery transmits signals to the brain. On the other side of the same coin, decoding these spikes provides insight on how the brain handles these encoded signals. In the context of training an SNN, the encoding mechanism does not constrain the decoding mechanism. Shifting our attention from the input of an SNN, how might we interpret the firing behavior of output neurons?

\begin{enumerate}
    \item \textbf{Rate coding} chooses the output neuron with the highest \textbf{firing rate}, or \textbf{spike count}, as the predicted class
    \item \textbf{Latency (or temporal) coding} chooses the output neuron that fires \textbf{first} as the predicted class
    \item \textbf{Population coding} applies the above coding schemes (typically a rate code) with \textbf{multiple neurons} per class
\end{enumerate}

\subsubsection{Rate Coded Outputs}\label{rco}
Consider a multi-class classification problem, where $N_C$ is the number of classes. A non-spiking neural network would select the neuron with the largest output activation as the predicted class. For a rate-coded spiking network, the neuron that fires with the highest frequency is used. As each neuron is simulated for the same number of time steps, simply choose the neuron with the highest spike count (\Cref{app:a4}).

\subsubsection{Latency Coded Outputs}
There are numerous ways a neuron might encode data in the timing of a spike. As in the case with latency-coded inputs, it could be that a neuron representing the correct class fires first. 
This addresses the energy burden that arises from the multiple spikes needed in rate codes. In hardware, the need for fewer spikes reduces the frequency of memory accesses which is another computational burden in deep learning accelerators. 

Biologically, does it make sense for neurons to operate on a time to first spike principle? How might we define `first' if our brains are not constantly resetting to some initial, default state? This is quite easy to address conceptually. The idea of a latency or temporal code is motivated by our response to a sudden input stimulus. For example, when viewing a static, unchanging visual scene, the retina undergoes rapid, yet subtle, saccadic motion. The scene projected onto the retina changes every few hundreds of milliseconds. It could very well be the case that the first spike must occur with respect to the reference signal generated by this saccade. 

\subsubsection{Rate vs. Latency Code} \label{rvl}
Whether neurons encode information as a rate, as latency, or as something wholly different, is a topic of much controversy. We do not seek to crack the neural code here, but instead aim to provide intuition on when SNNs might benefit from one code over the other.

\textbf{Advantages of Rate Codes}
\begin{itemize}
    \item \textbf{Error tolerance:} if a neuron fails to fire, there are ideally many more spikes to reduce the burden of this error.
    \item \textbf{More spiking promotes more learning:} additional spikes provide a stronger gradient signal for learning via error backpropagation. As will be described in \Cref{sec:4}, the absence of spiking can impede learning convergence (more commonly referred to as the `dead neuron problem').
\end{itemize}

\textbf{Advantages of Latency Codes}
\begin{itemize}
    \item \textbf{Power consumption:} generating and communicating fewer spikes means less dynamic power dissipation in tailored hardware. It also reduces memory access frequency due to sparsity, as a vector-matrix product for an all-zero input vector returns a zero output.
    \item \textbf{Speed:} the reaction time of a human is roughly in the ballpark of 250~ms. If the average firing rate of a neuron in the human brain is on the order of 10~Hz (which is likely an overestimation \cite{olshausen2006other}), then one can only process about 2-3 spikes in this reaction time window. In contrast, latency codes rely on a single spike to represent information. This issue with rate codes may be addressed by coupling it with a population code: if a single neuron is limited in its spike count within a brief time window, then just use more neurons \cite{mello2015scalable}. This comes at the expense of further exacerbating the power consumption problem of rate codes.
\end{itemize}

The power consumption benefit of latency codes is also supported by observations in biology, where nature optimises for efficiency. Olshausen and Field's work in `What is the other 85\% of V1 doing?' methodically demonstrates that rate-coding can only explain, at most, the activity of 15\% of neurons in the primary visual cortex (V1) \cite{olshausen2006other}. If our neurons indiscriminately defaulted to a rate code, this would consume an order of magnitude more energy than a temporal code. The mean firing rate of our cortical neurons must necessarily be rather low, which is supported by temporal codes.

Lesser explored encoding mechanisms in gradient-based SNNs include using spikes to represent a prediction or reconstruction error \cite{bengio2014auto}. The brain may be perceived as an anticipatory machine that takes action based on its predictions. When these predictions do not match reality, spikes are triggered to update the system. 

Some assert the true code must lie between rate and temporal codes \cite{shinomoto2007solution}, while others argue that the two may co-exist and only differ based on the timescale of observation: rates are observed for long timescales, latency for short timescales \cite{mehta2002role}. Some reject rate codes entirely \cite{brette2015philosophy}. This is one of those instances where a deep learning practitioner might be less concerned with what the brain does, and prefers to focus on what is most useful.


\subsection{Objective Functions}\label{sec:obj}
While it is unlikely that our brains use something as explicit as a cross-entropy loss function, it is fair to say that humans and animals have baseline objectives \cite{richards2019deep}. Biological variables, such as dopamine release, have been meaningfully related to objective functions from reinforcement learning \cite{schultz1997neural}. Predictive coding models often aim to minimise the information entropy of sensory encodings, such that the brain can actively predict incoming signals and inhibit what it already expects \cite{huang2011predictive}. The multi-faceted nature of the brain's function likely calls for the existence of multiple objectives \cite{marblestone2016toward}. How the brain can be optimised using these objectives remains a mystery, though we might be able to gain insight from multi-objective optimisation \cite{deb2014multi}.

A variety of loss functions can be used to encourage the output layer of a network to fire as a rate or temporal code. The optimal choice is largely unsettled, and tends to be a function of the network hyperparameters and complexity of the task at hand. All objective functions described below have successfully trained networks to competitive results on a variety of datasets, though come with their own trade-offs.

\subsubsection{Spike Rate Objective Functions} \label{sec:3sro}
A summary of approaches commonly adopted in supervised learning classification tasks with SNNs to promote the correct neuron class to fire with the highest frequency is provided in \Cref{tab:rateobj}. In general, either the cross-entropy loss or mean square error is applied to the spike count or the membrane potential of the output layer of neurons.

\begin{table*}[!ht] 
\centering \small
    \caption{Rate-coded objectives}
    \label{tab:rateobj}
    \begin{tabular}{  p{1.5cm}  p{6.85cm}  p{6.85cm} }
        \toprule
& \textbf{Cross-Entropy Loss}   
& \textbf{Mean Square Error} \\\midrule
\textbf{Spike Count}
& \textbf{Cross-Entropy Spike Rate:} The total number of spikes for each neuron in the output layer are accumulated over time into a spike count $\vec{c}\in \mathbb{N}^{N_C}$ (\Cref{beq:5} in \Cref{app:a4}), for $N_C$ classes. A multi-class categorical probability distribution is obtained by treating the spike counts as logits in the softmax function. Cross entropy minimisation is used to increase the spike count of the correct class, while suppressing the count of the incorrect classes \cite{amir2017low, esser2016convolutional} (\Cref{app:a5}).      
& \textbf{Mean Square Spike Rate:} The spike counts of both correct and incorrect classes are specified as targets. The mean square errors between the actual and target spike counts for all output classes are summed together. In practice, the target is typically represented as a proportion of the total number of time steps: e.g., the correct class should fire at 80\% of all time steps, while incorrect classes should fire 20\% of the time \cite{shrestha2018slayer, wu2018spatio, bellec2020solution, kaiser2020synaptic} (\Cref{app:a6}). \\\hline \rule{0pt}{2.5ex}\textbf{Membrane Potential}      
& \textbf{Maximum Membrane:} The logits are obtained by taking the maximum value of the membrane potential over time, which are then applied to a softmax cross entropy function. By encouraging the membrane potential of the correct class to increase, it is expected to encourage more regular spiking \cite{perez2021sparse, gutig2006tempotron, zenke2021remarkable}. A variant is to simply sum the membrane potential across all time steps to obtain the logits \cite{zenke2021remarkable} (\Cref{app:a7}).                       
& \textbf{Mean Square Membrane:} Each output neuron has a target membrane potential specified for each time step, and the losses are summed across both time and outputs. To implement a rate code, a superthreshold target should be assigned to the correct class across time steps (\Cref{app:a11}). \\ 

        \bottomrule
    \end{tabular}
\end{table*}

With a sufficient number of time steps, passing the spike count the objective function is more widely adopted as it operates directly on spikes. Membrane potential acts as a proxy for increasing the spike count, and is also not considered an observable variable which may partially offset the computational benefits of using spikes.

Cross-entropy approaches aim to suppress the spikes from incorrect classes, which may drive weights in a network to zero. This could cause neurons to go quiet in absence of additional regularisation. By using the mean square spike rate, which specifies a target number of spikes for each class, output neurons can be placed on the cusp of firing. Therefore, the network is expected to adapt to changing inputs with a faster response time than neurons that have their firing completely suppressed.

In networks that simulate a constrained number of time steps, a small change in weights is unlikely to cause a change in the spike count of the output. 
It might be preferable to apply the loss function directly to a more `continuous' signal, such as the membrane potential instead. This comes at the expense of operating on a full precision hidden state, rather than on spikes. 
Alternatively, using population coding can distribute the cost burden over multiple neurons to increase the probability that a weight update will alter the spiking behavior of the output layer. It also increases the number of pathways through which error backpropagation may take place, and improve the chance that a weight update will generate a change in the global loss.

All of these losses can be constructed in a single line each using the \textit{functional} module within snnTorch.

\begin{mintedbox}{python}
from snntorch import functional as SF

loss_1 = SF.ce_rate_loss() # cross-entropy spike rate
loss_2 = SF.mse_rate_loss() # mean square spike rate
loss_3 = SF.ce_max_membrane_loss() # maximum membrane
loss_4 = SF.mse_membrane_loss() # mean square membrane
\end{mintedbox}

\subsubsection{Spike Time Objectives} \label{sec:spiketimeobj}
Loss functions that implement spike timing objectives are less commonly used than rate-coded objectives. Two possible reasons may explain why: (1) error rates are typically perceived to be the most important metric in deep learning literature, and rate codes are more tolerant to noise, and (2) temporal codes are considerably more difficult to implement. A summary of approaches is provided in \Cref{tab:latencyobj}, with their snnTorch implementation below.

\begin{table*}[!ht] 
\centering \small
    \caption{Latency-coded objectives}
    \label{tab:latencyobj}
    \begin{tabular}{  p{1.5cm}  p{6.85cm}  p{6.85cm} }
        \toprule
& \textbf{Cross-Entropy Loss}   
& \textbf{Mean Square Error} \\\midrule
\textbf{Spike Time}
& \textbf{Cross-Entropy Spike Time:} The timing of the first spike of each neuron in the output layer is taken $\vec{f}\in\mathbb{R}^{N_C}$. As cross entropy minimisation involves maximising the likelihood of the correct class, a monotonically decreasing function must be applied to $\vec{f}$ such that early spike times are converted to large numerical values, while late spikes become comparatively smaller. These `inverted' values are then used as logits in the softmax function \cite{zhang2020spike} (\Cref{app:a8}). 

& \textbf{Mean Square Spike Time:} The spike time of all neurons are specified as targets. The mean square errors between the actual and target spike times of all output classes are summed together. This can be generalised to multiple spikes as well \cite{bohte2002error, shrestha2018slayer} (\Cref{app:a9}). \rule{6.765cm}{0.5pt} 
\rule{0pt}{2ex}\textbf{Mean Square Relative Spike Time:} A similar approach to above, but rather than specifying the precise time of each spike, only the relative time of the correct class must be specified. If the correct neuron fires a specified number of time steps earlier than incorrect neurons, the loss is fixed at zero \cite{kheradpisheh2020temporal} (\Cref{app:a10}).

\\\hline \rule{0pt}{2.5ex}\textbf{Membrane Potential}  

& Unreported in the literature.                       
& \textbf{Mean Square Membrane:} Analogous to the rate-coded case, each output neuron has a target membrane potential specified for each time step, and the losses are summed across both time and outputs. To implement a temporal code, the correct class should specify a target membrane greater than the threshold of the neuron at an early time (\Cref{app:a11}). \\ 

        \bottomrule
    \end{tabular}
\end{table*}

\begin{mintedbox}{python}
from snntorch import functional as SF

loss_1 = SF.ce_temporal_loss() # cross-entropy spike time
loss_2 = SF.mse_temporal_loss() # mean square spike time
loss_3 = SF.mse_membrane_loss() # mean square membrane - target must be latency-coded
\end{mintedbox}

The use cases of these objectives are analogous to the spike rate objectives. 
A subtle challenge with using spike times is that the default implementation assumes each neuron spikes at least once, which is not necessarily the case. This can be handled by forcing a spike at the final time step in the event a neuron does not fire \cite{kheradpisheh2020temporal}.


\begin{tcolorbox}[colback=bg,colframe=black,title=Practical Note: Output Decoding with a Read-Out Layer]
  Several state-of-the-art models wholly abandon spiking neurons at the output and train their models using a `read-out layer'. This often consists of an integrate-and-fire layer with infinitely high thresholds (i.e., they will never fire), or with typical artificial neurons that use standard activation functions (ReLU, sigmoid, softmax, etc.). While this often improves accuracy, this may not qualify as a fully spiking network. Does this actually matter? If one can still achieve power efficiency, then engineers will be happy and that's often all that matters.
\end{tcolorbox}

\subsection{Learning Rules} \label{sec:bml}

\subsubsection{Spatial and Temporal Credit Assignment}
Once a loss has been determined, it must somehow be used to update the network parameters with the hope that the network will iteratively improve at the trained task. Each weight takes some blame for its contribution to the total loss, and this is known as `credit assignment'. This can be split into the \textit{spatial} and \textit{temporal} credit assignment problems. Spatial credit assignment aims to find the spatial location of the weight contributing to the error, while the temporal credit assignment problem aims to find the time at which the weight contributes to the error. Backpropagation has proven to be an extremely robust way to address credit assignment, but the brain is far more constrained in developing solutions to these challenges.

Backpropagation solves spatial credit assignment by applying a distinct backward pass after a forward pass during the learning process \cite{guerguiev2017towards}. The backward pass mirrors the forward pass, such that the computational pathway of the forward pass must be recalled. In contrast, action potential propagation along an axon is considered to be unidirectional which may reject the plausibility of backprop taking place in the brain. Spatial credit assignment is not only concerned with calculating the weight's contribution to an error, but also assigning the error back to the weight. Even if the brain could somehow calculate the gradient (or an approximation), a major challenge would be projecting that gradient back to the synapse, and knowing which gradient belongs to which synapse.

This constraint of neurons acting as directed edges is increasingly being relaxed, which could be a mechanism by which errors are assigned to synapses~\cite{neftci2017event}. Numerous bi-directional, non-linear phenomena occur within individual neurons which may contribute towards helping errors find their way to the right synapse. 
For example, feedback connections are observed in most places where there are feedforward connections \cite{callaway2004feedforward}. 


\subsubsection{Biologically Motivated Learning Rules}
\begin{figure}[!t]
    \centering
    \includegraphics[scale=0.65]{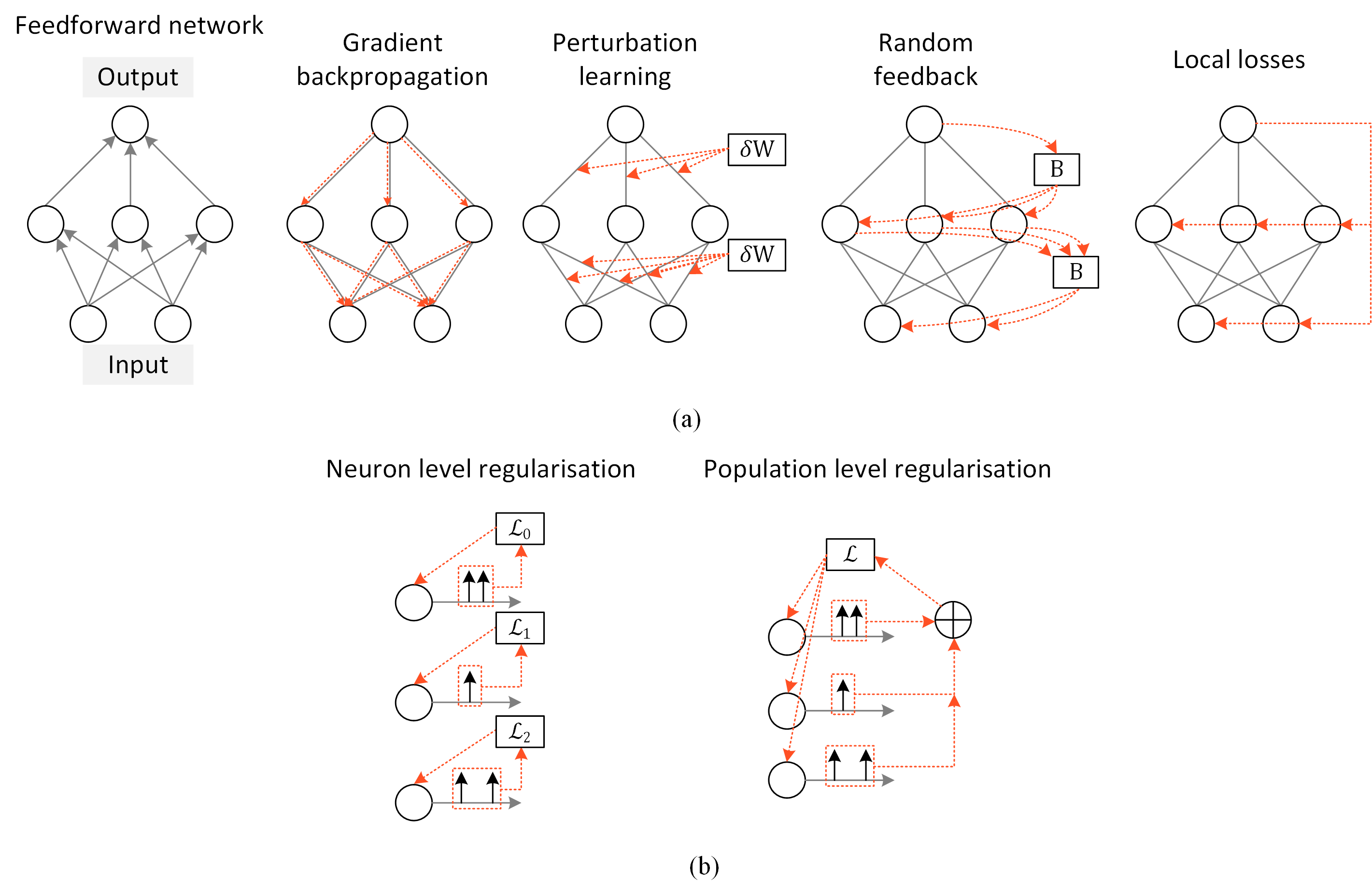}
    \caption{A variety of learning rules can be used to train a network. (a) Objective Functions. \textbf{Gradient backpropagation:} an unbiased gradient estimator of the loss is derived with respect to each weight. \textbf{Perturbation learning:} weights are randomly perturbed by $\delta W$, with the change accepted if the output error is reduced. \textbf{Random feedback:} all backward references to weights $W$ are replaced with random feedback $B$. \textbf{Local losses:} each layer is provided with an objective function avoiding error backpropagation through multiple layers. (b) Activity Regularisation. \textbf{Neuron level regularisation:} aims to set a baseline spike count per neuron. \textbf{Population level regularisation:} aims to set an upper limit on the total number of spikes emitted from all neurons.}
    \label{fig:bio_obj}
\end{figure}

With a plethora of neuronal dynamics that might embed variants of backpropagation, what options are there for modifying backprop to relax some of the challenges associated with biologically plausible spatial credit assignment? In general, the more broadly adopted approaches rely on either trading parts of the gradient calculation for stochasticity, or otherwise swapping a global error signal for localised errors (\Cref{fig:bio_obj}). Conjuring alternative methods to credit assignment that a real-time machine such as the brain can implement is not only useful for developing insight to biological learning\cite{lillicrap2020backpropagation}, but also reduces the cost of data communication in hardware \cite{laskin2020parallel}. For example, using local errors can reduce the length a signal must travel across a chip. Stochastic approaches can trade computation with naturally arising circuit noise \cite{lammie2021memristive, gaba2014memristive, cai2020power}. A brief summary of several common approaches to ameliorating the spatial credit assignment problem are provided below:


\begin{itemize}
    \item \textbf{Perturbation Learning:} A random perturbation of network weights is used to measure the change in error. If the error is reduced, the change is accepted. Otherwise, it is rejected \cite{williams1988toward, williams1992simple, werfel2005learning}. The difficulty of learning scales with the number of weights, where the effect of a single weight change is dominated by the noise from all other weight changes. In practice, it may take a huge number of trials to average this noise away \cite{seung2003learning}.
    \item \textbf{Random Feedback:} Backpropagation requires sequentially transporting the error signal through multiple layers, scaled by the forward weights of each layer. Random feedback replaces the forward weight matrices with random matrices, reducing the dependence of each weight update on distributed components of the network. While this does not fully solve the spatial credit assignment problem, it quells the \textit{weight transport problem} \cite{lillicrap2014random}, which is specifically concerned with a weight update in one layer depending upon the weights of far-away layers. Forward and backward-propagating data are scaled by symmetric weight matrices, a mechanism that is absent in the brain. Random feedback has shown similar performance to backpropagation on simple networks and tasks, which gives hope that a precise gradient may not be necessary for good performance \cite{lillicrap2014random}. Random feedback has struggled with more complex tasks, though variants have been proposed that reduce the gap \cite{moskovitz2018feedback, bartunov2018assessing, xiao2018biologically, frenkel2021learning}. Nonetheless, the mere fact that such a core piece of the backpropagation algorithm can be replaced with random noise and yet somehow still work is a marvel. It is indicative that we still have much left to understand about gradient backpropagation.
    
    \item \textbf{Local Losses:} It could be that the six layers of the cortex are each supplied with their own cost function, rather than a global signal that governs a unified goal for the brain \cite{marblestone2016toward}. Early visual regions may try to minimise the prediction error in constituent visual features, such as orientations, while higher areas use cost functions that target abstractions and concepts. For example, a baby learns how to interpret receptive fields before consolidating them into facial recognition. In deep learning, greedy layer-wise training assigns a cost function to each layer independently \cite{bengio2007greedy}. Each layer is sequentially assigned a cost function so as to ensure a shallow network is only ever trained. Target propagation is similarly motivated, by assigning a reconstruction criterion to each layer \cite{bengio2014auto}. Such approaches exploit the fact that training a shallow network is easier than training a deep one, and aim to address spatial credit assignment by ensuring the error signal does not need to propagate too far \cite{mostafa2018deep, neftci2017event}.

    \item \textbf{Forward-Forward Error Propagation:} The backward pass of a model is replaced with a second forward-pass where the input signal is altered based on error, or some related metric. Initially proposed by Dellaferrera \textit{et al.}~\cite{dellaferrera2022error}, Hinton's Forward-Forward learning algorithm generated more traction soon after~\cite{hinton2022forward}. These have not been ported to SNNs at the time of writing, though someone is bound to step up to the mantle soon.
\end{itemize}

These approaches to learning are illustrated in \Cref{fig:bio_obj}(a). While they are described in the context of supervised learning, many theories of learning place emphasis on self-organisation and unsupervised approaches. Hebbian plasticity is a prominent example \cite{hebb1949organisation}. But an intersection may exist in self-supervised learning, where the target of the network is a direct function of the data itself. Some types of neurons may be representative of facts, features, or concepts, only firing when exposed to the right type of stimuli. Other neurons may fire with the purpose of reducing a reconstruction error \cite{kushnir2019learning, deneve2017brain, yang2022weak}. By accounting for spatial and temporal correlations that naturally exist around us, such neurons may fire with the intent to predict what happens next. A more rigorous treatment of biological plausibility in objective functions can be found in \cite{marblestone2016toward}.

\begin{tcolorbox}[colback=bg,colframe=black,title=Practical Note: The Effectiveness of Credit Assignment Solutions]
  Gradient-based error backpropagation reigns supreme where final accuracy/performance is the key metric. Perturbation learning is effective with an infinitely large number of trials which is not practical, and the other methods struggle to scale to deep layers. At the same time, the brain is not considered `deep', so perhaps the focus should not be on how to make these algorithms scale to deep models, but rather, how shallow networks can perform to the same effectiveness as deeper models. This may call for evolutionary algorithms, where it has been shown that smaller networks can be optimized using a fitness function to outperform larger networks~\cite{parsa2021multi, schuman2020evolutionary}.
\end{tcolorbox}


\subsection{Activity Regularisation}
A huge motivator behind using SNNs comes from the power efficiency when processed on appropriately tailored hardware. This benefit is not only from single-bit inter-layer communication via spikes, but also the sparse occurrence of spikes. Some of the loss functions above, in particular those that promote rate codes, will indiscriminately increase the membrane potential and/or firing frequency without an upper bound, if left unchecked. Regularisation of the loss can be used to penalise excessive spiking (or alternatively, penalise insufficient spiking which is great for discouraging dead neurons). Conventionally, regularisation is used to constrain the solution space of loss minimisation, thus leading to a reduction in variance at the cost of increasing bias. Care must be taken, as too much activity regularisation can lead to excessively high bias. Activity regularisation can be applied to alter the behavior of individual neurons or populations of neurons, as depicted in \Cref{fig:bio_obj}(b).

\begin{itemize}
    \item \textbf{Population level regularisation:} this is useful when the metric to optimise is a function of aggregate behavior. For example, the metric may be power efficiency which is strongly linked to the total number of spikes from an entire network. L1-regularisation can be applied to the total number of spikes emitted at the output layer to penalise excessive firing, which encourages sparse activity at the output \cite{zenke2019spytorch}. Alternatively, for more fine-grain control over the network, an upper-activity threshold can be applied. If the total number of spikes for \textit{all} neurons in a layer exceeds the threshold, only then does the regularisation penalty kick in \cite{zenke2021remarkable, kaiser2020synaptic} (\Cref{app:regpop}).
    \item \textbf{Neuron level regularisation:} If neurons completely cease to fire, then learning may become significantly more difficult. Regularisation may also be applied at the individual neuron level by adding a penalty for each neuron. A lower-activity threshold specifies the lower permissible limit of firing for \textit{each} neuron before the regularisation penalty is applied (\Cref{app:regneu}).
\end{itemize}

Recent experiments have shown that rate-coded networks (at the output) are robust to sparsity-promoting regularisation terms \cite{zenke2021remarkable, kaiser2020synaptic, perez2021sparse}. However, networks that rely on time-to-first-spike schemes have had less success, which is unsurprising given that temporal outputs are already sparse.

Encouraging each neuron to have a baseline spike count helps with the backpropagation of errors through pathways that would otherwise be inactive. Together, the upper and lower-limit regularisation terms can be used to find the sweet spot of firing activity at each layer. As explained in detail in \cite{he2015delving}, the variance of activations should be as close as possible to `1' to avoid vanishing and exploding gradients. While modern deep learning practices rely on appropriate parameter initialization to achieve this, these approaches were not designed for non-differentiable activation functions, such as spikes. By monitoring and appropriately compensating for neuron activity, this may turn out to be a key ingredient to successfully training deep SNNs.

\begin{tcolorbox}[colback=bg,colframe=black,title=Practical Note: Avoiding Dead Neurons by Varying Thresholds]
  One of the most common reasons your network will not train is that neurons are not firing. Probing the spiking activity to identify the layer at which activity dies out is important, and can be used to guide how to regularise your objective function. An easier approach is to simply reduce the threshold of the at-risk layers. We do this threshold hack all the time!
\end{tcolorbox}

\newpage
\section{Training Spiking Neural Networks}\label{sec:4}
The rich temporal dynamics of SNNs give rise to a variety of ways in which a neuron's firing pattern can be interpreted. Naturally, this means there are several methods for training SNNs. They can generally be classified into the following methods:

\begin{itemize}
    \item \textbf{Shadow training:} A non-spiking ANN is trained and converted into an SNN by interpreting the activations as a firing rate or spike time
    \item \textbf{Backpropagation using spikes:} The SNN is natively trained using error backpropagation, typically through time as is done with sequential models
    \item \textbf{Local learning rules:} Weight updates are a function of signals that are spatially and temporally local to the weight, rather than from a global signal as in error backpropagation
\end{itemize}

Each approach has a time and place where it outshines the others. We will focus on approaches that apply backprop directly to an SNN, but useful insights can be attained by exploring shadow training and various local learning rules.

The goal of the backpropagation algorithm is loss minimisation. To achieve this, the gradient of the loss is computed with respect to each learnable parameter by applying the chain rule from the final layer back to each weight \cite{rumelhart1986learning, linnainmaa1970representation, werbos1982applications}. The gradient is then used to update the weights such that the error is ideally always decreased. If this gradient is `0', there is no weight update. This has been one of the main road blocks to training SNNs using error backpropagation due to the non-differentiability of spikes. This is also known as the dreaded `dead neuron' problem. There is a subtle, but important, difference between `vanishing gradients' and `dead neurons' which will be explained in \Cref{sec:bps}.

\begin{figure}[!t]
    \centering
    \includegraphics[scale=0.85]{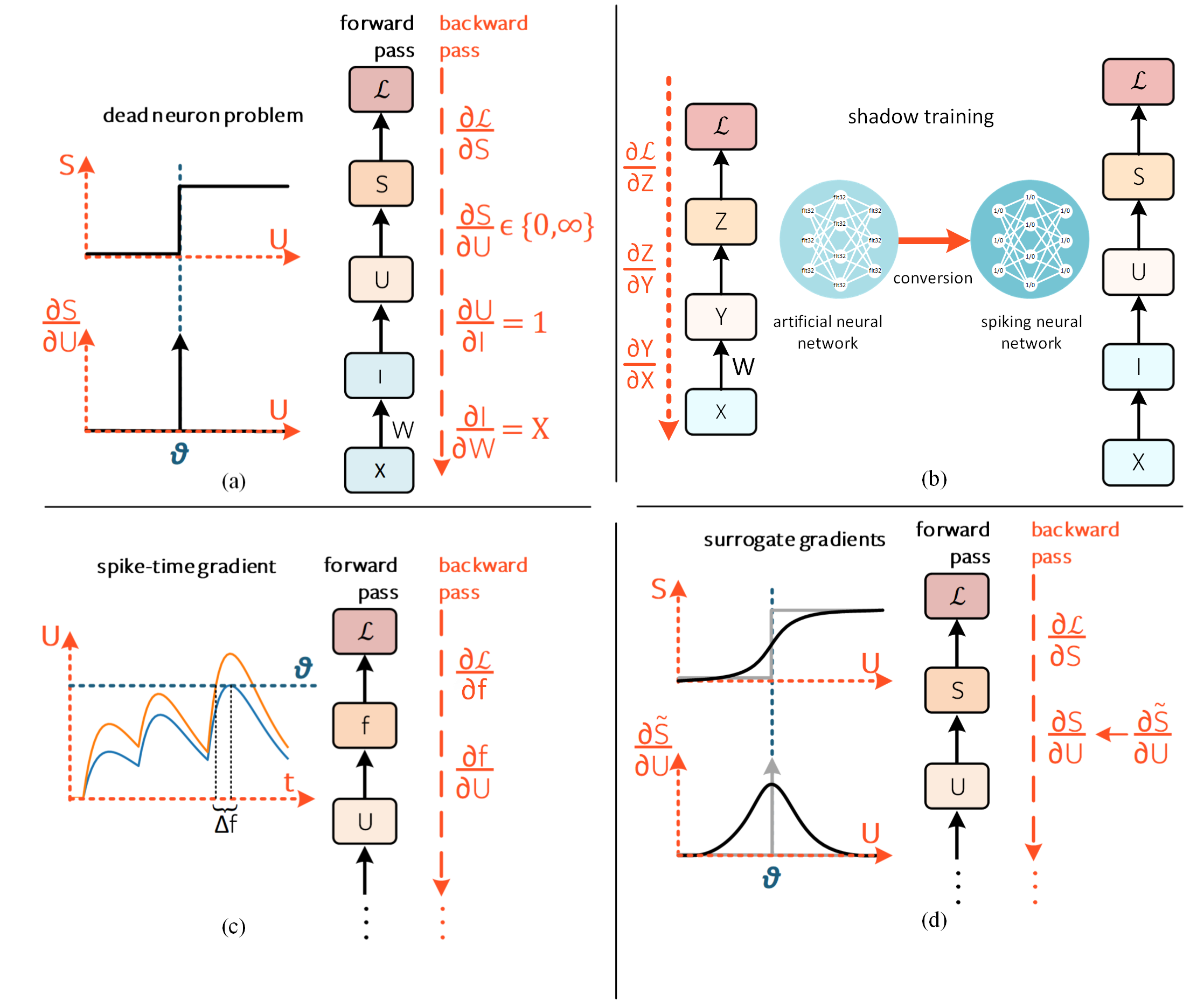}
    \caption{Addressing the dead neuron problem. Only one time step is shown, where temporal connections and subscripts from \Cref{fig:2.5} have been omitted for simplicity. (a) The dead neuron problem: the analytical solution of $\partial S/\partial U \in \{0, \infty\}$ results in a gradient that does not enable learning. (b) Shadow training: a non-spiking network is first trained and subsequently converted into an SNN. (c) Spike-time gradient: the gradient of spike time $f$ is taken instead of the gradient of the spike generation mechanism, which is a continuous function as long as a spike necessarily occurs \cite{bohte2002error}. (d) Surrogate gradients: the spike generation function is approximated to a continuous function during the backward pass \cite{zenke2021remarkable}. The left arrow ($\leftarrow$) indicates function substitution. This is the most broadly adopted solution to the dead neuron problem.} 
    \label{fig:5}
\end{figure}

To gain deeper insight behind the non-differentiability of spikes, recall the discretised solution of the membrane potential of the leaky integrate and fire neuron from \Cref{eq:2}: $U[t] = \beta U[t-1] + WX[t]$, where the first term represents the decay of the membrane potential $U$, and the second term is the weighted input $WX$. The reset term and subscripts have been omitted for simplicity. Now imagine a weight update $\Delta W$ is applied to the weight $W$ (\Cref{eq:2}). This update causes the membrane potential to change by $\Delta U$, but this change in potential fails to precipitate a further change to the spiking presence of the neuron (\Cref{eq:3}). That is to say, $dS/dU=0$ for all $U$, other than the threshold $\theta$, where $dS/dU\rightarrow\infty$. This drives the term we are actually interested in, $d\mathcal{L}/dW$, or the gradient of the loss in weight space, to either `0' or `$\infty$'. In either case, there is no adequate learning signal when backpropagating through a spiking neuron (\Cref{fig:5}(a)).

\subsection{Shadow Training} \label{sec:shadow}
The dead neuron problem can be completely circumvented by instead training on a shadow ANN and converting it into an SNN (\Cref{fig:5}(b)). The high precision activation function of each neuron is converted into either a spike rate \cite{perez2013mapping, hunsberger2015spiking, diehl2016conversion, hu2018spiking, rueckauer2017conversion} or a latency code \cite{stockl2021optimized}. One of the most compelling reasons to use shadow training is that advances in conventional deep learning can be directly applied to SNNs. For this reason, ANN-to-SNN conversion currently takes the crown for static image classification tasks on complex datasets, such as CIFAR-10 and ImageNet. Where inference efficiency is more important than training efficiency, and if input data is not time-varying, then shadow training could be the optimal way to go.

In addition to the inefficient training process, there are several drawbacks. Firstly, the types of tasks that are most commonly benchmarked do not make use of the temporal dynamics of SNNs, and the conversion of sequential neural networks to SNNs is an under-explored area \cite{diehl2016conversion}. Secondly, converting high-precision activations into spikes typically requires a long number of simulation time steps which may offset the power/latency benefits initially sought from SNNs. But what really motivates doing away with ANNs is that the conversion process is necessarily an approximation. Therefore, a shadow-trained SNN is very unlikely to reach the performance of the original network.

The issue of long time sequences can be partially addressed by using a hybrid approach: start with a shadow-trained SNN, and then perform backpropagation on the converted SNN \cite{rathi2019enabling}. Although this appears to degrade accuracy (reported on CIFAR-10 and ImageNet), it is possible to reduce the required number of steps by an order of magnitude. A more rigorous treatment of shadow training techniques and challenges can be found in \cite{pfeiffer2018deep}.

\subsection{Backpropagation Using Spike Times} \label{sec:4bpst}
An alternative method to side step the dead neuron problem is to instead take the derivative at spike times. In fact, this was the first proposed method to training multi-layer SNNs using backpropagation \cite{bohte2002error}. The original approach in \textit{SpikeProp} observes that while spikes may be discontinuous, time is continuous. Therefore, taking the derivative of spike \textit{timing} with respect to the weights achieves functional results. A thorough description is provided in \Cref{app:c1}.

\begin{tcolorbox}[colback=bg,colframe=black,title=Practical Note: Gradients at Spike Times]
This approach continues to be widely researched, but has taken a back seat to backpropagation through time using surrogate gradient descent, as it does not perform as well at optimizing a loss function. Why is this? Our best guess is the lower performance occurs because there are fewer edges to backpropagate errors through, making credit assignment more challenging. More details on surrogate gradient descent in the following section.
\end{tcolorbox}

Intuitively, SpikeProp calculates the gradient of the error with respect to the spike time. A change to  the weight by $\Delta W$ causes a change of the membrane potential by $\Delta U$, which ultimately results in a change of spike timing by $\Delta f$, where $f$ is the firing time of the neuron. In essence, the non-differentiable term $\partial S/ \partial U$ has been traded with $\partial f/\partial U$. This also means that each neuron \textit{must} emit a spike for a gradient to be calculable. This approach is illustrated in \Cref{fig:5}(c). Extensions of SpikeProp have made it compatible with multiple spikes \cite{booij2005gradient}, which are highly performant on data-driven tasks some of which have surpassed human level performance on MNIST and N-MNIST \cite{zhang2020spike, xu2013supervised, jin2023bplc+, wunderlich2021event}.

Several drawbacks arise. Once neurons become inactive, their weights become frozen. In most instances, no closed-form solutions exist to solving for the gradient if there is no spiking \cite{comsa2020temporal}. SpikeProp tackles this by modifying parameter initialization (i.e., increasing weights until a spike is triggered). But since the inception of SpikeProp in 2002, the deep learning community's understanding of weight initialization has gradually matured. We now know initialization aims to set a constant activation variance between layers, the absence of which can lead to vanishing and exploding gradients through space and time \cite{he2015delving, glorot2010understanding}. Modifying weights to promote spiking may detract from this. Instead, a more effective way to overcome the lack of firing is to lower the firing thresholds of the neurons. One may consider applying activity regularization to encourage firing in hidden layers, though this has degraded classification accuracy when taking the derivative at spike times. This result is unsurprising, as regularization can only be applied at the spike time rather than when the neuron is quiet. 

Another challenge is that it enforces stringent priors upon the network (e.g., each neuron must fire only once) that are incompatible with dynamically changing input data. This may be addressed by using periodic temporal codes that refresh at given intervals, in a similar manner to how visual saccades may set a reference time. But it is the only approach that enables the calculation of an unbiased gradient without any approximations in multi-layer SNNs. Whether this precision is necessary is a matter of further exploration on a broader range of tasks. 

\subsection{Backpropagation Using Spikes}\label{sec:bps}

\begin{figure}[!t]
    \centering
    \includegraphics[scale=0.8]{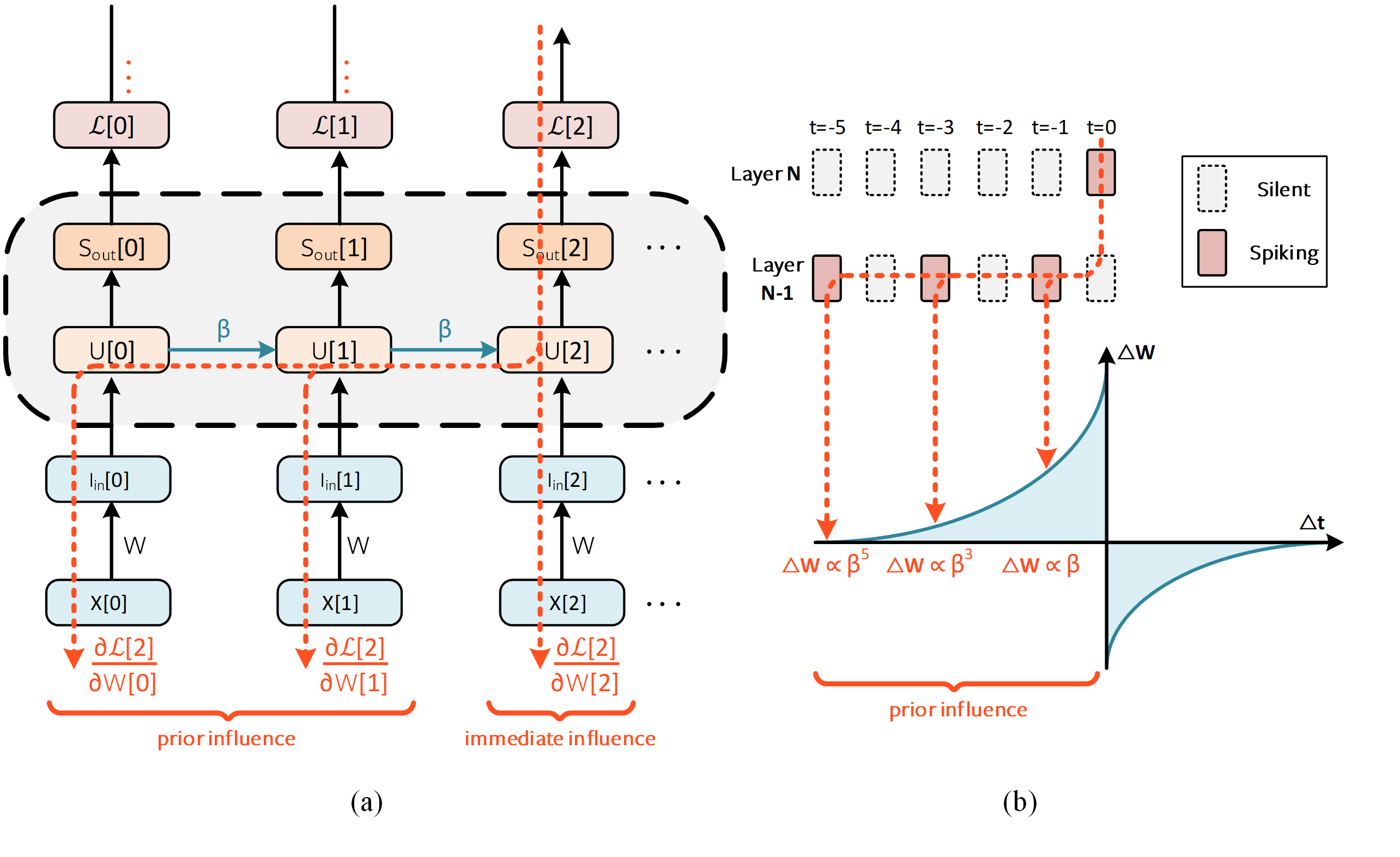}
    \caption{Backpropagation through time. (a) The present time application of $W$ is referred to as the immediate influence, with historical application of $W$ described as the prior influence. Reset dynamics and explicit recurrence have been omitted for brevity. The error pathways through $\mathcal{L}[0]$ and $\mathcal{L}[1]$ are also hidden but follow the same idea as that of $\mathcal{L}[2]$. (b) The hybrid approach defaults to a non-zero gradient only at spike times. For present time $t=0$, the derivative of each application of $W[s]$ with respect to the loss decays exponentially moving back in time. The magnitude of the weight update $\Delta W$ for prior influences of $W[s]$ follows a relationship qualitatively resembling that of STDP learning curves, where the strength of the synaptic update is dependent on the order and firing time of a pair of connected neurons \cite{bi1998synaptic}. }
    \label{fig:bptt}
\end{figure}

Instead of computing the gradient with respect to spike times, the most commonly adopted approach over the past several years is to apply the generalised backpropagation algorithm to the unrolled computational graph (\Cref{fig:2.5}(b)) \cite{hunsberger2015spiking, shrestha2018slayer, bellec2018long, esser2016convolutional, huh2017gradient}, i.e., backpropagation through time (BPTT). Working backwards from the final output of the network, the gradient flows from the loss to all descendants. In this way, computing the gradient through an SNN is mostly the same as that of an RNN by iterative application of the chain rule. \Cref{fig:bptt}(a) depicts the various pathways of the gradient $\partial \mathcal{L}/\partial W$ from the parent ($\mathcal{L}$) to its leaf nodes ($W$). In contrast, backprop using spike times only follows the gradient pathway whenever a neuron fires, whereas this approach takes every pathway regardless of the neuron firing. The final loss is the sum of instantaneous losses $\sum_t \mathcal{L}[t]$, though the loss calculation can take a variety of other forms as described in \Cref{sec:obj}.

Finding the derivative of the total loss with respect to the parameters allows the use of gradient descent to train the network, so the goal is to find $\partial\mathcal{L}/\partial W$. The parameter $W$ is applied at every time step, and the application of the weight at a particular step is denoted $W[s]$. Assume an instantaneous loss $\mathcal{L}[t]$ can be calculated at each time step (taking caution that some objective functions, such as the mean square spike rate loss (\Cref{sec:3sro}), must wait until the end of the sequence to accumulate all spikes and generate a loss). As the forward pass requires moving data through a directed acyclic graph, each application of the weight will only affect present and future losses. The influence of $W[s]$ on $\mathcal{L}[t]$ at $s=t$ is labelled the \textit{immediate influence} in \Cref{fig:bptt}(a). For $s<t$, we refer to the impact of $W[s]$ on $L[t]$ as the \textit{prior influence}. The influence of all parameter applications on present and future losses are summed together to define the global gradient:

\begin{equation} \label{eq:bptt}
    \frac{\partial \mathcal{L}}{\partial W} = \sum_t \frac{\partial \mathcal{L}[t]}{\partial W} = \sum_t \sum_{s\leq t}\frac{\partial \mathcal{L}[t]}{\partial W[s]} \frac{\partial W[s]}{\partial W}
\end{equation}

A recurrent system will constrain the weight to be shared across all steps: $W[0]=W[1]=\cdots=W$. Therefore, a change in $W[s]$ will have an equivalent effect on all other values of $W$, which suggests that $\partial W[s]/\partial W=1$, and \Cref{eq:bptt} simplifies to:

\begin{equation}\label{eq:bptt2}
     \frac{\partial \mathcal{L}}{\partial W} = \sum_t \sum_{s\leq t}\frac{\partial \mathcal{L}[t]}{\partial W[s]}
\end{equation}


\begin{tcolorbox}[colback=bg,colframe=black,title=Practical Note: Do I need to remember all of this to use SNNs?]
  Thankfully, gradients rarely need to be calculated by hand as most deep learning packages come with an automatic differentiation engine. Using SNNs does not require an intricate knowledge of the internals. But advancing SNN research most certainly does!
\end{tcolorbox}

Isolating the immediate influence at a single time step as in \Cref{fig:5}(d) makes it clear that we run into the spike non-differentiability problem in the term $\partial S/\partial U \in \{0, \infty\}$. The act of thresholding the membrane potential is functionally equivalent to applying a shifted Heaviside operator, which is non-differentiable.

The solution is actually quite simple. During the forward pass, as per usual, apply the Heaviside operator to $U[t]$ in order to determine whether the neuron spikes. But during the backward pass, substitute the Heaviside operator with a continuous function, $\tilde{S}$ (e.g., sigmoid). The derivative of the continuous function is used as a \textbf{}substitute $\partial S/\partial U \leftarrow \partial \tilde{S}/\partial U$, and is known as the surrogate gradient approach (\Cref{fig:5}(d)).

\subsubsection{Surrogate Gradients}

A major advantage of surrogate gradients is they help with overcoming the dead neuron problem. To make the dead neuron problem more concrete, consider a neuron with a threshold of $\theta$, and one of the following cases occurs:

\begin{enumerate}
    \item The membrane potential is below the threshold: $U < \theta$
    \item The membrane potential is above the threshold: $U > \theta$
    \item The membrane potential is exactly at the threshold: $U = \theta$
\end{enumerate}

In Case 1, no spike is elicited, and the derivative would be $\partial S/\partial U_{U<\theta}=0$. In Case 2, a spike would fire, but the derivative remains $\partial S/\partial U_{U>\theta}=0$. Applying either of these to the chain of equations in \Cref{fig:5}(a) will null $\partial\mathcal{L}/\partial W=0$. In the improbable event of Case 3, $\partial S/\partial U_{U=\theta}=\infty$, which swamps out any meaningful gradient when applied to the chain rule\footnote{Whether or not a spike occurs in Case 3 depends on your code implementation.}. But approximating the gradient, $\partial \tilde{S}/\partial U$, solves this.

One example is to replace the non-differentiable term with the threshold-shifted sigmoid function, but only during the backward pass. This is illustrated in \Cref{fig:5}(d). More formally:

\begin{equation}
    \sigma (\cdot) = \frac{1}{1+e^{\theta-U}},
\end{equation}

and therefore,

\begin{equation}
    \frac{\partial S}{\partial U} \leftarrow \frac{\partial \tilde{S}}{\partial U} =  \sigma '(\cdot) = \frac{e^{\theta - U}}{(e^{\theta-U}+1)^2}.
\end{equation}

This means learning only takes place if there is spiking activity. Consider a synaptic weight attached to the input of a spiking neuron, $W_{\rm in}$ and another weight at the output of the same neuron, $W_{\rm out}$. Say the following sequence of events occurs:

\begin{wrapfigure}{r}{3.5cm}
\hspace{0.75cm}\includegraphics[scale=0.85]{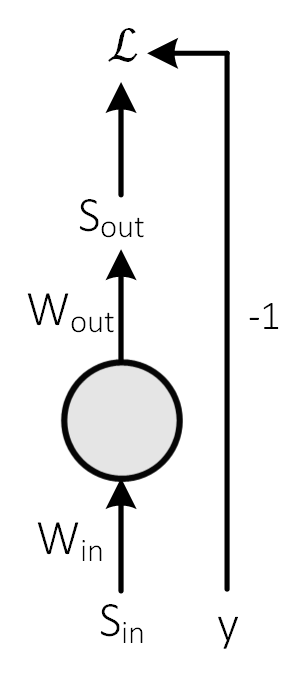}
\caption{Sequence of steps during the forward pass.}\label{wrap-fig:1}
\end{wrapfigure} 

\begin{enumerate}
    \item An input spike, $S_{\rm in}$ is scaled by $W_{\rm in}$
    \item The weighted spike is added as an input current injection to the spiking neuron (\Cref{eq:2})
    \item This may cause the neuron to trigger a spike, $S_{\rm out}$
    \item The output spike is weighted by the output weight $W_{\rm out}$
    \item This weighted output spike varies some arbitrary loss function, $\mathcal{L}$
\end{enumerate}


Let the loss function be the Manhattan distance between a target value $y$ and the weighted spike:

\begin{equation*}
    \mathcal{L} = | W_{\rm out} S_{\rm out} - y|,
\end{equation*}

where updating $W_{\rm out}$ requires:

\begin{equation*}
    \frac{\partial \mathcal{L}}{\partial W_{\rm out}} = S_{\rm out}
\end{equation*}

More generally: \textbf{a spike must be triggered for a weight to be updated.} The surrogate gradient does not change this.

Now consider the case for updating $W_{\rm in}$, where the following derivative must be calculated:

\begin{equation*}
    \frac{\partial \mathcal{L}}{\partial W_{\rm in}} = \underbrace{\frac{\partial \mathcal{L}}{\partial S_{\rm out}}}_{A} \underbrace{\frac{\partial S_{\rm out}}{\partial U_{}}}_{B} \underbrace{\frac{\partial U}{\partial W_{\rm in}}}_{C}
\end{equation*}

\begin{itemize}
    \item \textbf{Term A} is simply $W_{\rm out}$ based on the above equation for $\mathcal{L}$
    \item \textbf{Term B} would almost always be 0, unless substituted for a surrogate gradient
    \item \textbf{Term C} is $S_{\rm in}$ (see \Cref{eq:2} where $X=S_{\rm in}$)
\end{itemize}

To summarize: the surrogate gradient enables errors to propagate to earlier layers, regardless of spiking. \textbf{But spiking is still needed to trigger a weight update.}

\begin{tcolorbox}[colback=bg,colframe=black,title=Practical Note: The `Best' Surrogate Gradient?]
Various work empirically explore different surrogate gradients. These include triangular functions, fast sigmoid and sigmoid functions, straight-through estimators, and various other weird shapes. Is there a best surrogate gradient? In our experience, we have found the following function to be the best starting point:

\begin{equation*}
\frac{\partial\tilde{S}}{\partial U} = \frac{1}{\pi} \frac{1}{1+(\pi U)^2}
\end{equation*}

You might see this referred to as the `arctan' surrogate gradient, first proposed in Ref.~\cite{fang2021deep}. This is because the integral of this function is:

\begin{equation*}
    \tilde{S} = \frac{1}{\pi}{\rm arctan}(\pi U)
\end{equation*}

As of 2023, this is the default surrogate gradient in snnTorch. We have no idea why it works so well.
\end{tcolorbox}

To reiterate, surrogate gradients will not enable learning in the absence of spiking. This provokes an important distinction between the dead neuron problem and the vanishing gradient problem. A dead neuron is one that does not fire, and therefore does not contribute to the loss. This means the weights attached to that neuron have no `credit' in the credit assignment problem. The relevant gradient terms during the training process will remain at zero. Therefore, the neuron cannot learn to fire later on and so is stuck \textit{forever}, not contributing to learning.

On the other hand, vanishing gradients can arise in ANNs as well as SNNs. For deep networks, the gradients of the loss function can become vanishingly small as they are successively scaled by values less than `1' when using several common activation functions (e.g., a sigmoid unit). In much the same way, RNNs are highly susceptible to vanishing gradients because they introduce an additional layer to the unrolled computational graph at each time step. Each layer adds another multiplicative factor in calculating the gradient, which makes it susceptible to vanishing if the factor is less than `1', or exploding if greater than `1'. The ReLU activation became broadly adopted to reduce the impact of vanishing gradients, but remains underutilised in surrogate gradient implementations \cite{he2015delving}.

Surrogate gradients do not need to be explicitly defined in snnTorch as the arctan surrogate is applied by default. But the following code snippet shows how you might use an alternative with a leaky integrate-and-fire neuron:

\begin{mintedbox}{python}
import snntorch as snn
from snntorch import surrogate

lif_1 = snn.Leaky(beta=0.9, spike_grad=surrogate.fast_sigmoid())
lif_2 = snn.Leaky(beta=0.9, spike_grad=surrogate.sigmoid())
lif_3 = snn.Leaky(beta=0.9, spike_grad=surrogate.straight_through_estimator())
lif_4 = snn.Leaky(beta=0.9, spike_grad=surrogate.triangular())
\end{mintedbox}

Surrogate gradients have been used in most state-of-the-art experiments that natively train an SNN \cite{hunsberger2015spiking, shrestha2018slayer, bellec2018long, esser2016convolutional, huh2017gradient}. A variety of surrogate gradient functions have been used to varying degrees of success, and the choice of function can be treated as a hyperparameter. 
While several studies have explored the impact of various surrogates on the learning process \cite{zenke2021remarkable, neftci2019surrogate}, our understanding tends to be limited to what is known about biased gradient estimators. There is a lot left unanswered here. For example, if we can get away with approximating gradients, then perhaps surrogate gradients can be used in tandem with random feedback alignment. This involves replacing weights with random matrices during the backward pass. Rather than pure randomness, perhaps local approximations can be made that follow the same spirit of a surrogate gradient.

In summary, taking the gradient only at spike times provides an unbiased estimator of the gradient, at the expense of losing the ability to train dead neurons. Surrogate gradient descent flips this around, enabling dead neurons to backpropagate error signals by introducing a biased estimator of the gradient. There is a tug-of-war between bringing dead neurons back to life and introducing bias. Given how prevalent surrogate gradients have become, we will linger a little longer on the topic in describing their relation to model quantization. Understanding how approximations in gradient descent impacts learning will very likely lead to a deeper understanding of why surrogate gradients are so effective, how they might be improved, and how backpropagation can be simplified by making approximations that reduce the cost of training without harming an objective.

\subsubsection{The Link Between Surrogate Gradients and Quantized Neural Networks}
Surrogate gradients have been around under a few disguises for over a decade now. Hinton overcame the challenge of thresholding activations and weights in binarized neural networks by simply ignoring them during the backward pass \cite{hinton2012}. He coined the term, `straight-through-estimator', as the gradient passes `straight through' the non-differentiable operator. Equivalently, this is like setting the surrogate gradient to be $\partial \tilde{S}/\partial U=1$. The exact same methodology is applied when training quantized neural networks \cite{eshraghian2022navigating, eshraghian2022memristor}.

\begin{tcolorbox}[colback=bg,colframe=black,title=A Brief History of Surrogate Gradients]
A couple of years after Hinton introduced the straight-through-estimator, Hunsberger and Eliasmith used `softened' functions for the firing rate of a leaky integrator neuron, making them amenable to backpropagation. Lee, Delbruck and Pfeiffer subsequently provided the first demonstration of approximating the backward path of spike signals in 2016~\cite{lee2016training}. This is akin to what we presently do today. But Shrestha and Orchard were the first to release code that was compatible PyTorch with SLAYER in 2018, where a fast sigmoid gradient was used on the backward pass~\cite{shrestha2018slayer}. 2019 is when the term `surrogate gradient' was coined by Zenke, Mostafa and Neftci~\cite{neftci2019surrogate}, and Zenke's codebase \textit{SpyTorch} showed how features in PyTorch could be used to implement gradient approximations very easily~\cite{zenke2019spytorch}. This spawned a proliferation of various applications, projects and libraries, including snnTorch, that continue to extend these techniques.
\end{tcolorbox}

Training quantized neural networks involves adjusting the weights and activations to use lower-precision fixed-point representations while maintaining acceptable performance and accuracy \cite{fiesler1990weight, balzer1991weight}. Quantized-fixed point arithmetic requires fewer computational resources and lower memory storage compared to floating-point arithmetic, and commonly used in accelerators, both neuromorphic and otherwise, as a result. Several methods have been proposed to construct quantized neural networks, and they can be broadly categorized into the following approaches:

\begin{enumerate}
    \item \textbf{Post-training quantization:} A neural network is first trained using standard floating-point arithmetic. Once training is complete, the weights and activations are quantized to lower-precision fixed-point representations. 
    Post-training quantization is simple to implement and computationally efficient, but it may result in a significant loss of accuracy for certain models or tasks.
    \item \textbf{Quantization-aware training:} This method involves training the neural network with quantization built into the forward pass during the training process. The quantization process is non-differentiable, and is thus ignored during the gradient calculation step by applying Hinton's straight-through-estimator. The weight update is applied to the full precision weight, which is quantized only during the forward-pass. This allows the model to learn how to compensate for the quantization errors during training, leading to better performance and accuracy compared to post-training quantization. However, quantization-aware training is more computationally intensive and may require modifications to the training algorithm \cite{hou2018loss}.
    \item \textbf{Mixed-precision training:} Different parts of the neural network use different levels of numerical precision. For example, the forward pass may use lower-precision fixed-point arithmetic, while the backward pass and weight updates use higher-precision floating-point arithmetic. This approach can help maintain the benefits of reduced computational complexity and memory requirements while minimizing the impact on model accuracy \cite{micikevicius2017mixed}.
    \item \textbf{Binary and ternary neural networks:} These are extreme cases of quantized neural networks, where the weights and activations are quantized to binary or ternary values, typically $\{$-1, 0, 1$\}$ for ternary networks and $\{$-1, 1$\}$ for binary networks. Training such networks often involves learning a real-valued scaling factor alongside the binary or ternary weights to improve the model's expressive power. These ultra-low precision networks can significantly reduce computational requirements and power consumption, but they may suffer from reduced accuracy or increased model complexity.
\end{enumerate}

Several studies have shown that SNNs are extremely robust to quantization when quantization-aware training is used. In the extreme case, binarized weights appeared to have a far weaker impact on a range of classification problems than an equivalent non-spiking neural network. Our working theory is because approximations and truncation errors are likely to be absorbed in the sub-threshold dynamics of the neuron~\cite{eshraghian2022navigating, eshraghian2022memristor}.

The following code sample shows the use of the Python library, \textit{Brevitas}, in constructing quantized SNNs. Brevitas already accounts for the straight-through-estimator gradient during training so the developer does not need to make any modifications during the backward-pass~\cite{brevitas}. It should be noted that this approach models a reduced precision network, but does not represent variables in a reduced precision format.
\begin{mintedbox}{python}
import snntorch as snn
import torch.nn as nn
import brevitas.nn as qnn

# full precision model
net = nn.Sequential(nn.Linear(784, 10),
                    snn.Leaky(beta=0.9, init_hidden=True))

# quantized model
num_bits = 8
quant_net = nn.Sequential(qnn.QuantLinear(784, 10, weight_bit_width=num_bits),
                          snn.Leaky(beta=0.9, init_hidden=True))
\end{mintedbox}

The above code snippets quantize weights and activations. The membrane potential and hidden states are often neglected, and post-quantized outside of the training process. State-based quantization aware training is also possible, where the membrane potential is discretized during the forward-pass. This is straightforward to account for in snnTorch by passing an argument to a neuron model, which triggers quantization of the membrane potential during the forward pass:

\begin{mintedbox}{python}
from snntorch.functional import quant

# set the quantization parameters
q_lif = quant.state_quant(num_bits=4)

# state-quantized model
quant_net = nn.Sequential(qnn.QuantLinear(784, 10, weight_bit_width=num_bits),
                          snn.Leaky(beta=0.9, init_hidden=True, state_quant=q_lif))
\end{mintedbox}

\begin{tcolorbox}[colback=bg,colframe=black,title=Practical Note: Quantization Ranges]
Say you have $N$ bits available to represent weights and states. How do you determine the range these $N$ bits should span? Weights will use the minimum and maximum full precision weight values as the range. States are a lesser explored in how they should best be represented, but our early experimental results indicate that for a normalized resting potential of $0$~V, negative values can be safely clipped. This provides positive states a finer resolution rather than wasting them on ranges that do not trigger spikes. Naturally, this does not hold true where inhibitory spikes can be triggered from negative thresholds.
\end{tcolorbox}


\subsubsection{A Bag of Tricks in BPTT with SNNs}
Many advances in deep learning stem from a series of incremental techniques that bolster the learning capacity of models. These techniques are applied in conjunction to boost model performance. For example, He \textit{et al.}'s work in `\textit{Bag of tricks for image classification with convolutional neural networks}' not only captures the honest state of deep learning in the title alone, but also performs an ablation study of `hacks' that can be combined to improve optimization during training \cite{he2019bag}. Some of these techniques can be ported straight from deep learning to SNNs, while others are SNN specific. A non-exhaustive list of these techniques are provided in this section. These techniques are quite empirical and each bullet would have its own `Practical Note' text box, but then this paper would just turn into a bunch of boxes.

\begin{itemize}
    \item \textbf{The reset mechanism} in \Cref{eq:2} is a function of the spike, and is also non-differentiable. It is important to ensure the surrogate gradient is not cloned into the reset function as it has been empirically shown to degrade network performance \cite{zenke2021remarkable}. Quite simply, we ignore it during the backward pass. snnTorch does this automatically by detaching the reset term in \Cref{eq:2} from the computational graph by calling the `\textit{.detach()}' function.
    \item \textbf{Residual connections} work remarkably well for non-spiking nets and spiking models alike. Direct paths between layers are created by allowing the output of an earlier layer to be added to the output of a later layer, effectively skipping one or more layers in between. They are used to address the vanishing gradient problem and improve the flow of information during both forward and backward propagation, which enabled the neural network community to construct far deeper architectures, starting with the ResNet family of models and now commonly used in Transformers \cite{he2016deep}. Unsurprisingly, they work extremely well for SNNs, too \cite{fang2021deep}.
    \item \textbf{Learnable decay:} Rather than treating the decay rates of neurons as hyperparameters, it is also common practice to make them learnable parameters. This makes SNNs resemble conventional RNNs much more closely. Doing so has shown to improve testing performance on datasets with time-varying features \cite{perez2021neural}.
    \item \textbf{Graded Spikes:} Passive dendritic properties can attenuate action potentials, as can the cable-like properties of the axon. This feature can be coarsely accounted for as graded spikes. Each neuron has an additional learnable parameter that determines how to scale an output spike. Neuronal activations are no longer constrained to $\{$1, 0$\}$. Can this still be thought of as a SNN? From an engineering standpoint, if a spike must be broadcast to a variety of downstream neurons with a 8 or 16-bit destination address, then adding another several bits to the payload can be worth it. The 2nd generation Loihi chip from Intel Labs incorporates graded spikes in such a way that sparsity is preserved. Furthermore, the vector of learnt values scales linearly with the number of neurons in a network, rather than quadratically with weights. It therefore contributes a minor cost in comparison to other components of an SNN.
    \item \textbf{Learnable Thresholds} have \textit{not} been shown to help the training process. This is likely due to the discrete nature of thresholds, giving rise to non-differentiable operators in a computational graph. On the other hand, normalizing the values that are passed into a threshold significantly helps. Adopting batch-normalization in convolutional networks helps boost performance, and learnable normalization approaches may act as an effective surrogate for learnable thresholds \cite{ioffe2015batch, kim2021revisiting, duan2022temporal}.
    \item \textbf{Pooling} is effective for downsampling large spatial dimensions in convolutional networks, and achieving translational invariance. If max-pooling is applied to a sparse, spiking tensor, then tie-breaking between 1's and 0's does not make much sense. One might expect we can borrow ideas from training binarized neural networks, were pooling is applied to the activations \textit{before} they are thresholded to binarized quantities. This corresponds to applying pooling to the membrane potential, in a manner that resembles a form of `local lateral inhibition'. But this does not necessarily lead to optimal performance in SNNs. Interestingly, Yu \textit{et al.} applied pooling to the spikes instead. Where multiple spikes occurred in a pooling window, a tie-break would occur randomly among them \cite{fang2021deep}. While no reason was given for doing this, it nonetheless achieved state-of-the-art (at the time) performance on a series of computer vision problems. Our best guess is that this randomness acted as a type of regularization. Whether max-pooling or average-pooling is used can be treated as a hyperparameter. As an alternative, SynSense's neuromorphic hardware adopts sum-pooling, where spatial dimensions are reduced by re-routing the spikes in a receptive field to a common post-synaptic neuron.
    \item \textbf{Optimizer:} Most SNNs default to the Adam optimizer as they have classically been shown to be robust when used with sequential models \cite{kingma2014adam}. As SNNs become deeper, stochastic gradient descent with momentum seems to increase in prevalence over the Adam optimizer. The reader is referred to Godbole \textit{et al.}'s Deep Learning Tuning Playbook for a systematic approach to hyperparameter optimization that applies generally \cite{tuningplaybookgithub}.
\end{itemize}

\subsubsection{The Intersection Between Backprop and Local Learning}  \label{sec:hybrid}
An interesting result arises when comparing backpropagation pathways that traverse varying durations of time. The derivative of the hidden state over time is $\partial U[t]/\partial U[t-1]=\beta$ as per \Cref{eq:2}. A gradient that backpropagates through $n$ time steps is scaled by $\beta^n$. For a leaky neuron we get $\beta < 1$, which causes the magnitude of a weight update to exponentially diminish with time between a pair of spikes. This proportionality is illustrated in \Cref{fig:bptt}(b). This result shows how the strength of a synaptic update is exponentially proportional to the spike time difference between a pre- and post-synaptic neuron. In other words, weight updates from BPTT closely resemble weight updates from spike-timing dependent plasticity (STDP) learning curves (\Cref{app:bpus}) \cite{bi1998synaptic}.

Is this link just a coincidence? BPTT was derived from function optimization. STDP is a model of a biological observation. Despite being developed via completely independent means, they converge upon an identical result. This could have immediately practical implications, where hardware accelerators that train models can excise a chunk of BPTT and replace it with the significantly cheaper and local STDP rule. Adopting such an approach might be thought of as an online variant of BPTT, or as a gradient-modulated form of STDP.

\subsection{Long-Term Temporal Dependencies}
Neural and synaptic time constants span timescales typically on the order of 1-100s of milliseconds. With such time scales, it is difficult to solve problems that require long-range associations that are larger than the slowest neuron or synaptic time constant. 
Such problems are common in natural language processing and reinforcement learning, and are key to understanding behavior and decision making in humans.
This challenge is a a huge burden on the learning process, where vanishing gradients drastically slow the convergence of the neural network.
LSTMs \cite{hochreiter1997long} and, later, GRUs \cite{cho2014learning} introduced slow dynamics designed to overcome memory and vanishing gradient problems in RNNs.
Thus, a natural solution for networks of spiking neurons is to complement the fast timescales of neural dynamics with a variety of slower dynamics.
Mixing discrete and continuous dynamics may enable SNNs to learn features that occur on a vast range of timescales. Examples of slower dynamics include:


\begin{itemize}
    \item \textbf{Adaptive thresholds:} After a neuron fires, it enters a refractory period during which it is more difficult to elicit further spikes from the neuron. This can be modeled by increasing the firing threshold of the neuron $\theta$ every time the neuron emits a spike. After a sufficient time in which the neuron has spiked, the threshold relaxes back to a steady-state value. 
    Homeostatic thresholds are known to promote neuronal stability in correlated learning rules, such as STDP which favours long term potentiation at high frequencies regardless of spike timing \cite{watt2010homeostatic, sjostrom2001rate}. More recently, it has been found to benefit gradient-based learning in SNNs as well \cite{bellec2018long} (\Cref{app:c2}).

    \item \textbf{Recurrent attention:} Hugely popularized from natural language generation, self-attention finds correlations between tokens of vast sequence lengths by feeding a model with all sequential inputs at once. This representation of data is not quite how the brain processes data. Several approaches have approximated self-attention into a sequence of recurrent operations, where SpikeGPT is the first application in the spiking domain and successfully achieved language generation~\cite{zhu2023spikegpt}. In addition to more complex state-based computation, SpikeGPT additionally employs dynamical weights that vary over time.
    
    \item \textbf{Axonal delays:} The wide variety of axon lengths means there is a wide range of spike propagation delays. Some neurons have axons as short as 1~mm, whereas those in the sciatic nerve can extend up to a meter in length. The axonal delay can be a learned parameter spanning multiple time steps \cite{shrestha2018slayer, schrauwen2004extending, taherkhani2015dl}. A lesser explored approach accounts for the varying delays in not only axons, but also across the dendritic tree of a neuron. Coupling axonal and dendritic delays together allows for a fixed delay per synapse.
    
    \item \textbf{Membrane Dynamics:} 
    We already know how the membrane potential can trigger spiking, but how does spiking impact the membrane? Rapid changes in voltage cause an electric field build-up that leads to temperature changes in cells. Joule heating scales quadratically with voltage changes, which affects the geometric structure of neurons and cascades into a change in membrane capacitance (and thus, time constants). Decay rate modulation as a function of spike emission can act as a second-order mechanism to generate neuron-specific refractory dynamics.
    
    \item \textbf{Multistable Neural Activity:}
    Strong recurrent connections in biological neural networks can support multistable dynamics \cite{renart2004mean}, which facilitates stable information storage over time. Such dynamics, often called attractor neural networks \cite{amit1992modeling}, are believed to underpin working memory in the brain \cite{renart2003robust, rigotti2010internal}, and is often attributed to the prefrontal cortex. The training of such networks using gradient descent is challenging, and has not been attempted using SNNs as of yet \cite{miller2019stable}.
\end{itemize}

Several rudimentary slow timescale dynamics have been tested in gradient-based approaches to training SNNs with a good deal of success \cite{shrestha2018slayer, bellec2018long}, but there are several neuronal dynamics that are yet to be explored. LSTMs showed us the importance of temporal regulation of information, and effectively cured the short-term memory problem that plagued RNNs. Translating more nuanced neuronal features into gradient-based learning frameworks can undoubtedly strengthen the ability of SNNs to represent dynamical data in an efficient manner.

\newpage 

\section{Online Learning}\label{sec:ol}
\subsection{Temporal Locality}
As incredible as our brains are, sadly, they are not time machines. It is highly unlikely our neurons are breaching the space-time continuum to explicitly reference historical states to run the BPTT algorithm.
As with all computers, brains operate on a physical substrate which dictates the operations it can handle and where memory is located. While conventional computers operate on an abstraction layer, memory is delocalised and communicated on demand, thus paying a considerable price in latency and energy. 
Brains are believed to operate on local information, which means the best performing approaches in temporal deep learning, namely BPTT, are biologically implausible. 
This is because BPTT requires the storage of the past inputs and states in memory. As a result, the required memory scales with time, a property which limits BPTT to small temporal dependencies.
To solve this problem, BPTT assumes a finite sequence length before making an update, while truncating the gradients in time. 
This, however, severely restricts the temporal dependencies that can be learned.

The constraint imposed on brain-inspired learning algorithms is that the calculation of a gradient should, much like the forward pass, be temporally local, \emph{i.e.} that they only depend on values available at either present time $t$ or $t-1$. To address this, we turn to online algorithms that adhere to \textit{temporal locality}. Real-time recurrent learning (RTRL) proposed back in 1989 is one prominent example. 

\subsection{Real-Time Recurrent Learning}
RTRL estimates the same gradients as BPTT, but relies on a set of different computations that make it temporally, but not spatially, local \cite{williams1989learning}. 
Since RTRL's memory requirement does not grow with time, then why is it not used in favour of BPTT? BPTT's memory usage scales with the product of time and the number of neurons; it is $\mathcal{O}(nT)$. For RTRL, an additional set of computations must be introduced to enable the network to keep track of a gradient that evolves with time. These additional computations result in a $\mathcal{O}(n^3)$ memory requirement, which often exceeds the demands of BPTT. But the push for continuously-learning systems that can run indefinitely long has cast a spotlight back on RTRL (and variants \cite{tallec2017unbiased, mujika2018approximating, roth2018kernel, murray2019local, marschall2020unified}), with a focus on improving computational and memory efficiency.

Let us derive what new information needs to be propagated forward to enable real-time gradient calculation for an SNN. As in \Cref{eq:bptt2}, let $t$ denote real time in the calculation of $\partial\mathcal{L}/\partial W$, and let the instantaneous loss $\mathcal{L}[t]$ be a measure of how well the instantaneously predicted output $\hat{Y}[t]$ matches the target output $Y[t]$. Depending on the type of loss function in use, $\hat{Y}[t]$ might simply be the spike output of the final layer $S_{\rm out}[t]$ or the membrane potential $U[t]$.
In either case, $\partial\mathcal{L}[t]/U[t]$ does not depend on any values that are not present at $t$, so it is natural to calculate this term in an online manner. The key problem is deriving $\partial U[t]/\partial W$ such that it only relies on values presently available at $t-1$ and $t$.

First we define the influence of parameter $W$ on the membrane potential $U[t]$ as $m[t]$, which serves to track the derivative of the present-time membrane potential with respect to the weight. We then unpack it by one time step: 

\begin{equation} \label{eq:inf}
    m[t] = \frac{\partial U[t]}{\partial W} = \sum_{s\leq t}\frac{\partial U[t]}{\partial W[s]} = \underbrace{\sum_{s\leq t-1}\frac{\partial U[t]}{\partial W[s]}}_{\text{\rm prior}} + \underbrace{\frac{\partial U[t]}{\partial W[t]}}_{\text{\rm immediate}}
\end{equation}

The immediate and prior influence components are graphically illustrated in \Cref{fig:bptt}(a). The immediate influence is also natural to calculate online, and evaluates to the unweighted input to the neuron $X[t]$. The prior influence relies on historical components of the network:


\begin{equation}\label{eq:temp}
    \sum_{s\leq t-1}\frac{\partial U[t]}{\partial W[s]} = \sum_{s\leq t-1}\underbrace{\frac{\partial U[t]}{\partial U[t-1]}}_{\text{\rm temporal}}\frac{\partial U[t-1]}{\partial W[s]}
\end{equation}

Based on \Cref{eq:2}, in the absence of explicitly recurrent connections, the temporal term evaluates to $\beta$. From \Cref{eq:inf}, the second term is the influence of parameters on $U[t-1]$, which is by definition $m[t-1]$. Substituting these back into \Cref{eq:inf} gives:

\begin{equation}\label{eq:online}
    m[t] = \beta m[t-1] + x[t]
\end{equation}

\begin{figure}
    \centering
    \includegraphics[width=0.85\linewidth]{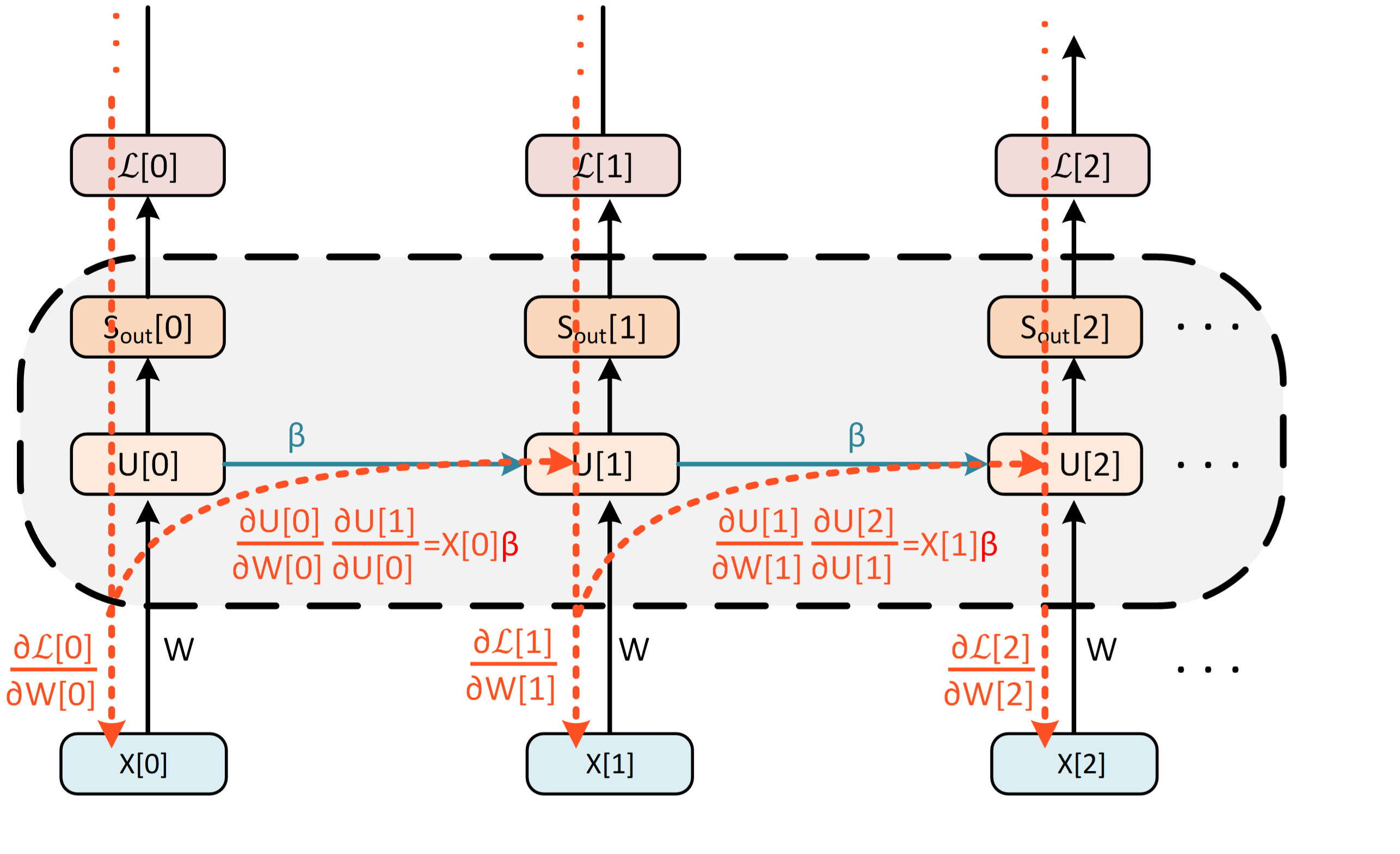}
    \caption{Real-time recurrent learning gradient pathways. The node for synaptic current, $I$, has been removed as it does not alter the result here.}
    \label{fig:rtrl}
\end{figure}

This recursive formula is updated by passing the unweighted input directly to $m[t]$, and recursively decaying the influence term by the membrane potential decay rate $\beta$. The gradient that is ultimately used with the optimizer can be derived with the chain rule:

\begin{equation}
    \frac{\partial\mathcal{L}[t]}{\partial W} = \frac{\partial \mathcal{L}[t]}{\partial U[t]}\frac{\partial U[t]}{\partial W} \equiv \bar{c}[t] m[t]
\end{equation}

where $\bar{c}[t]=\partial \mathcal{L}[t]/\partial U[t]$ is the immediate credit assignment value obtained by backpropagating the instantaneous loss to the hidden state of the neuron, for example, by using a surrogate gradient approach. The calculation of $m[t]$ only ever depends on present time inputs and the influence at $t-1$, thus enabling the loss to be calculated in an online manner. The input spike now plays a role in not only modulating the membrane potential of the neuron, but also the influence $m[t]$. The general flow of gradients is depicted in \Cref{fig:rtrl}.

An intuitive, though incomplete, way to think about RTRL is as follows. By reference to \Cref{fig:rtrl}, at each time step, a backward-pass that does not account for the history of weight updates is applied: $\partial \mathcal{L}[0]/\partial W[0]$ (the immediate influence). Rather than directing gradients backwards through time, the partial derivative $\partial U[0]/\partial W[0]$ is `pushed' forward in time. In doing so, it is scaled by the temporal term, $X[0]\beta$. This term modulates the immediate influence at the next time step. This can be thought of as a gradient term that `snowballs' forward in time as a result of modulating and accumulating with the immediate influence term, but also loses a bit of `momentum' every time the temporal term $\beta$ decays it.

In the example above, the RTRL approach to training SNNs was only derived for a single neuron and a single parameter. A full scale neural network replaces the influence value with an influence matrix $\vect{M}[t]\in\mathbb{R}^{n\times P}$, where $n$ is the number of neurons and $P$ is the number of parameters (approximately $\mathcal{O}(n^2$) memory). Therefore, the memory requirements of the influence matrix scales with $\mathcal{O}(n^3)$.

Recent focus in online learning aims to reduce the memory and computational demands of RTRL. This is generally achieved by decomposing the influence matrix into simpler parts, approximating the calculation of $\vect{M}[t]$ by either completely removing terms or trading them for stochastic noise instead \cite{tallec2017unbiased, mujika2018approximating, roth2018kernel, murray2019local}. Marschall~\textit{et al.} provides a systematic treatment of approximations to RTRL in RNNs in \cite{marschall2020unified}, and variations of online learning have been applied specifically to SNNs in \cite{kaiser2020synaptic, bellec2020solution, bohnstingl2020online}.

\subsubsection{RTRL Variants in SNNs}
Since 2020, a flurry of forward-mode learning algorithms have been tailored to SNNs. All such works either modify, re-derive, or approximate RTRL:

\begin{itemize}
    \item \textbf{e-prop (\textit{Bellec et al., 2020}~\cite{bellec2020solution}):} RTRL is combined with surrogate gradient descent. Recurrent spiking neurons are used where output spikes are linearly transformed and then fed back to the input of the same neurons. The computational graph is detached at the explicit recurrent operation, but retained for implicit recurrence (i.e., where membrane potential evolves over time). Projecting output spikes into a higher-dimensional recurrent space acts like a reservoir, though leads to biased gradient estimators that underperforms compared to BPTT.
    \item \textbf{decolle (\textit{Kaiser et al., 2020} \cite{kaiser2020synaptic}):} `Deep continuous online learning' also combines RTRL with surrogate gradient descent. This time, greedy local losses are applied at \textbf{every} layer \cite{bengio2007greedy}. As such, errors only need to be propagated back to a single layer at a time. This means that errors do not need to traverse through a huge network, which reduces the burden of the spatial credit assignment problem. This brings about two challenges: 1) not many problems can be cast into a form with definable local losses, and 2) greedy local learning prioritizes immediate gains without considering an overall objective.
    \item \textbf{OSTL (\textit{Bonhstingl et al., 2022} \cite{bohnstingl2022online}):} `Online spatio-temporal learning' re-derives RTRL. The spatial components of backpropation and temporal components are factored into two separate terms; e.g., one that tracks the `immediate' influence, and one that tracks the `prior influence' from \Cref{eq:inf}. 
    \item \textbf{ETLP (\textit{Quintana et al., 2023} \cite{quintana2023etlp}):} `Event-based Three-Factor Local Plasticity' combines e-prop with direct random target projection (DRTP: Frenkel and Lefebvre, 2019, \cite{frenkel2021learning}). In other words, the weights in the final layer are updated based on an approximation of RTRL. Earlier layers are updated based on partial derivatives that do not rely on a global loss, and are spatially `local' to the layer. Instead, the target output is used to modulate these gradients. This addresses spatial credit assignment by using signals from a target, rather than backpropagating gradients in the immediate influence term of \Cref{eq:inf}. The cost is that it both inherits drawbacks from e-prop and DRTP. DRTP prioritizes immediate gains without considering an overall objective, similar to greedy local learning.
    \item \textbf{OSTTP (\textit{Ortner and Pes, et al., 2023} \cite{ortner2023online}):} `Online Spatiotemporal Learning with Target Projection' combines OSTL (functionally equivalent to RTRL) with DRTP. It inherits the drawbacks of DRTP, while addressing the spatial credit assignment problem.
    \item \textbf{FPTT (\textit{Kag, et al., 2021} \cite{kag2021training}):} `Forward Propagation Through Time' considers RTRL for sequence-to-sequence models with time varying losses. A regularization term is applied to the loss at each step to ensure stability during the training process. \textit{Yin et al.} subsequently applied FPTT to SNNs with more complex neuron models with richer dynamics \cite{yin2023accurate}.
\end{itemize}

This is a non-exhaustive list of RTRL alternatives, and can appear quite daunting at first. But all approaches effectively stem from RTRL. The dominant trends include:
\begin{enumerate}
    \item Approximating RTRL to test how much of an approximation the training procedure can tolerate without completely failing \cite{bellec2020solution}
    \item Replacing the immediate influence with global-modulation of a loss or target to address spatial credit assignment \cite{kaiser2020synaptic, quintana2023etlp, ortner2023online}
    \item Modifying the objective to promote stable training dynamics \cite{kag2021training}
    \item Identifying similarities to biology by factorizing RTRL into eligibility traces and/or three-factor learning rules \cite{bellec2020solution, yin2023accurate, quintana2023etlp}
\end{enumerate}

Several RTRL-variants claim to outperform BPTT in terms of loss minimization, though we take caution with such claims as the two effectively become identical to BPTT for the case where weight updates are deferred to the end of a sequence. We also note caution with claims that suggest improvements over RTRL, as RTRL can be thought of as the most general case of forward-model learning applied to any generic architecture. Most reductions in computational complexity arise because they are narrowly considered for specific architectures, or otherwise introduce approximations into their models. In contrast, Tallec and Ollivier developed an `unbiased online recurrent optimization' scheme where stochastic noise is used and ultimately cancelled out, leading to quadratic (rather than cubic) computational complexity with network size \cite{tallec2017unbiased}.

\subsubsection{Practical Considerations with RTRL}

Several practical considerations should be accounted for when implementing online learning algorithms. For an approach that closely resembles BPTT, the gradient accumulated at the end of the sequence can be used to update the network, which is referred to as a `deferred' update. Alternatively, it is possible to update the network more regularly as a gradient is consistently available. While this latter option is a more accurate reflection of  biological learning (i.e., training and inference are not decoupled processes), there are two issues that must be treated with care. Firstly, adaptive optimizers such as Adam naturally reduce the learning rate as parameters approach optimal values \cite{kingma2014adam}. When applying frequent updates on a given batch of data, future batches will have less influence on weight updates. The result is a learning procedure that assigns a higher weighting to early data than to later data. If the sampled data does not satisfy the i.i.d assumption, which is the case when a system experiences data in an online fashion, learning may not perform well. Secondly, the reverse problem is catastrophic forgetting where new information causes the network to forget what it has previously learnt \cite{mccloskey1989catastrophic}. This is especially problematic in real-time systems because a ``real-world batch size is equal to 1''. Several approaches to overcome catastrophic forgetting in continual learning have been proposed, including using higher dimensional synapses \cite{zenke2017continual}, ensembles of networks \cite{rusu2016progressive}, pseudo-replay \cite{shin2017continual}, and penalizing weights that change excessively fast \cite{kirkpatrick2017overcoming}.

\subsection{Spatial Locality}
While temporal locality relies on a learning rule that depends only on the present state of the network, spatial locality requires each update to be derived from a node immediately adjacent to the parameter. The biologically motivated learning rules described in \Cref{sec:bml} address the spatial credit assignment problem by either replacing the global error signal with local errors, or replacing analytical/numerical derivatives with random noise \cite{lillicrap2014random}.

The more `natural' approach to online learning is perceived to be via unsupervised learning with synaptic plasticity rules, such as STDP \cite{bi1998synaptic, diehl2015unsupervised} and variants of STDP (\Cref{app:bpus}) \cite{brader2007learning, zhao2014feedforward, beyeler2013categorization,querlioz2013immunity}. These approaches are directly inspired by experimental relationships between spike times and changes to synaptic conductance. Input data is fed to a network, and weights are updated based on the order and firing times of each pair of connected neurons (\Cref{fig:bptt}(b)). The interpretation is that if a neuron causes another neuron to fire, then their synaptic strength should be increased. If a pair of neurons appear uncorrelated, their synaptic strength should be decreased. It follows the Hebbian mantra of \textit{`neurons that fire together wire together'} \cite{hebb1949organisation}.

There is a common misconception that backprop and STDP-like learning rules are at odds with one other, competing to be the long-term solution for training connectionist networks. On the one hand, it is thought that STDP deserves more attention as it scales with less complexity than backprop. STDP adheres to temporal and spatial locality, as each synaptic update only relies on information from immediately adjacent nodes. However, this relationship necessarily arises as STDP was reported using data from `immediately adjacent' neurons. On the other hand, STDP fails to compete with backprop on remotely challenging datasets. But backprop was designed with function optimization in mind, while STDP emerged as a physiological observation. The mere fact that STDP is capable at all of obtaining competitive results on tasks originally intended for supervised learning (such as classifying the MNIST dataset), no matter how simple, is quite a wonder. Rather than focusing on what divides backprop and STDP, the pursuit of more effective learning rules will more likely benefit by understanding how the two intersect.

We demonstrated in \Cref{sec:hybrid} how surrogate gradient descent via BPTT subsumes the effect of STDP. Spike time differences result in exponentially decaying weight update magnitudes, such that half of the learning window of STDP is already accounted for within the BPTT algorithm (\Cref{fig:bptt}(b)). Bengio \textit{et al.} previously made the case that STDP resembles stochastic gradient descent, provided that STDP is supplemented with gradient feedback \cite{bengio2015towards, hinton2016can}. This specifically relates to the case where a neuron's firing rate is interpreted as its activation. Here, we have demonstrated that no modification needs to be made to the BPTT algorithm for it to account for STDP-like effects, and is not limited to any specific neural code, such as the firing rate. The common theme is that STDP may benefit from integrating error-triggered plasticity to provide meaningful feedback to training a network \cite{payvand2020chip}.


\newpage
\section{Outlook}
Designing a neural network was once thought to be strictly an engineering problem whereas mapping the brain was a scientific curiosity \cite{crick1989recent}. With the intersection between deep learning and neuroscience broadening, and brains being able to solve complex problems much more efficiently, this view is poised to change. 
From the scientist's view, deep learning and brain activity have shown many correlates, which lead us to believe that there is much untapped insight that ANNs can offer in the ambitious quest of understanding biological learning. For example, the activity across layers of a neural network have repeatedly shown similarities to experimental activity in the brain. This includes links between convolutional neural networks and measured activity from the visual cortex \cite{cichy2016comparison, rajalingham2018large, schrimpf2020brain}, and auditory processing regions \cite{kell2018task}. Activity levels across populations of neurons have been quantified in many studies, but SNNs might inform us of the specific nature of such activity. 

From the engineer's perspective, neuron models derived from experimental results have allowed us to design extremely energy-efficient networks when running on hardware tailored to SNNs \cite{davies2018loihi, merolla2014million, furber2014spinnaker, neckar2018braindrop, pei2019towards, roy2019towards, kornijcuk2019recent}. Improvements in energy consumption of up to 2--3 orders of magnitude have been reported when compared to conventional ANN acceleration on embedded hardware, which provides empirical validation of the benefits available from the three S's: spikes, sparsity and static data suppression (or event-driven processing) \cite{rahimiazghadi2020hardware, ceolini2020hand, sun2022intelligence, davies2021advancing, sharifshazileh2021electronic}. These energy and latency benefits are derived from simply applying neuron models to connectionist networks, but there is so much more left to explore.


It is safe to say the energy benefits afforded by spikes are uncontroversial. But a more challenging question to address is: are spikes actually good for computation? It could be that years of evolution determined spikes solved the long-range signal transmission problem in living organisms, and everything else had to adapt to fit this constraint. If this were true, then spike-based computation would be pareto optimal with a proclivity towards energy efficiency and latency. But until we amass more evidence of a spike's purpose, we have some intuition as to where spikes shine in computation:

\begin{itemize}
    \item \textbf{Hybrid Dynamical Systems:} SNNs can model a broad class of dynamical systems by coupling discrete and continuous time dynamics into one system. Discontinuities are present in many physical systems, and spiking neuron models are a natural fit to model such dynamics.
    \item \textbf{Discrete Function Approximators:} Neural networks are universal function approximators, where discrete functions are considered to be modelled sufficiently well by continuous approximations. Spikes are capable of precisely defining discrete functions without approximation.
    \item \textbf{Multiplexing:} Spikes can encode different information in spike rate, times, or burst counts. Re-purposing the same spikes offers a sensible way to condense the amount of computation required by a system.
    \item \textbf{Message Packets:} By compressing the representation of information, spikes can be thought of as packets of messages that are unlikely to collide as they travel across a network. In contrast, a digital system requires a synchronous clock to signal that a communication channel is available for a message to pass through (even when modelling asynchronous systems).
    \item \textbf{Coincidence Detection:} Neural information can be encoded based on spatially disparate but temporally proximate input spikes on a target neuron. It may be the case that isolated input spikes are insufficient to elicit a spike from the output neuron. But if two incident spikes occur on a timescale faster than the target neuron membrane potential decay rate, this could push the potential beyond the threshold and trigger an output spike. In such a case, associative learning is taking place across neurons that are not directly connected. Although coincidence detection can be programmed in a continuous-time system without spikes, a theoretical analysis has shown that the processing rate of a coincidence detector neuron is faster than the rate at which information is passed to a neuron \cite{krips2009stochastic, brette2012computing}.
    \item \textbf{Noise Robustness:} While analog signals are highly susceptible to noise, digital signals are far more robust in long-range communication. Neurons seem to have figured this out by performing analog computation via integration at the soma, and digital communication along the axon. It is possible that any noise incident during analog computation at the soma is subsumed into the subthreshold dynamics of the neuron, and therefore eliminated. In terms of neural coding, a similar analogy can be made to spike rates and spike times. Pathways that are susceptible to adversarial attacks or timing perturbations could learn to be represented as a rate, which otherwise mitigates timing disturbances in temporal codes.
    \item \textbf{Modality normalisation:} A unified representation of sensory input (e.g., vision, auditory) as spikes is nature's way of normalising data. While this benefit is not exclusive to spikes (i.e., continuous data streams in non-spiking networks may also be normalised), early empirical evidence has shown instances where multi-modal SNNs outperform convolutional neural networks on equivalent tasks \cite{ceolini2020hand, rahimiazghadi2020hardware}.
    \item \textbf{Mixed-mode differentiation:} 
    While most modern deep learning frameworks rely on reverse-mode autodifferentiation \cite{paszke2017automatic}, it is in stark contrast to how the spatial credit assignment problem is treated in biological organisms. 
    If we are to draw parallels between backpropagation and the brain, it is far more likely that approximations of forward-mode autodifferentation are being used instead. \Cref{eq:online} in \Cref{sec:ol} describes how to propagate gradient-related terms forward in time to implement online learning, where such
    terms could be approximated by eligibility traces that keep track of pre-synaptic neuron activity in the form of calcium ions, and fades over time \cite{sanhueza2013camkii, bellec2020solution}. SNNs offer a natural way to use mixed-mode differentiation by projecting temporal terms in the gradient calculation from \Cref{eq:temp} into the future via forward-mode differentation, while taking advantage of the computational complexity of reverse-mode autodifferentation for spatial terms \cite{kaiser2020synaptic, zenke2021brain}.
\end{itemize}

A better understanding of the types of problems spikes are best suited for, beyond addressing just energy efficiency, will be important in directing SNNs to meaningful tasks. The above list is a non-exhaustive start to intuit where that might be. Thus far, we have primarily viewed the benefits of SNNs by examining individual spikes. For example, the advantages derived from sparsity and single-bit communication arise at the level of an individual spiking neuron: how a spike promotes sparsity, how it contributes to a neural encoding strategy, and how it can be used in conjuction with modern deep learning, backprop, and gradient descent. Despite the advances yielded by this spike-centric view, it is important not to develop tunnel vision. New advances are likely to come from a deeper understanding of spikes acting collectively, much like the progression from atoms to waves in physics.

Designing learning rules that operate with brain-like performance is far less trivial than substituting a set of artificial neurons with spiking neurons. It would be incredibly elegant if a unified principle governed how the brain learns. But the diversity of neurons, functions, and brain regions imply that a heterogeneous system rich in objectives and synaptic update rules is more likely, and might require us to use all of the weapons in our arsenal of machine learning tools. It is likely that a better understanding of biological learning will be amassed by observing the behavior of a collection of spikes distributed across brain regions. Ongoing advances in procuring large-scale electrophysiological recordings at the neuron-level can give us a window into observing how populations of spikes are orchestrated to handle credit assignment so efficiently, and at the very least, give us a more refined toolkit to developing theories that may advance deep learning \cite{jun2017fully, steinmetz2021neuropixels}. After all, it was not a single atom that led to the silicon revolution, but rather, a mass of particles, and their collective fields. A stronger understanding of the computational benefits of spikes may require us to think at a larger scale, in terms of the `fields' of spikes.

As the known benefits of SNNs manifest in the physical quantities of energy and latency, it will take more than just a machine learning mind to navigate the tangled highways of 100 trillion synapses. It will take a concerted effort between machine learning engineers, neuroscientists, and circuit designers to put spikes in the front seat of deep learning. 


\section*{Acknowledgements}
We would like to thank Sumit Bam Shrestha, Garrick Orchard, Albert Albesa Gonz\'alez and Ruijie Zhu for their insightful discussions over the course of putting together this paper, and iDataMap Corporation for their support.

\section*{Additional Materials}
A series of interactive tutorials complementary to this paper are available in the documentation for our Python package designed for gradient-based learning using spiking neural networks, \textit{snnTorch}\cite{snntorch2021}, at the following link: \url{https://snntorch.readthedocs.io/en/latest/tutorials/index.html}.







\appendix

\renewcommand\thefigure{S.\arabic{figure}}    
\setcounter{figure}{0}

\newpage
\section{Appendix A: From Artificial to Spiking Neural Networks} 
\subsection{Forward Euler Method to Solving Spiking Neuron Models}\label{app:a1}
The time derivative $dU(t)/dt$ is substituted into \Cref{eq:1} without taking the limit $\Delta t \rightarrow 0$:

\begin{equation}\label{aeq:1}
    \tau \frac{U(t+\Delta t) - U(t)}{\Delta t} = -U(t) + I_{\rm in}(t)R
\end{equation}

For small enough values of $\Delta t$, this provides a sufficient approximation of continuous-time integration. Isolating the membrane potential at the next time step on the left side of the equation gives:

\begin{equation}\label{aeq:2}
    U(t+\Delta t) = (1-\frac{\Delta t}{\tau})U(t) + \frac{\Delta t}{\tau}I_{\rm in}(t)R
\end{equation}

To single out the leaky membrane potential dynamics, assume there is no input current $I_{\rm in}(t) = 0 A$:

\begin{equation} \label{aeq:2.1}
    U(t+\Delta t) = (1-\frac{\Delta t}{\tau})U(t)
\end{equation}

Let the ratio of subsequent values of $U$, i.e., $U(t+\Delta t)/U(t)$ be the decay rate of the membrane potential, also known as the inverse time constant. From \Cref{aeq:2}, this implies that $\beta = (1-\Delta t/\tau$). 

Assume $t$ is discretised into sequential time-steps, such that $\Delta t=1$. To further reduce the number of hyperparameters from \Cref{aeq:2}, assume $R=1 \Omega$. This leads to the result in \Cref{eq:euler}, where the following representation is shifted by one time step:

\begin{equation} \label{aeq:2.2}
    \beta = (1-\frac{1}{\tau}) \implies U[t+1] = \beta U[t] + (1-\beta)I_{\rm in}[t+1]
\end{equation}

The input current is weighted by $(1-\beta)$ and time-shifted by one step such that it can instantaneously contribute to membrane potential. While this is not a physiologically precise assumption, it casts the neuron model into a form that better resembles an RNN. $\beta$ can be solved using the continuous-time solution from \Cref{eq:generalsolution}. In absence of current injection:

\begin{equation}\label{aeq:init}
    U(t) = U_0e^{-t/\tau}
\end{equation}

where $U_0$ is the initial membrane potential at $t=0$. Assuming \Cref{aeq:init} is computed at discrete steps of $t$, $(t+\Delta t)$, $(t+2\Delta t) ...$, then the ratio of membrane potential across two subsequent steps can be calculated using:

\begin{equation}\label{aeq:beta}
\begin{split}
    \beta = &\frac{U_0e^{-(t+\Delta t)/\tau}}{U_0e^{-t/\tau}} = \frac{U_0e^{-(t+2\Delta t)/\tau}}{U_0e^{-(t+\Delta t)/\tau}} = ... \\
    \implies & \beta = e^{-\Delta t/\tau}
    \end{split}
\end{equation}

It is preferable to calculate $\beta$ using \Cref{aeq:beta} rather than $\beta = (1-\Delta t/\tau$), as the latter is only precise for $\Delta t << \tau$. This result for $\beta$ can then be used in \Cref{aeq:2.2}.

A second non-physiological assumption is made, where the effect of $(1-\beta)$ is absorbed by a learnable weight $W$:

\begin{equation} \label{aeq:2.3}
    WX[t] = I_{\rm in}[t]
\end{equation}

This can be interpreted the following way. $X[t]$ is an input voltage, spike, or unweighted current, and is scaled by the synaptic conductance $W$ to generate a current injection to the neuron. This leads to the following result:

\begin{equation}
    U[t+1] = \beta U[t] + WX[t+1]
\end{equation}

where the effects of $W$ and $\beta$ are decoupled, thus favouring simplicity over biological precision. 





To arrive at \Cref{eq:2}, a reset function is appended which activates every time an output spike is triggered. The reset mechanism can be implemented by either subtracting the threshold at the onset of a spike as in \Cref{eq:2}, or by forcing the membrane potential to zero:

\begin{equation}\label{aeq:5}
    U[t+1] =\underbrace{\beta U[t]}_{\text{\rm decay}} + \underbrace{WX[t]}_{\text{\rm input}} - \underbrace{S_{\rm out}(\beta U[t] + WX[t])}_{\text{\rm reset-to-zero}}
\end{equation}

In general, reset-by-subtraction is thought to be better for performance as it retains residual superthreshold information, while reset-to-zero is more efficient as $U[t]$ will always be forced to zero when a spike is triggered. This has been formally demonstrated in ANN-SNN conversion approaches (\Cref{sec:shadow}), though has not yet been characterised for natively trained SNNs. The two approaches will converge for a small enough time window where $U[t]$ is assumed to increase in a finite period of time:

\begin{figure}[!ht]
    \centering
    \includegraphics[scale=0.5]{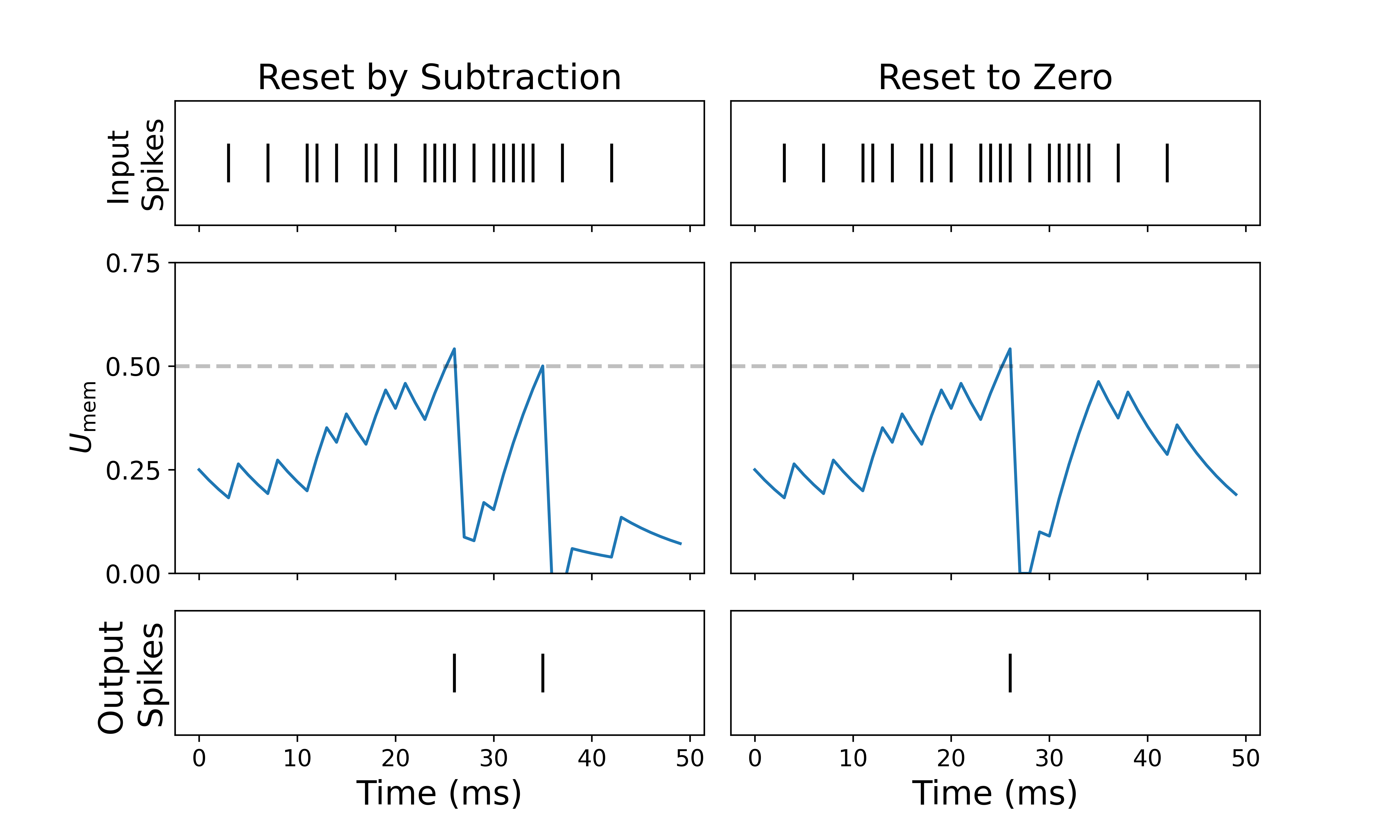}
    \caption{Reset by subtraction vs reset-to-zero. Threshold set to $\theta = 0.5$.}
    \label{fig:sa1}
\end{figure}





\section{Appendix B: Spike Encoding}
The following spike encoding mechanisms and loss functions are described with respect to a single sample of data. They can be generalised to multiple samples as is common practice in deep learning to process data in batches. 

\subsection{Rate Coded Input Conversion} \label{app:a2}
An example of conversion of an input sample to a rate coded spike train follows. Let $\vect{X} \in \mathbb{R}^{m \times n}$, be a sample from the MNIST dataset, where $m=n=28$. We wish to convert $\vect{X}$ to a rate-coded 3-D tensor $\tensorsym{R} \in \mathbb{R}^{m \times n \times t}$, where $t$ is the number of time steps. Each feature of the original sample $X_{ij}$ is encoded separately, where the normalised pixel intensity (between 0 and 1) is the probability a spike occurs at any given time step. This can be treated as a Bernoulli trial, a special case of the binomial distribution $R_{ijk}\sim B(n, p)$ where the number of trials is $n=1$, and the probability of success (spiking) is $p=X_{ij}$. Explicitly, the probability a spike occurs is:

\begin{equation}\label{beq:1}
    {\rm P}(R_{ijk}=1) = X_{ij} = 1 - {\rm P}(R_{ijk}=0)
\end{equation}

Sampling from the Bernoulli distribution for every feature at each time step will populate the 3-D tensor $\tensorsym{R}$ with 1's and 0's. For an MNIST image, a pure white pixel $X_{ij}=1$ corresponds to a 100\% probability of spiking. A pure black pixel $X_{ij}=0$ will never generate a spike. A gray pixel of value $X_{ij}=0.5$ will have an equal probability of sampling either a `1' or a `0'. As the number of time steps $t \rightarrow \infty$, the proportion of spikes is expected to approach 0.5. 

\begin{figure}[!ht]
    \centering
    \includegraphics{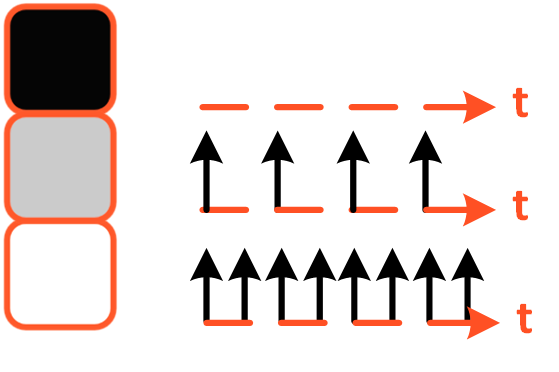}
    \caption{Rate coded input pixel. An input pixel of greater intensity corresponds to a higher firing rate.}
    \label{fig:sb1}
\end{figure}

\subsection{Latency Coded Input Conversion} \label{app:a3}
The logarithmic dependence between input feature intensity and spiking timing can be derived using an RC circuit model. Starting with the general solution of the membrane potential with respect to the input current in \Cref{eq:generalsolution} and nulling out the initial conditions $U_0=0$, we obtain:

\begin{equation}\label{beq:2}
    U(t) = I_{\rm in}R(1 - e^{-t/\tau})
\end{equation}

For a constant current injection, $U(t)$ will exponentially relax towards a steady-state value of $I_{\rm in}R$. Say a spike is emitted when $U(t)$ reaches a threshold $\theta$. We solve for the time $U(t)=\theta$:

\begin{equation}\label{beq:3}
    t=\tau\Big[{\rm ln}\Big(\frac{I_{\rm in}R}{I_{\rm in}R-\theta}\Big)\Big]
\end{equation}

The larger the input current, the faster $U(t)$ charges up to $\theta$, and the faster a spike occurs. The steady-state potential, $I_{\rm in}R$ is set to the input feature $x$:

\begin{equation}\label{beq:4}
    t(x) =
    \begin{cases}
      \tau\Big[{\rm ln}\Big(\frac{x}{x-\theta}\Big)\Big], & x > \theta\\
      \infty, & \text{otherwise} \\
    \end{cases}  
\end{equation}

\begin{figure}[!ht]
    \centering
    \includegraphics{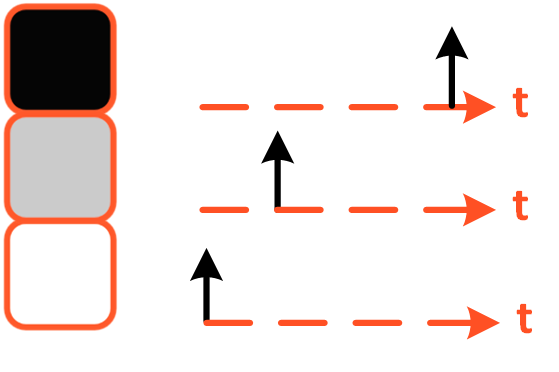}
    \caption{Latency coded input pixel. An input pixel of greater intensity corresponds to an earlier spike time.}
    \label{fig:sb2}
\end{figure}

\subsection{Rate Coded Outputs} \label{app:a4}
A vectorised implementation of determining the predicted class from rate-coded output spike trains is described. Let $\vec{S}[t] \in \mathbb{R}^{N_C}$ be a time-varying vector that represents the spikes emitted from each output neuron across time, where $N_C$ is the number of output classes. Let $\vec{c}\in \mathbb{R}^{N_C}$ be the spike count from each output neuron, which can be obtained by summing $\vec{S[t]}$ over $T$ time steps:

\begin{equation}\label{beq:5}
    \vec{c} = \sum_{j=0}^T\vec{S}[t]
\end{equation}

The index of $\vec{c}$ with the maximum count corresponds to the predicted class:

\begin{equation}\label{beq:6}
    \hat{y} = \argmax_ic_i
\end{equation}

\begin{figure}[!h]
    \centering
    \includegraphics[scale=0.75]{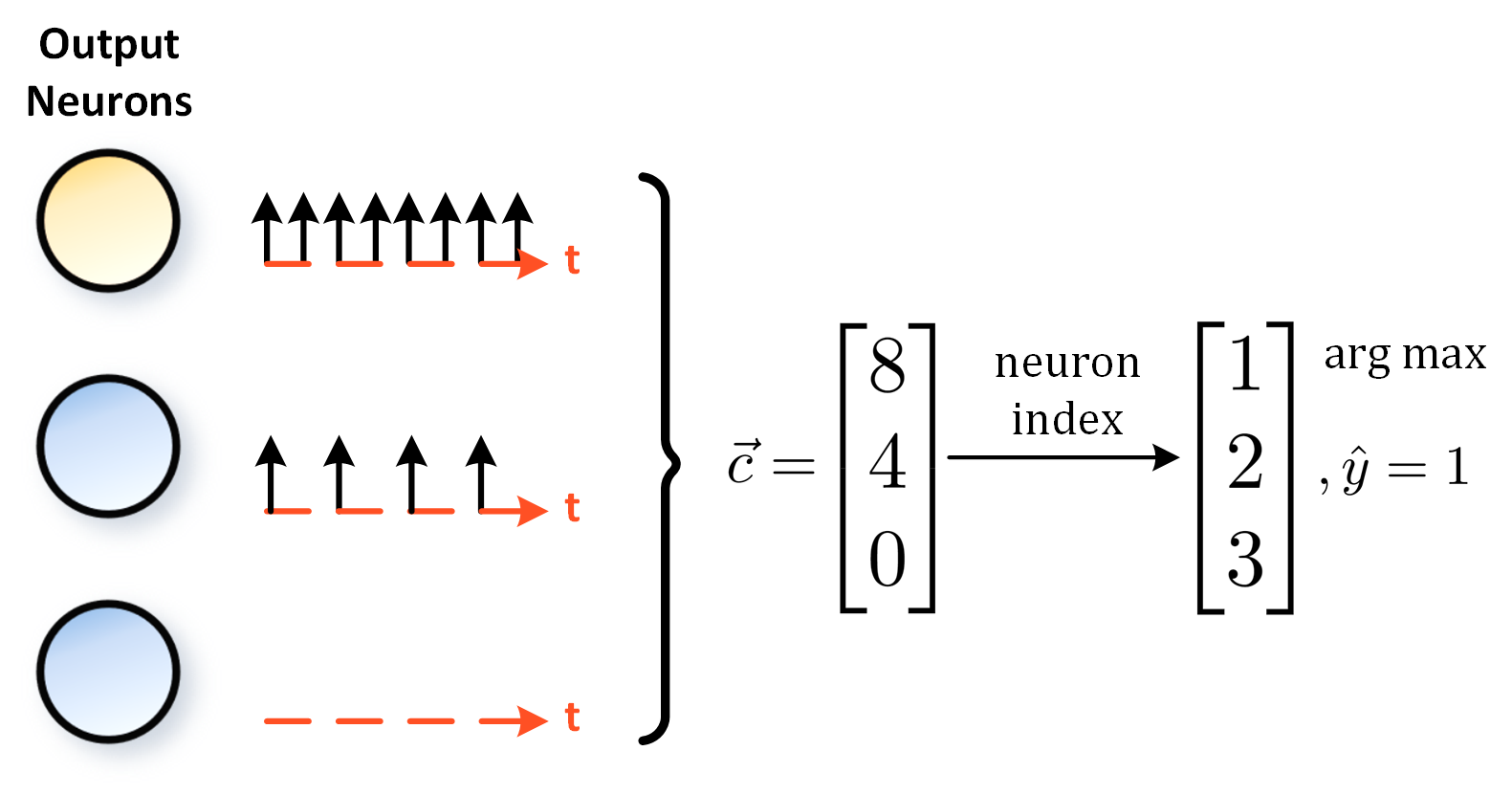}
    \caption{Rate coded outputs. $\vec{c} \in \mathbb{R}^{N_C}$ is the spike count from each output neuron, where the example above shows the first neuron firing a total of 8 times. $\hat{y}$ represents the index of the predicted output neuron, where it indicates the first neuron is the correct class.}
    \label{fig:s4}
\end{figure}

\subsection{Cross Entropy Spike Rate} \label{app:a5}
The spike count of the output layer $\vec{c}\in \mathbb{R}^{N_C}$ is obtained as in \Cref{beq:5}. $c_i$ is the $i^{th}$ element of $\vec{c}$, treated as the logits in the softmax function:
    
\begin{equation}\label{beq:7}
    p_i=\frac{e^{c_i}}{\sum_{i=1}^{N_C}e^{c_i}}
\end{equation}

The cross entropy between $p_i$ and the target $y_i \in \{0,1\}^{N_C}$, which is a one-hot target vector, is obtained using:

\begin{equation}\label{beq:8}
     \mathcal{L}_{CE} = \sum_{i=0}^{N_C}y_i{\rm log}(p_i)
\end{equation}

\begin{figure}[!ht]
    \centering
    \includegraphics[scale=0.75]{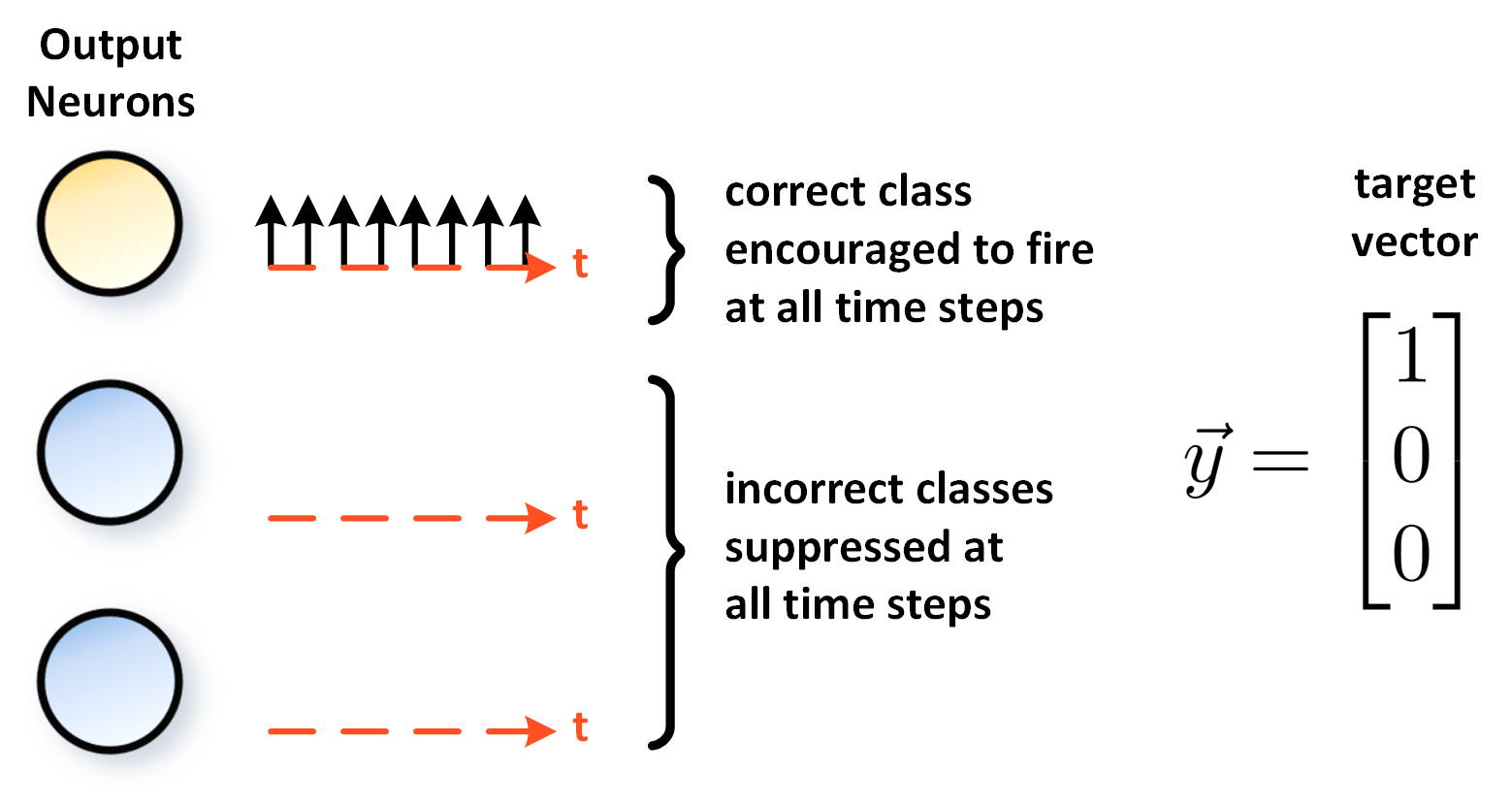}
    \caption{Cross Entropy Spike Rate. The target vector $\vec{y}$ specifies the correct class as a one-hot encoded vector.}
    \label{fig:s5}
\end{figure}

\subsection{Mean Square Spike Rate}\label{app:a6}
As in \Cref{beq:5}, the spike count of the output layer $\vec{c}\in \mathbb{R}^{N_C}$ is obtained. $c_i$ is the $i^{th}$ element of $\vec{c}$, and let $y_i\in \mathbb{R}$ be the target spike count over a period of time $T$ for the $i^{th}$ output neuron. The target for the correct class should be greater than that of incorrect classes:

\begin{equation}\label{beq:9}
    \mathcal{L}_{MSE} = \sum_i^{N_C}(y_i - c_i)^2
\end{equation}

\begin{figure}[!ht]
    \centering
    \includegraphics[scale=0.75]{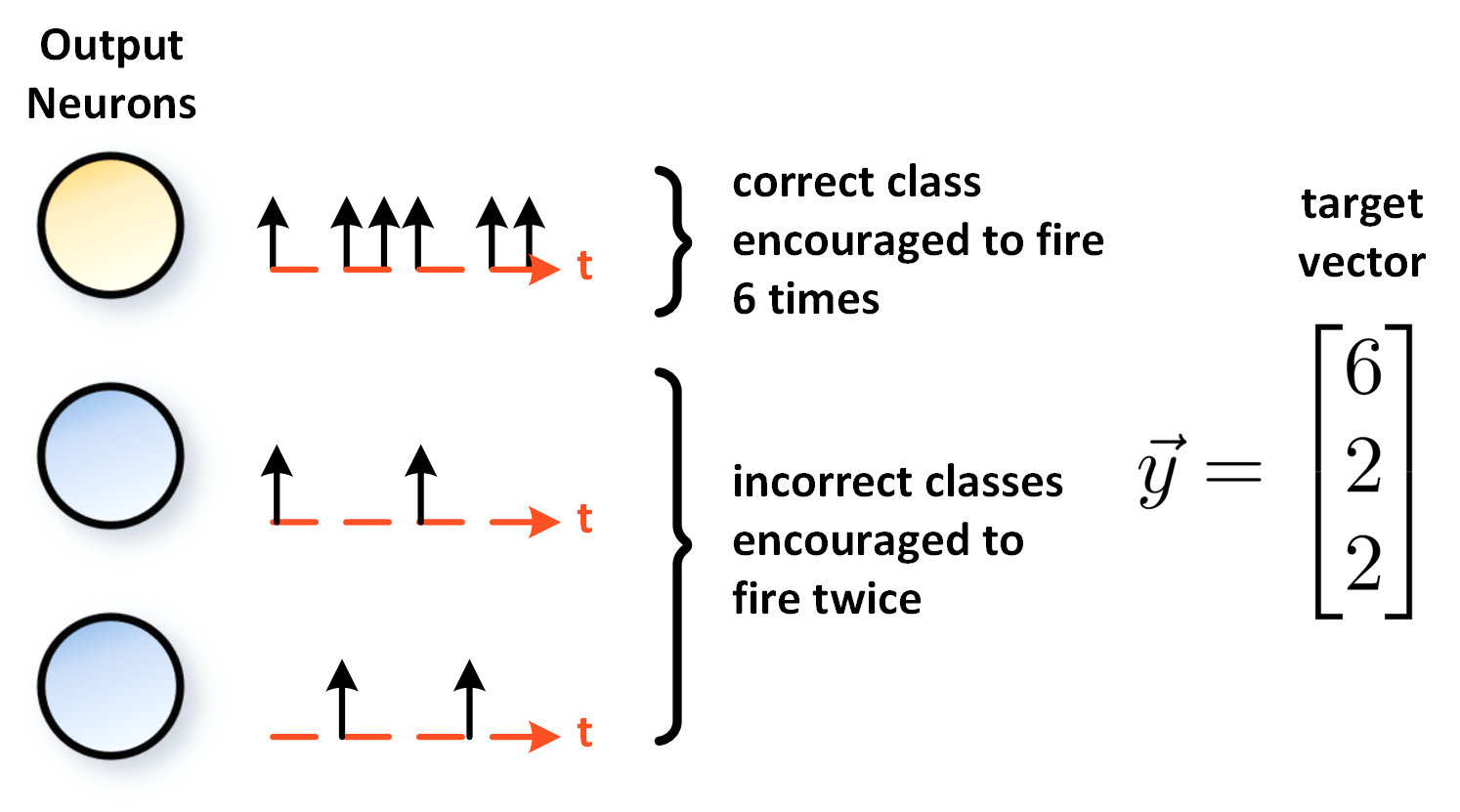}
    \caption{Mean Square Spike Rate. The target vector $\vec{y}$ specifies the total desired number of spikes for each class.}
    \label{fig:s6}
\end{figure}


\subsection{Maximum Membrane}\label{app:a7}
The logits $\vec{m}\in \mathbb{R}^{N_C}$ are obtained by taking the maximum value of the membrane potential of the output layer $\vec{U}[t]\in \mathbb{R}^{N_C}$ over time:

\begin{equation}\label{beq:10}
    \vec{m} = {\rm max}_t\vec{U}[t]
\end{equation}
The elements of $\vec{m}$ replace $c_i$ in the softmax function from \Cref{beq:7}, with the cross entropy of the result measured with respect to the target label.

\begin{figure}[!ht]
    \centering
    \includegraphics[scale=0.75]{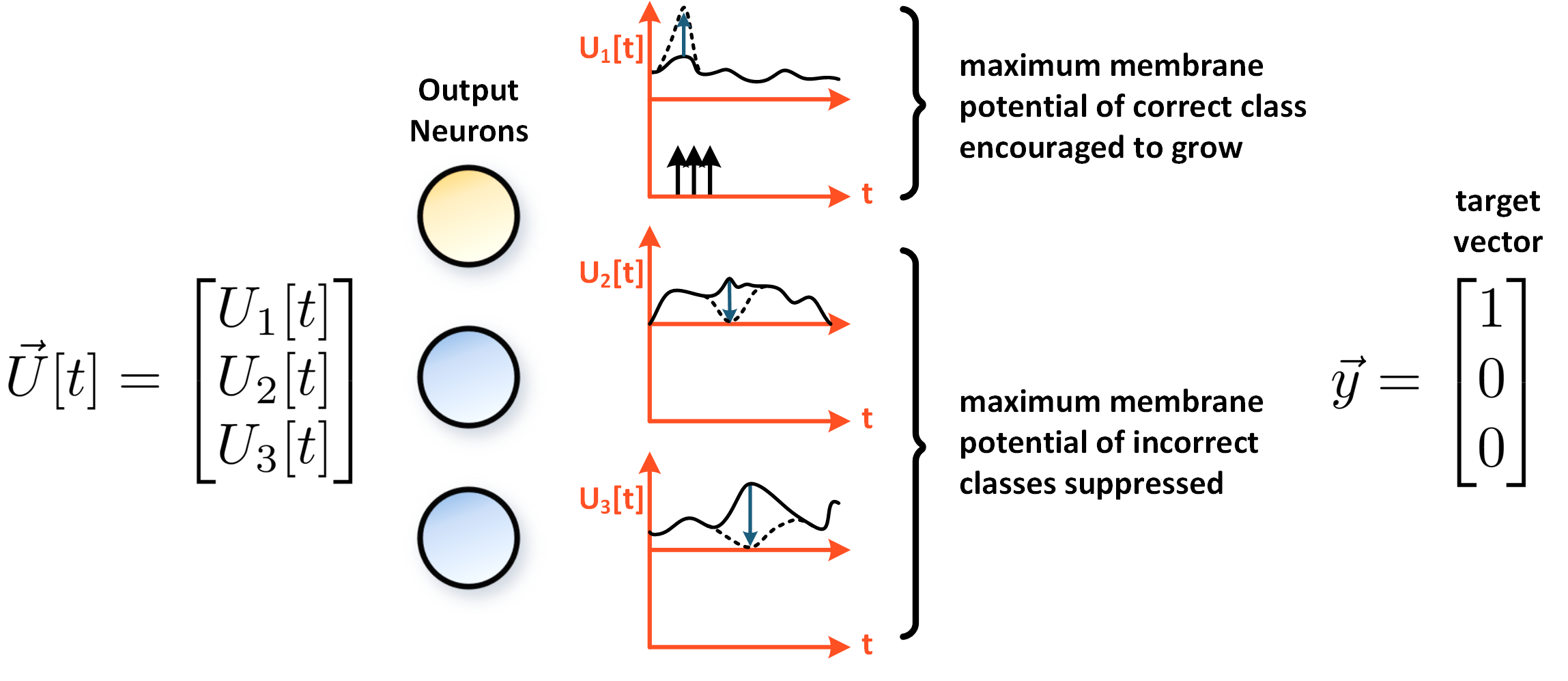}
    \caption{Maximum Membrane. The peak membrane potential for each neuron is used in the cross entropy loss function. This encourages the peak of the correct class to grow, while that of the incorrect class is suppressed. The effect of this is to promote more firing from the correct class and less from the incorrect class.}
    \label{fig:s7}
\end{figure}

Alternatively, the membrane potential is summed over time to obtain the logits:

\begin{equation}
    \vec{m} = \sum_t^T\vec{U}[t]
\end{equation}

\subsection{Mean Square Membrane}\label{app:a11}
Let $y_{i}[t]$ be a time-varying value that specifies the target membrane potential of the $i^{th}$ neuron at each time step. The total mean square error is calculated by summing the loss for all $T$ time steps and for all $N_C$ output layer neurons:

\begin{equation}\label{beq:15}
    \mathcal{L}_{MSE} = \sum_i^{N_C}\sum_t^T(y_i[t]-U[t])^2
\end{equation}

Alternatively, the time-varying target $y_{i}[t]$ can be replaced with a time-static target to drive the membrane potential of all neurons to a constant value. This can be an efficient implementation for a rate code, where the correct class target exceeds the threshold and all other targets are subthreshold values.

\begin{figure}[!ht]
    \centering
    \includegraphics[scale=0.75]{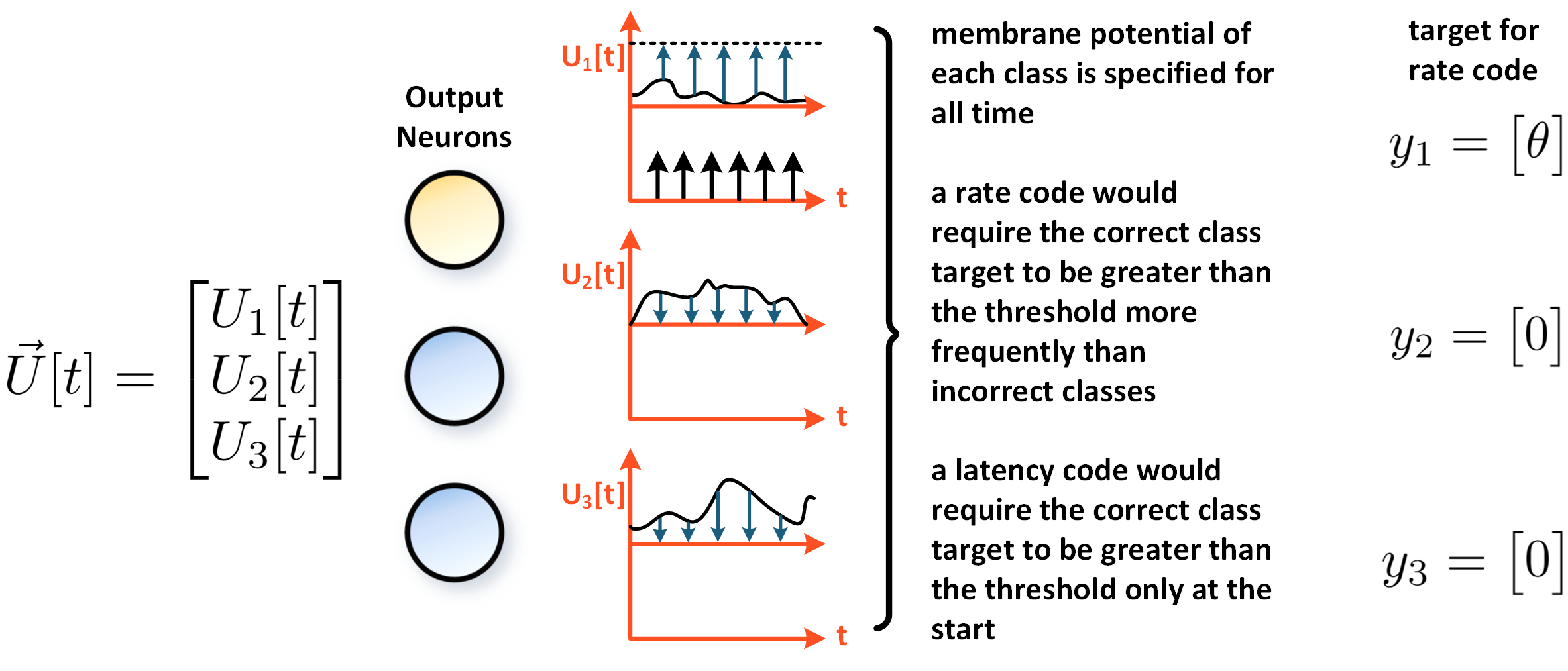}
    \caption{Mean Square Membrane. The membrane potential at each time step is applied to the mean square error loss function. This allows a defined membrane target. The example above sets the target at all time steps at the firing threshold for the correct class, and to zero for incorrect classes.}
    \label{fig:sb10}
\end{figure}

\subsection{Cross Entropy Latency Code}\label{app:a8}
Let $\vec{f}\in\mathbb{R}^{N_C}$ be a vector containing the first spike time of each neuron in the output layer. Cross entropy minimisation aims to maximise the logit of the correct class and reduce the logits of the incorrect classes. However, we wish for the correct class to spike first, which corresponds to a smaller value. Therefore, a monotonically decreasing function must be applied to $\vec{f}$. A limitless number of options are available. The work in \cite{zhang2020spike} simply negates the spike times:

\begin{equation}\label{beq:11}
    \vec{f}:=-\vec{f}
\end{equation}

Taking the inverse of each element $f_i$ of $\vec{f}$ is also a valid option:
\begin{equation}\label{beq:12}
    f_i := \frac{1}{f_i}
\end{equation}

The new values of $f_i$ then replace $c_i$ in the softmax function from \Cref{beq:7}. \Cref{beq:12} must be treated with care, as it precludes spikes from occurring at $t=0$, otherwise $f_i \rightarrow \infty$.

\begin{figure}[!ht]
    \centering
    \includegraphics[scale=0.75]{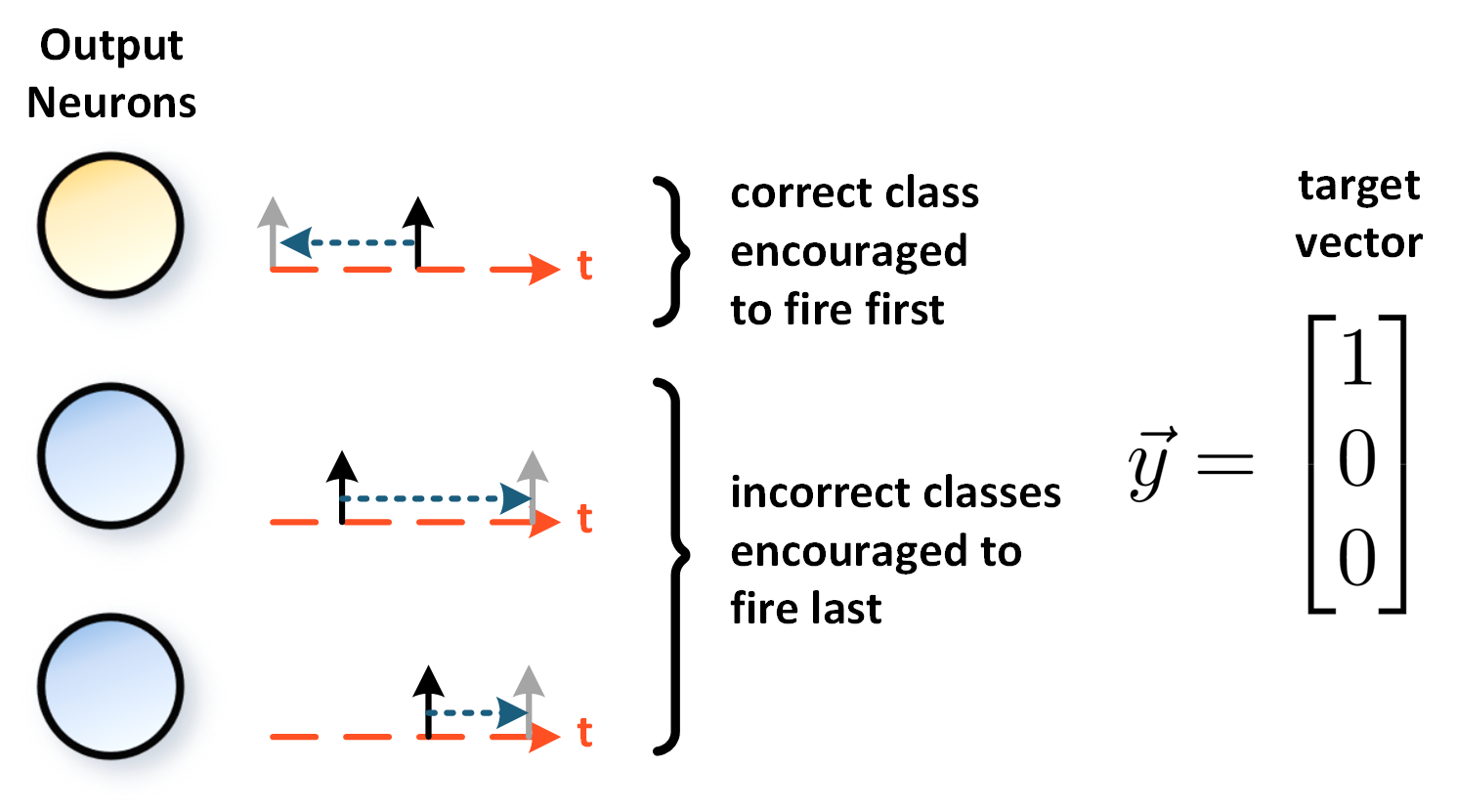}
    \caption{Cross Entropy Latency Code. Applying the inverse (or negated) spike time to the cross entropy loss pushes the correct class to fire first, and incorrect classes to fire later.}
    \label{fig:s8}
\end{figure}


\subsection{Mean Square Spike Time}\label{app:a9}
The spike time(s) of all neurons are specified as targets. In the case where only the first spike matters, $\vec{f}\in\mathbb{R}^{N_C}$ contains the first spike time of each neuron in the output layer, $y_i\in \mathbb{R}$ is the target spike time for the $i^{th}$ output neuron. The mean square errors between the actual and target spike times of all output classes are summed together:

\begin{equation}\label{beq:13}
    \mathcal{L}_{MSE} = \sum_i^{N_C}(y_i - f_i)^2
\end{equation}

This can be generalised to account for multiple spikes \cite{shrestha2018slayer}. In this case, $\vec{f_i}$ becomes a list of emitted spike times and $\vec{y_i}$ becomes a vector desired spike times for the $i^{th}$ neuron, respectively. The $k^{th}$ spike is sequentially taken from $\vec{f_i}$ and $\vec{y_i}$, and the mean square error between the two is calculated. This process is repeated $n$ times, where $n$ is the number of spike times that have been specified and the errors are summed together across spikes and classes:

\begin{equation}\label{beq:14}
    \mathcal{L}_{MSE} = \sum_k^{n}\sum_i^{N_C}(y_{i,k} - f_{i,k})^2
\end{equation}

\begin{figure}[!ht]
    \centering
    \includegraphics[scale=0.75]{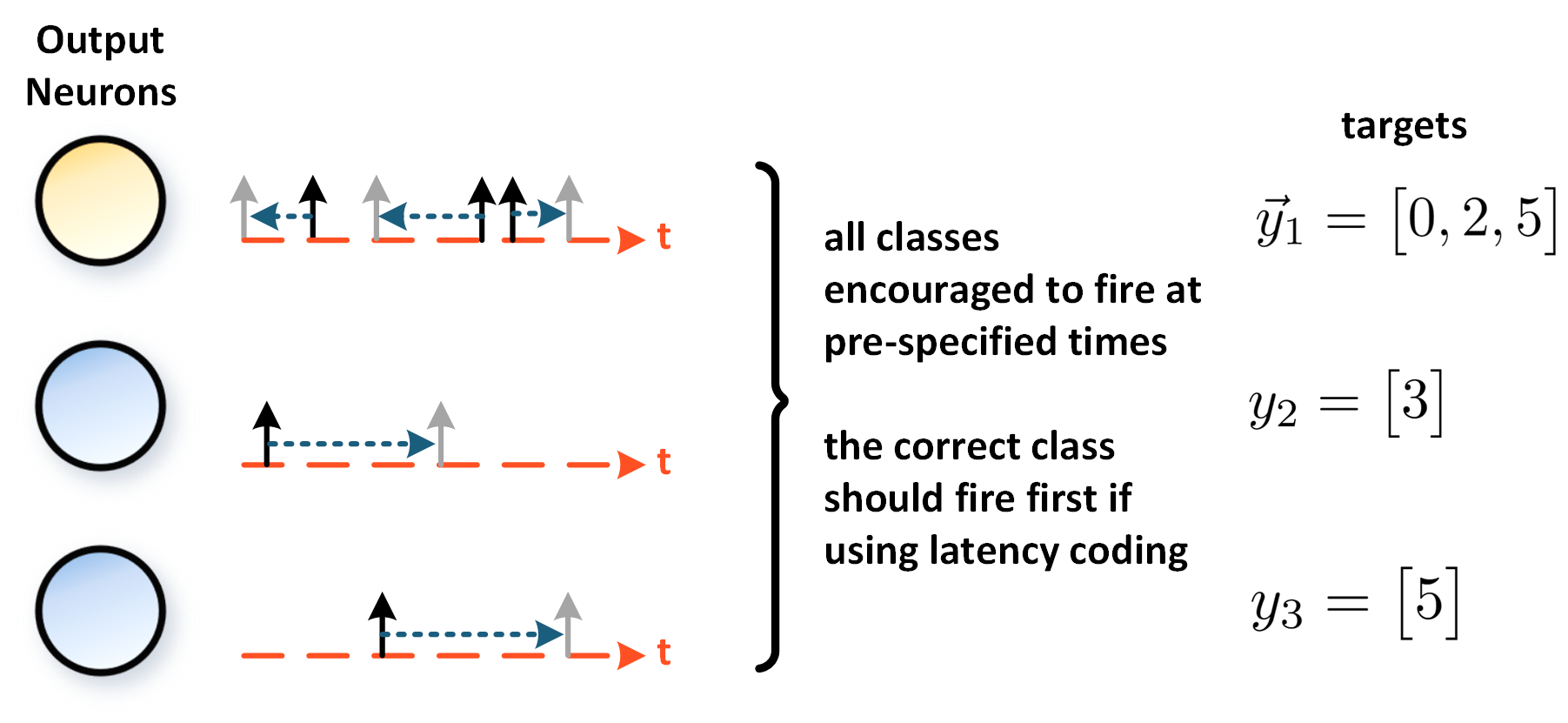}
    \caption{Mean Square Spike Time. The timing of all spikes are iterated over, and sequentially applied to the mean square error loss function. This enables the timing for multiple spikes to be precisely defined.}
    \label{fig:s9}
\end{figure}

\subsection{Mean Square Relative Spike Time} \label{app:a10}
The difference between the spike time of correct and incorrect neurons is specified as a target. As in \Cref{app:a9}, $y_i$ is the desired spike time for the $i^{th}$ neuron and $f_i$ is the actual emitted spike time. The key difference is that $y_i$ can change throughout the training process.

Let the minimum possible spike time be $f_0 \in \mathbb{R}$. This sets the target firing time of the correct class. The target firing time of incorrect neuron classes $y_i$ is set to:

\begin{equation}
y_i =
    \begin{cases}
     f_0 + \gamma, & \text{\rm if $f_i<f_0+\gamma$}\\
     f_i, & \text{\rm if $f_i \geq f_0+\gamma$} \\
    \end{cases}
\end{equation}

where $\gamma$ is a pre-defined latency, treated as a hyperparameter. In the first case, if an incorrect neuron fires at some time before the latency period $\gamma$ then a penalty will be applied. In the second case, where the incorrect neuron fires at $\gamma$ steps after the correct neuron, then the target is simply set to the actual spike time. These zero each other out during the loss calculation. This target $y_i$ is then applied to the mean square error loss (\Cref{beq:14}). 

\begin{figure}[!ht]
    \centering
    \includegraphics[scale=0.75]{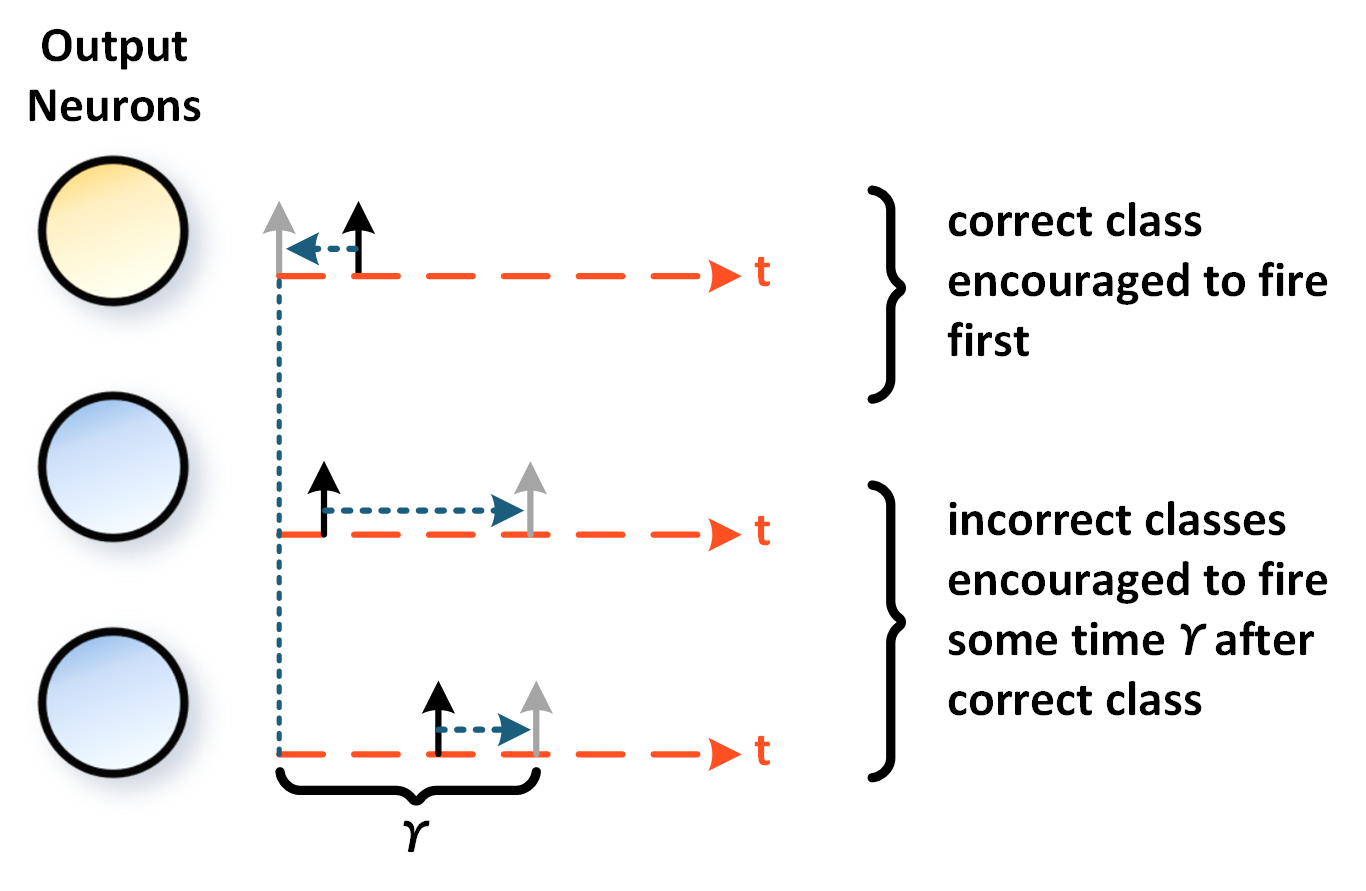}
    \caption{Mean Square Relative Spike Time. The relative timing between all spikes are applied to the mean square error loss function, enabling a defined time window $\gamma$ to occur between the correct class firing and incorrect classes firing.}
    \label{fig:s10}
\end{figure}

\subsection{Population Level Regularisation}\label{app:regpop}
L1-regularisation can be applied to the total number of spikes emitted at the output layer to penalise excessive firing \cite{zenke2019spytorch}, thus encouraging sparse activity at the output: 

\begin{equation}
     \mathcal{L}_{L1} = \lambda_1\sum_t^T\sum_i^{N_C}S_{i}[t]
\end{equation}

where $\lambda_1$ is a hyperparameter controlling the influence of the regularisation term, and $S_{i}[t]$ is the spike of the $i^{th}$ class at time $t$. 

Alternatively, an upper-activity threshold $\theta_U$ can be applied where if the total number of spikes for \textit{all} neurons in layer $l$ exceeds this threshold, only then does the regularisation penalty apply:

\begin{equation}
    \mathcal{L}_{U} = \lambda_U\Big(\Big[\sum_i^{N}c_{i}^{(l)} - \theta_U\Big]_+\Big)^L
\end{equation}

where $c_{i}$ is the total spike count over time for the $i^{th}$ neuron in layer $l$, and $N$ is the total number of neurons in layer $l$. $\lambda_U$ is a hyperparameter influencing the strength of the upper-activity regularisation, and $[\cdot]_+$ is a linear rectification: if the total number of spikes from the layer is less than $\theta_U$, the rectifier clips the negative result to zero such that a penalty is not added. $L$ is typically chosen to be either 1 or 2 \cite{zenke2021remarkable}. It is possible to swap out the spike count for a time-averaged membrane potential as well, if using hidden-state variables is permissible \cite{kaiser2020synaptic}.

\subsection{Neuron Level Regularisation}\label{app:regneu}
A lower-activity threshold $\theta_L$ that specifies the lower permissible limit of firing for \textit{each} neuron before the regularisation penalty is applied:

\begin{equation}
    \mathcal{L}_{L} = \frac{\lambda_L}{N}\sum_i^N\Big(\Big[\theta_L - c_i^{(l)}\Big]_+\Big)^2
    \end{equation}

The rectification $[\cdot]_+$ now falls within the summation, and is applied to the firing activity of each individual neuron, rather than a population of neurons, where $\lambda_L$ is a hyperparameter that influences the strength of lower-activity regularisation \cite{zenke2021remarkable}. As with population-level regularisation, the spike count can also be substituted for a time-averaged membrane potential \cite{kaiser2020synaptic}.

\section{Appendix C: Training Spiking Neural Networks}
\subsection{Backpropagation Using Spike Times}\label{app:c1}
In the original description of SpikeProp from \cite{bohte2002error}, a spike response model is used:

\begin{equation} \label{ceq:srm_mem}
    U_j(t) = \sum_{i,k} W_{i,j}I_i^{(k)}(t),\nonumber
\end{equation}
\begin{equation} \label{ceq:srm_mem2}
    I_i^{(k)}(t)=\epsilon(t-f_i^{(k)}),
\end{equation}

where $W_{i,j}$ is the weight between the $i^{th}$ presynaptic and $j^{th}$ postsynaptic neurons, $f_i^{(k)}$ is the firing time of the $k^{th}$ spike from the $i^{th}$ presynaptic neuron, and $U_j(t)$ is the membrane potential of the $j^{th}$ neuron. 
For simplicity, the `alpha function' defined below is frequently used for the kernel:



\begin{equation} \label{ceq:ker}
    \epsilon(t) = \frac{t}{\tau}e^{1-\frac{t}{\tau}}\Theta\left(t\right),
\end{equation}

where $\tau$ and $\Theta$ are the time constant of the kernel and Heaviside step function, respectively. 

Consider an SNN where each target specifies the timing of the output spike emitted from the $j^{th}$ output neuron ($y_j$). This is used in the mean square spike time loss (\Cref{beq:13}, \Cref{app:a9}), where $f_j$ is the actual spike time. Rather than backpropagating in time through the entire history of the simulation, only the gradient pathway through the spike time of each neuron is taken. The gradient of the loss in weight space is then:


\begin{equation} \label{eq:spikeprop}
    \frac{\partial \mathcal{L}}{\partial W_{i,j}} = \frac{\partial \mathcal{L}}{\partial f_j}\frac{\partial f_j}{\partial U_j}\frac{\partial U_j}{\partial W_{i,j}}\Bigr|_{t=f_j}.
\end{equation}

The first term on the right side evaluates to:

\begin{equation}
    \frac{\partial \mathcal{L}}{\partial f_j} = 2(y_j - f_j).
\end{equation}

The third term can be derived from \Cref{ceq:srm_mem}:


\begin{equation}
    \frac{\partial U_{j}}{\partial W_{i,j}}\Bigr|_{t=f_j} = \sum_{k}I_i^{(k)}(f_j) = \sum_k\epsilon(f_j-f_i^{(k)}).
\end{equation}

The second term in \Cref{eq:spikeprop} can be calculated by calculating $-\partial U_j/\partial f_j\Bigr|_{t=f_j}$ instead, and then taking the inverse. In \cite{bohte2002error}, the evolution of $U_j(t)$ 
can be analytically solved using \Cref{ceq:srm_mem,ceq:ker}:

\begin{equation}\label{ceq:appr} 
    \frac{\partial f_j}{\partial U_j} \leftarrow -\left(\frac{\partial U_j}{\partial t}\Bigr|_{t=f_j}\right)^{-1} = -\left(\sum_{i,k} W_{i,j}\frac{\partial I_i^{(k)}}{\partial t}\Bigr|_{t=f_j}\right)^{-1} = 
    \Bigg(\sum_{i,k}W_{i,j}\frac{f_j-f_i^{(k)}-\tau}{\tau^2}\bigg(e^{\frac{f_j-f_i^{(k)}}{\tau}-1}\bigg)\Bigg)^{-1}.
\end{equation} 

\begin{figure}[!ht]
    \centering
    \includegraphics[scale=0.85]{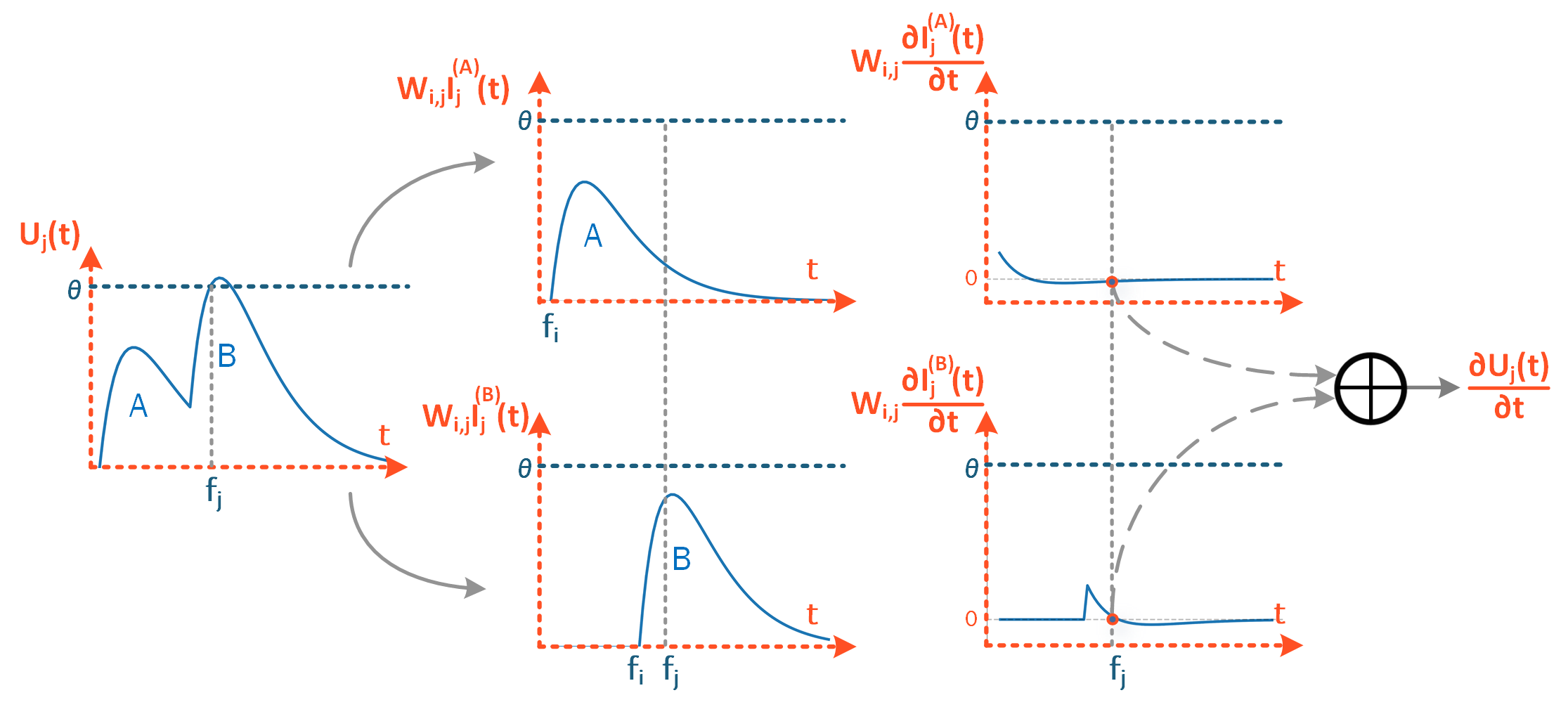}
    \caption{Calculation of derivative of membrane potential with respect to spike time. The superscripts $^{(A)}$ and $^{(B)}$ denote the separate contributions from each application of the kernel.}
    \label{fig:sc1}
\end{figure}

Note, the input current is triggered at the onset of the pre-synaptic spike $t=f_i$, but is evaluated at the time of the post-synaptic spike $t=f_j$. The results can be combined to give:

\begin{equation}
    \frac{\partial\mathcal{L}}{\partial W_{i,j}} = -\frac{2(y_j-f_j)\sum_k I_i^{(k)}(f_j)}{\sum_{i,k}W_{i,j}(\partial I_j^{(k)}/\partial t)\Bigr|_{t=f_j}}
\end{equation}

This approach can be generalized to handle deeper layers, and the original formulation also includes delayed response kernels that are not included above for simplicity. 

\subsection{Backpropagation Using Spikes}\label{app:bpus}
\textbf{Spike Timing Dependent Plasticity}

The connection between a pair of neurons can be altered by the spikes emitted by both neurons. Several experiments have shown the relative timing of spikes between pre- and post-synaptic neurons can be used to define a learning rule for updating the synaptic weight \cite{bi1998synaptic}. Let $t_{\rm pre}$ and $t_{\rm post}$ represent the timing of the pre- and post-synaptic spikes, respectively. The difference in spike time is:

\begin{equation}
    \Delta t = t_{\rm pre} - t_{\rm post}
\end{equation}

\begin{figure}[!ht]
    \centering
    \includegraphics[scale=1]{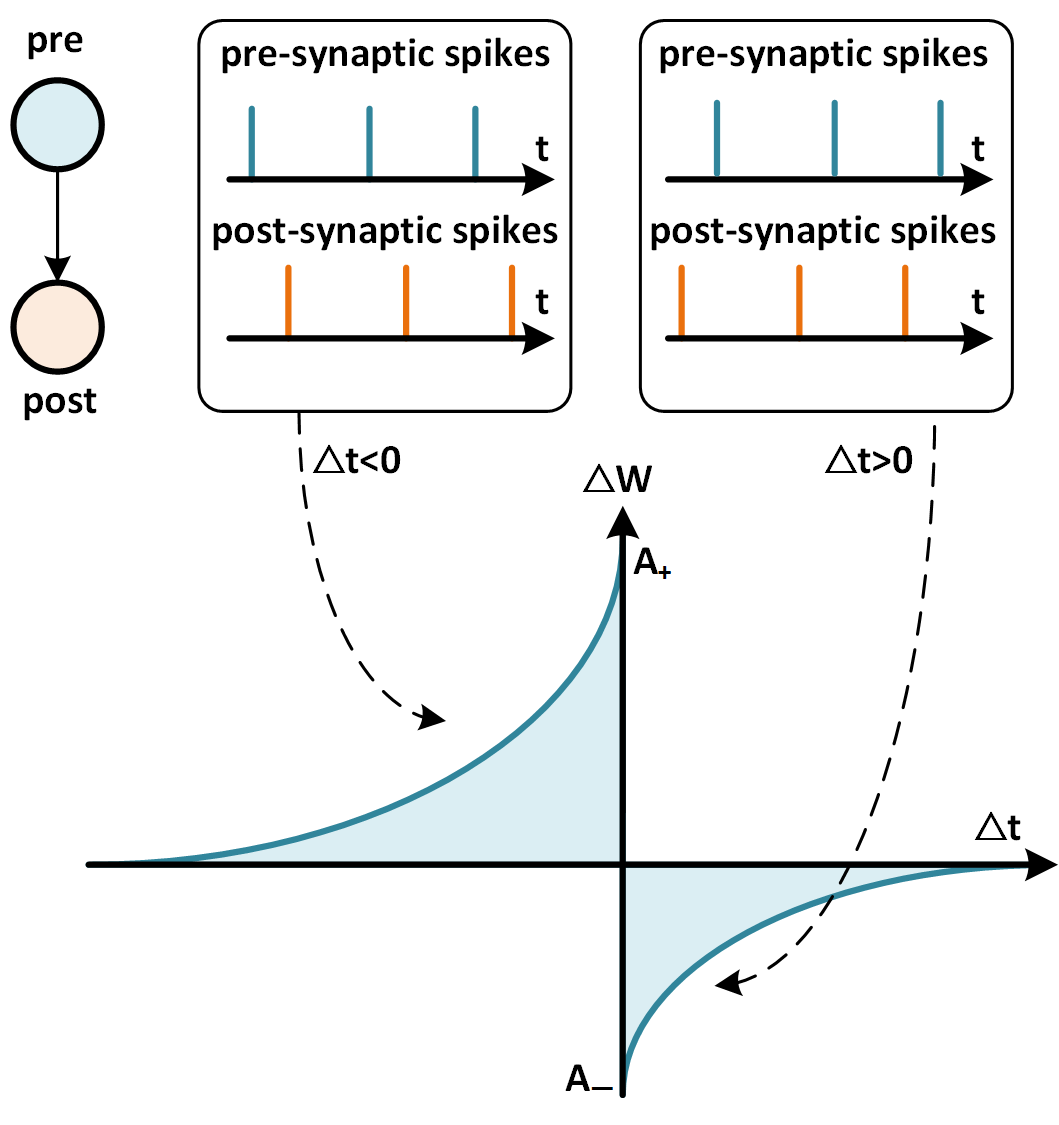}
    \caption{STDP Learning Window. If the pre-synaptic neuron spikes before the post-synaptic neuron, $\Delta t < 0 \implies \Delta W > 0$ and the synaptic strength between the two neurons is increased. If the pre-synaptic neuron spikes after the post-synaptic neuron, $\Delta t > 0 \implies \Delta W < 0$ and the synaptic strength is decreased.}
    \label{fig:stdp}
\end{figure}

When the pre-synaptic neuron emits a spike before the post-synaptic neuron, such that the pre-synaptic spike may have caused the post-synaptic spike, then the synaptic strength is expected to increase (`potentiation'). When reversed, i.e., the post-synaptic neuron spikes before the pre-synaptic neuron, the synaptic strength decreases (`depression'). This rule is known as spike timing dependent plasticity (STDP), and has been shown to exist in various brain regions including the visual cortex, somatosensory cortex and the hippocampus. Fitting curves to experimental measurements take the following form \cite{bi1998synaptic}:

\begin{equation} \label{eq:appc}
    \Delta W =
    \begin{cases}
      A_+e^{\Delta t/\tau_+},  & \text{\rm if $t_{\rm post} > t_{\rm pre}$} \\ 
      A_-e^{-\Delta t/\tau_-},  & \text{\rm if $t_{\rm post} < t_{\rm pre}$}
    \end{cases}  
\end{equation}

where $\Delta W$ is the change in synaptic weight, $A_+$ and $A_-$ represent the maximum amount of synaptic modulation that takes place as the difference between spike times approaches zero, $\tau_+$ and $\tau_-$ are the time constants that determine the strength of the update over a given interspike interval. This mechanism is illustrated in \Cref{fig:stdp}.

For a strong, excitatory synaptic connection, a pre-synaptic spike will trigger a large post-synaptic potential (refer to $U$ in \Cref{eq:2}). As membrane potential approaches the threshold of neuronal firing, such an excitatory case suggests that a post-synaptic spike will likely follow a pre-synaptic spike. This will lead to a positive change of the synaptic weight, thus increasing the chance that a post-synaptic spike will follow a pre-synaptic spike in future. This is a form of causal spiking, and STDP reinforces causal spiking by continuing to increase the strength of the synaptic connection.

Input sensory data is typically correlated in both space and time, so a network's response to a correlated spike train will be to increase the weights much faster than uncorrelated spike trains. This is a direct result of causal spiking. Intuitively, a group of correlated spikes from multiple pre-synaptic neurons will arrive at a post-synaptic neuron within a close time interval, causing stronger depolarization of the neuron membrane potential, and a higher probability of a post-synaptic spike being triggered.

However, without an upper bound, this will lead to unstable and indefinitely large growth of the synaptic weight. In practice, an upper limit should be applied to constrain potentiation. Alternatively, homeostatic mechanisms can also be used to offset this unbounded growth, such as an adaptive threshold that increases each time a spike is triggered from the neuron (\Cref{app:c2}).

\subsection{Long-Term Temporal Dependencies} \label{app:c2}
One of the simplest implementations of an adaptive threshold is to choose a steady-state threshold $\theta_0$ and a decay rate $\alpha$:

\begin{equation}
    \theta[t] = \theta_0 + b[t]
\end{equation}
\begin{equation}
    b[t+1] = \alpha b[t] + (1-\alpha)S_{\rm out}[t]
\end{equation}

Each time a spike is triggered from the neuron, $S_{\rm out}[t]=1$, the threshold jumps by $(1-\alpha)$. This is added to the threshold through an intermediary state variable, $b[t]$. This jump decays at a rate of $\alpha$ at each subsequent step, causing the threshold to tend back to $\theta_0$ in absence of further spikes. The above form is loosely based on \cite{bellec2018long}, though the decay rate $\alpha$ and threshold jump factor $(1-\alpha)$ can be decoupled from each other. $\alpha$ can be treated as either a hyperparameter or a learnable parameter.

\bibliographystyle{unsrt}  
\bibliography{references}  

\end{document}